%% file: main.tex
\documentclass[11pt]{article}
\usepackage[margin=1in]{geometry}
\usepackage{amsmath,amssymb,amsthm}
\usepackage{booktabs}
\usepackage{longtable}
\usepackage{xcolor}
\usepackage{listings}
\usepackage{appendix}
\usepackage{graphicx}
\usepackage{float}
\usepackage{authblk}
\graphicspath{{.}{figures/}}

\usepackage{hyperref}
\hypersetup{
    colorlinks=true,  
    linkcolor=black,  
    citecolor=blue,   
    urlcolor=blue     
}

\newcommand{\Fmac}{\textsf{F}}
\newcommand{\sFmac}{\textsf{sF}}
\newcommand{\sP}{\ensuremath{\mathrm{s}P}}
\newcommand{\code}[1]{\texttt{\small #1}}
\newcommand{\NONHALT}{\ensuremath{\bot}}

\newcommand\blfootnote[1]{%
  \begingroup
  \renewcommand\thefootnote{}\footnote{#1}%
  \addtocounter{footnote}{-1}%
  \endgroup
}

\title{Measuring in-context algorithmic reasoning in language
models against an exact Bayes-optimal reference}

\author[1]{Luan Ozelim}
\author[1,2]{Hector Zenil\thanks{Corresponding author: hector.zenil@kcl.ac.uk}}

\affil[1]{\normalsize\text{ }Oxford Immune Algorithmics, Oxford University Innovation \& London Institute for Healthcare Engineering, U.K}
\affil[2]{\normalsize\text{ }Department of Biomedical Computing, School of Biomedical Engineering and Imaging Sciences \& King's Institute for AI, King's College London, U.K}

\date{}
\begin{document}
\maketitle
\blfootnote{$^\dagger$These authors contributed equally to this work.}

\begin{abstract}
Whether large language models perform algorithmic inference or pattern
completion is hard to test, because most benchmarks supply answers but no
distributional reference for what the shown evidence licenses. F-ICL
supplies one exactly: we exhaustively enumerate the 86 million valid programs
of length at most 13 on a Turing-complete machine F, complement-symmetrised
to remove output-polarity bias, and compute the exact posterior under a
declared bounded Levin--Solomonoff prior. It is Bayes-optimal for that stated prior rather
than universal, and models are never told it exists, so the score reads the
inductive prior their served distribution already encodes. Across 105 serving
configurations spanning open models from 0.8B to 675B and frontier systems,
models answer up to 92\% of queries correctly, yet 45 of the 46 exposing
distributions sit farther from the F reference than a keystroke
reference. This is not an artefact of task selection: on the bit coordinate,
the half the length quota cannot distort, 69 of 80 runs stay below the anchor.
Fidelity is inert to scale, which accuracy tracks; continuation improves late
without converging; and models un-solve a solved task once per two gains, where
the F reference does so once per nine and always repairs it. Because
absolute distances are reference-dependent, we prove sequential bounds holding
for rival priors: any predictor whose prior gives the reference positive weight
has bounded cumulative excess loss, and, in a loss never invoking the
reference, any Bayesian mixture giving the realised truth positive mass has a
bounded truth-loss budget. On 23,998 trajectories, 86.7\% already spend over 10
bits of it. Sequences ending by position nine cannot exclude an arbitrarily
large finite constant, so these are lower bounds on what a rival prior must
already pay. F-ICL is an open benchmark and toolkit.
\end{abstract}

\section{Introduction}
In-context learning, the ability of a sequence model to infer a task from
a few input--output examples placed in its prompt and then generalise to a
new query, is among the most consequential capabilities of modern language
models\cite{brown2020gpt3}. It is also among the hardest to interpret. When
a model answers a few-shot prompt correctly, it is rarely clear whether it
has inferred the underlying rule or has retrieved a surface pattern that
happens to fit. The difficulty is not only empirical but definitional: on
most benchmarks there is no agreed distributional account of what the
examples license, so a model's behaviour can usually be compared only against
other models or against human answers, not against an explicit posterior.

A principled family of answers to ``what should be inferred from these
examples?'' exists in algorithmic information theory. Solomonoff's theory of
inductive inference defines priors over computable hypotheses and corresponding
Bayes predictors; for a declared prior no other predictor achieves lower
expected loss under that same generative law\cite{solomonoff1964,solomonoff1978,
li2019kolmogorov,hutter2005uai}. The obstacle to using this standard as a
yardstick is that it is uncomputable in general and
expensive even when bounded. Exhaustive enumeration of small machines has
long been used to estimate algorithmic complexity
numerically\cite{delahaye2012,solertoscano2014,zenil2018bdm}; what has
been missing is its use as an \emph{exact inferential standard} for
learning systems. As a result, the rich recent literature
connecting ICL to Bayesian inference\cite{xie2022bayesian,
mueller2022pfn,garg2022what,min2022rethinking,chan2022data} and language
modelling to compression\cite{deletang2024compression,mahoney1999,
huang2024compression} has had to rely on tractable but
non-universal hypothesis classes (e.g.\ linear functions or
parametric mixtures), trading the exactness of the \Fmac{} reference for analytic
convenience. Benchmarks of algorithmic reasoning face the same gap from
the other side: suites such as ARC\cite{chollet2019arc} and
CLRS\cite{velickovic2022clrs,velickovic2021nar} probe rule induction and
algorithm execution but score against the single intended answer, not
against what the evidence licenses; SuperARC\cite{superarc2026} instead
scores frontier models on human-agnostic, algorithmic-information terms, but
along an orthogonal axis to ours, grading compression-based model abstraction
and recursive prediction \emph{ability} against approximate complexity
estimates rather than grading a served \emph{distribution} against an exactly
enumerated posterior; meta-learners trained toward
Solomonoff induction\cite{grau2024universal} sample a closely related
Brainfuck-family machine to build a training stream, but sampling leaves the
target itself out of reach, so models there are compared against a loose
upper bound on the log-loss rather than against the posterior (we return to
this in the Discussion); and mechanistic
accounts of ICL\cite{olsson2022induction,akyurek2023what,
vonoswald2023transformers} describe \emph{how} models update in context,
but not how close that update is to a designated posterior.

\begin{figure}[!htbp]
  \centering
  \includegraphics[width=0.95\linewidth]{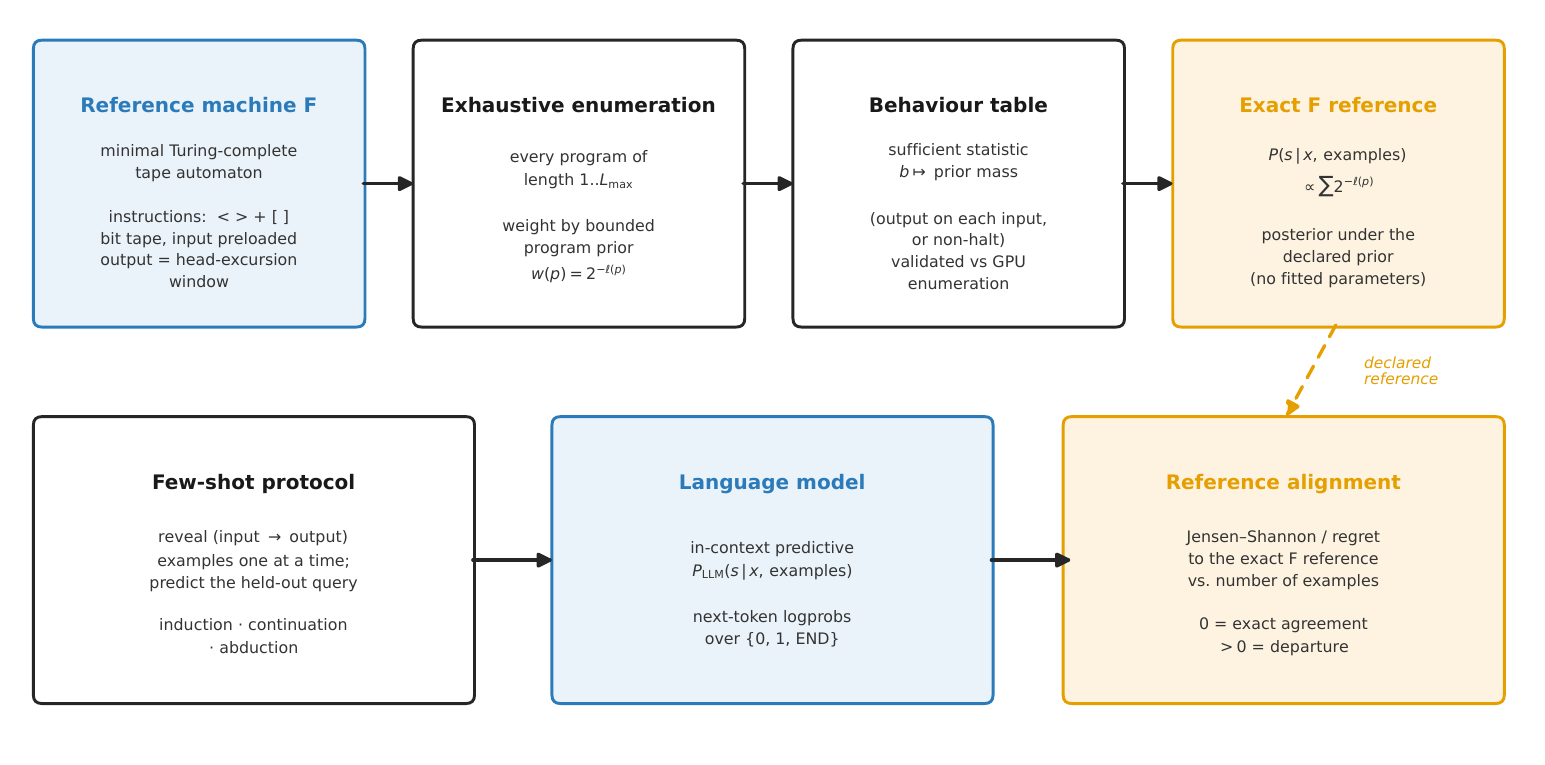}
  \caption{\textbf{Measuring in-context learning against an exact
  algorithmic reference.} Top: the reference machine \Fmac{} is enumerated
  exhaustively up to a length budget; weighting each program by the bounded
  program prior $w(p)=2^{-\ell(p)}$ gives a behaviour-to-mass table, a
  sufficient statistic from which its exact reference posterior follows in
  closed form (both defined in \S\ref{sec:standard}).
  An independent reference implementation cross-checks the accelerated
  enumeration. Bottom: in the few-shot protocol, examples are revealed one
  at a time and a language model's predictive distribution is compared
  against the \Fmac{} reference; the divergence to the \Fmac{} reference as a function of the
  number of examples is the measure of algorithmic reasoning ($0$ = matches
  the \Fmac{} reference exactly; larger values indicate greater departure from it).}
  \label{fig:principle}
\end{figure}

Here we make one algorithmic reference explicit and exact at a small but
genuinely algorithmic scale. We fix a minimal Turing-complete reference
machine, a five-instruction tape automaton we call \Fmac{}, and
\emph{exhaustively enumerate} every program up to a length budget,
weighting each by a bounded program prior. Because the enumeration is
exhaustive, the induced map from program behaviour to prior mass is a
\emph{sufficient statistic} for Bayesian inference, and the posterior
predictive for any few-shot context can be computed in closed form. This
yields, for each task, the exact posterior prediction under the declared
\Fmac{} prior, with no parameters tuned to the evaluated models
(Fig.~\ref{fig:principle}). We call this frozen target the \emph{\Fmac{}
reference}. It is Bayes-optimal for that bounded prior under its own
uncurated generative law. Optimality is therefore a property of the declared
prior and machine, and its scope is exactly that: it does not extend to ideal
Solomonoff induction, to the curated panel we evaluate on, or to what a text
model is obliged to hold. The comparison asks a correspondingly narrow
empirical question: how closely does the model's served predictive
distribution align with this specified algorithmic measure, and how does that
alignment change as evidence accumulates?

The epistemic condition is equally explicit. The standard prompt says that one
fixed program generated the shown input--output behaviour, but does not name
\Fmac{}, state the program prior or bounds, or disclose the length--complexity
selection rule. Consequently a gap from the \Fmac{} reference cannot be read as
irrationality: it combines the model's updating ability with its default prior
over what ``a fixed program'' means. The released prompt ablation varies
wording, template, ordering, output constraint and deliberation, but does
\emph{not} supply the complete \Fmac{} generator; a generator-disclosure
experiment remains a distinct, necessary control (\S\ref{sec:audit}).

Turning this idea into a usable benchmark requires solving two problems.
First, the enumeration and the \Fmac{} reference must be \emph{certified}: an
exact reference is only as trustworthy as the machine that defines
it. We address this with an accelerated GPU enumeration that scales to more
than a billion programs at the longest length, an independent pure-Python reference
implementation, and a chain of exact cross-checks: byte-for-byte agreement
of per-output program counts, re-execution of every stored witness program,
and bit-identical agreement between the served few-shot posterior and the
exhaustive joint. Second, the comparison must be \emph{discriminative and
fair}: it should reward genuine inference, expose pattern completion, and
summarise a model's behaviour in a way that is robust to task difficulty.
We instantiate three task families with exact \Fmac{} references; models are scored on
two (induction and continuation) by a bounded, symmetric divergence, and
the third (abduction) validates the design. Models are summarised by a
single difficulty-aware score alongside an interpretable facet profile.

We describe the benchmark, establish its validity, and use it to evaluate a
panel of open-weight and frontier language models. \Fmac{}-ICL separates its
reference update from strong heuristic foils, and reveals that current open
models track that update but fall well short of its sample efficiency: they solve the easiest
tasks while needing far more examples than necessary on harder ones, and
produce token-level distributions that are confidently mismatched
with the algorithmic reference. The benchmark is exact by construction,
inexpensive to run once the \Fmac{} references are frozen, and model-agnostic;
its interpretation is reference-relative rather than a claim of uniquely
rational prediction.

\section{Results}

\subsection{An exact reference for in-context inference}
\label{sec:standard}

This subsection defines the standard the model results are read against;
its certification is summarised in \S\ref{sec:certified} and the full
construction is given in Methods. \Fmac{} is a
minimal Turing-complete tape automaton in the Brainfuck family
\cite{bohm1964}: five
instructions acting on a bit tape preloaded with a binary input
(Table~\ref{tab:instructions}). It is small enough to enumerate
exhaustively up to a length and runtime budget,
yet expressive enough that its short programs realise
counting, arithmetic and bounded recursion. Weighting each program by a
bounded Levin--Solomonoff-style prior makes the corresponding reference
inference computable: for any few-shot context the posterior predictive
under that declared prior is computed
\emph{exactly}, in closed form, with nothing tuned to the models
(Eq.~\eqref{eq:posterior}). That \Fmac{} reference is relative to a fixed set of
conventions, stated once here and tabulated in
Table~\ref{tab:conventions}: the symmetrised machine \sFmac{}, programs of
length $L\le13$, a $1024$-step and $128$-cell run budget, normalisation over
halting outputs, and a per-family weighting that is length-only for
induction and Levin-style (length plus an Elias-$\delta$ halt-time code) for
the accumulating-evidence families.

Enumerating every program on a fixed ordered set of inputs and recording
each program's \emph{behaviour} gives $w(b)$, the total prior mass of each
distinct behaviour under the length prior $w(p)=2^{-\ell(|p|)}$
(Eq.~\eqref{eq:table}, Methods), a sufficient statistic for
Bayesian inference over the \Fmac{}-prior restricted to those inputs.
Given examples $\mathcal{E}=\{(x_i,o_i)\}$, a behaviour is
\emph{consistent} if $b(x_i)=o_i$ for all $i$, and the exact posterior
predictive for a query input $x_q$ is
\begin{equation}
  P\!\left(s \mid x_q,\mathcal{E}\right)=
  \frac{\sum_{b\,\text{consistent},\,b(x_q)=s} w(b)}
       {\sum_{b\,\text{consistent}} w(b)} .
  \label{eq:posterior}
\end{equation}
When tasks are drawn from the bounded \Fmac{}-prior, no predictor attains
lower expected log-loss than Eq.~\eqref{eq:posterior}. The evaluated
sample is \emph{curated} from that prior (quotas over output length and
complexity, plus the learnability gates of Methods), so
Eq.~\eqref{eq:posterior} is Bayes-optimal for the prior and not for the
curated evaluation law. \S\ref{sec:audit} measures the consequences of
that gap directly, and reports the headline under corrections for it.
Appendix~\ref{app:priorfree} separates what follows from a change of prior:
on a genuine sequential reveal, every coherent predictor that finitely
covers the reference has a fixed pathwise regret budget, while every
deterministic Bayesian mixture assigning positive mass to the truth has a
fixed cumulative truth-loss budget. These statements motivate extensible
continuation and prequential-induction audits; they do not turn the present
finite held-out induction curves into a proof against every finite prior.

\Fmac{}'s all-zero tape makes zeros cheap, so the raw prior prefers
zero-heavy outputs (first-bit law $P(\code1\mid\varepsilon)=0.278$; mean
conditional complement asymmetry $0.24$ bits Jensen--Shannon), a
confound a frozen benchmark would reward. To keep the standard free of that
bias, all \Fmac{} references are served from the
complement-symmetrised machine \sFmac{}, and every task ships with its
bitwise-complemented \emph{twin}, on which the \Fmac{} reference provably scores
identically (Methods~\ref{sec:methods-sf}). The enumeration, the stored \Fmac{} references and
the serving path (an exhaustive small-program joint over programs
$L\!\le\!6$, and a full-budget $L\le13$ program index; two backends that
cross-check each other) are certified by exact, independent checks
(\S\ref{sec:certified}; Methods).

Models are scored on two task families. In \emph{induction}, a hidden
program is shown through input$\to$output examples and the model predicts
its output on a held-out query; the target is the exact posterior over
outcomes. In \emph{continuation}, a program's output is revealed one token
at a time and the model gives its next-token distribution over
$\{\code0,\code1,\textsc{end}\}$; the target is the exact conditional at
every position. A third family, \emph{abduction}, validates the design (worked example in
Appendix~\ref{app:abd});
\S\ref{sec:foils}. The headline \Fmac{}-ICL-A score anchors the
Jensen--Shannon divergence between model and \Fmac{} reference to three exactly
computable points (Eq.~\eqref{eq:skill}, Methods): the worst possible
predictor scores $0$, the keystroke reference (a uniform random keystroke over $\{\code0,\code1,\textsc{end}\}$ at every step) scores exactly
$\tfrac12$, and the \Fmac{} reference scores $1$. The keystroke reference is a
fixed, computable, model-independent point, but not an uninformed one: it is
itself an algorithmic mixture, the one induced by a print-only machine with
no loops (\S\ref{sec:audit}). A score's position relative to $\tfrac12$
therefore locates a model between that loop-free mixture and the
loop-bearing \Fmac{} reference, and its distance states how much of the corresponding
gap it closes. We call
the quantity this score measures \emph{reference fidelity}: how closely its
served predictive distribution matches the exact \Fmac{} reference
posterior. We avoid the term ``calibration'', which in the machine-learning
literature denotes confidence--accuracy agreement\cite{guo2017calibration};
reference fidelity is the distribution-level property of matching that
declared posterior. Raw Jensen--Shannon geometry compresses all
predictors toward the top of its scale (the keystroke reference scores
${\approx}0.85$), which is why the headline is the anchored score rather
than the raw one; the raw per-family divergences and the keystroke
references are reported alongside it and never blended. Two robustness
summaries accompany it: a difficulty-balanced score weighting easy and hard
tasks equally, and a median \emph{sample-complexity efficiency} (reference
examples-to-solve divided by model examples-to-solve; $1$ matches the
reference). Every
induction task additionally carries its \emph{Bayes sample complexity}
($\mathrm{BSC}$), the number of examples the \Fmac{} reference itself
needs to identify the answer, including zero when the prior alone
identifies it; the frozen set is balanced across
$\mathrm{BSC}\in\{0,1,2,3{+}\}$. One consequence of scoring against an
exact posterior should be stated here: the induction candidate set contracts
as evidence eliminates programs, and collapses to a single candidate in
$63\%$ of (task, shot) cells overall and $93\%$ by the final shot, where the
divergence is identically zero for model, keystroke reference and worst-case
predictor alike. The anchors therefore dilute together and the anchored
ratio is preserved, but the induction divergence facet carries its signal on
the roughly one third of cells where the posterior has not yet collapsed.
Reference foils and classical nulls are
scored through the identical pipeline, so the region around and below the
keystroke reference is itself charted (\S\ref{sec:foils}; Methods). With these
anchors in hand we first establish, on data drawn from the declared mixture,
that distance to the \Fmac{} reference separates reference-consistent Bayesian
updating from pattern completion, then report the panel on
that established axis. The construction of the instrument is given in
Methods.

\begin{table}[t]
\centering
\caption{\textbf{The reference machine \Fmac{}.} Five instructions over a
bit tape; $\code{[}\,\dots\,\code{]}$ is a while-the-current-bit-is-1 loop.
The only validity test is bracket matching. Input is preloaded on the tape;
output is the bit string on the head-excursion window at halt.}
\label{tab:instructions}
\small
\begin{tabular}{@{}cll@{}}
\toprule
Symbol & Name & Effect \\
\midrule
\code{<} & left  & move head one cell left (abort if off-tape) \\
\code{>} & right & move head one cell right (abort if off-tape) \\
\code{+} & flip  & flip the bit under the head \\
\code{[} & loop-begin & if current bit $=0$, jump past the matching \code{]} \\
\code{]} & loop-end   & if current bit $=1$, jump back past the matching \code{[} \\
\bottomrule
\end{tabular}
\end{table}

\subsection{The standard is certified}
\label{sec:certified}

A benchmark whose targets are computed by a bespoke accelerator invites the
worry that the accelerator, not the model, is being measured. A chain of
exact checks rules this out (Methods):
\begin{itemize}
  \item \textbf{Gate 1: count cross-check.} An independent,
  dependency-free pure-Python \Fmac{} reproduces the GPU enumeration's
  per-(input, output) program counts \emph{byte-for-byte} ($L=5$, $M=8$,
  all $511$ inputs of length $0..M$ for $M=8$, $M$ denoting the maximum input
  length of the enumerated input set).
  \item \textbf{Gate 2: witness re-execution.} Sampled stored entries
  decode their \emph{witness} program and re-execute it with exact CPU
  semantics, confirming halt time and output: every recorded behaviour is
  a real, re-runnable program.
  \item \textbf{Gate 3: rule-set identity.} The dumped program index
  serving the multi-example \Fmac{} references is re-enumerated and checked, per
  (input, output), for identity of the exact \emph{set} of program
  identities, not just their count.
  \item \textbf{Serving check.} The posterior served from that index is
  \emph{bit-identical} to the exhaustive small-program joint across
  $0$--$5$ examples, including non-halting examples
  (Appendix~\ref{app:index}).
\end{itemize}
A build is accepted only on exact agreement. Two metric-compatible backends
compute the \Fmac{} reference and cross-check each other (a small-program exact
joint, and the production dumped program index serving the complete $L\le13$
budget; Methods), and a pre-release audit recomputed the frozen datasets'
stored targets from independent code paths, catching two builder defects
invisible to every accuracy-style metric. The build therefore asserts
twin-mirror equality, normalisation and the exact first-position law on
every dataset it writes.

\subsection{The benchmark separates inference from pattern completion}
\label{sec:foils}

\begin{figure}[!htbp]
  \centering
  \includegraphics[width=\linewidth]{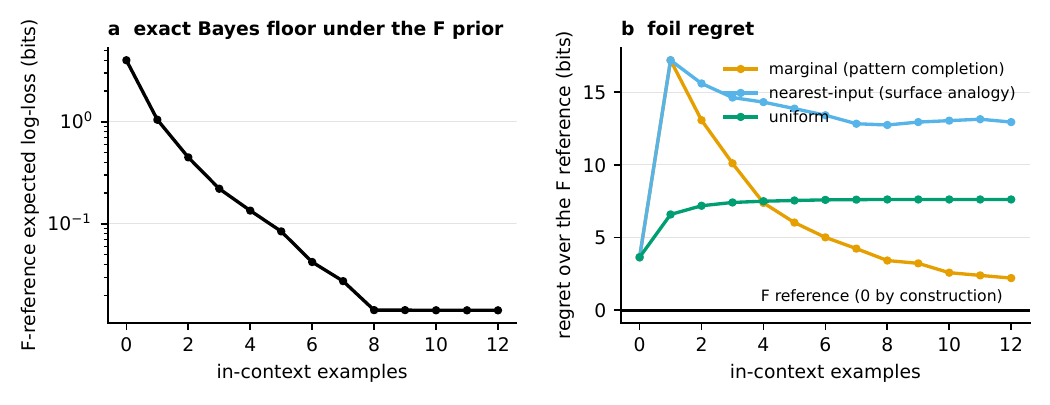}
  \caption{\textbf{The benchmark separates reference Bayesian inference from
  pattern-matching foils} (realisable \sFmac{} family, $M=4$, $L_{\max}=6$,
  $300$ tasks; sF \Fmac{} references under the mass--time weighting of
  Eq.~\eqref{eq:masst}, the same joint that serves the pipeline-scale build induction
  \Fmac{} references).
  \textbf{a}, The exact Bayes floor under the declared mixture (expected
  log-loss, log scale)
  falls from $3.99$ bits to $\approx0.01$ bits as examples accumulate; no
  predictor can lie below this curve. \textbf{b}, Regret (excess bits over
  the \Fmac{} reference) versus number of in-context examples. The \Fmac{} reference
  is at zero by construction; the marginal (output-frequency) foil spikes to
  $17.2$ bits at one example by confidently committing to an unsupported
  output and still pays $2.2$ bits at twelve; the uniform foil's regret
  climbs to $\log_2|\mathcal O|\approx7.6$ bits as the floor falls; the
  nearest-input (surface-analogy) foil remains near $13$ bits.
  None approaches the \Fmac{} reference.}
  \label{fig:foils}
\end{figure}

We verify that approaching the \Fmac{} reference is genuinely diagnostic on
data generated by its own declared mixture by scoring predictors ranging from
the reference Bayesian rule to purely superficial heuristics
(Fig.~\ref{fig:foils}), on a realisable few-shot family drawn from the
\sFmac{} prior ($M=4$, $L_{\max}=6$, the same small-program joint that
cross-checks the induction \Fmac{} references; $683$ distinct behaviours over $199$
outcomes). The exact Bayes floor falls from $3.99$ bits with no examples
to $\approx\!0.01$ bits by eight (Fig.~\ref{fig:foils}a); the \Fmac{}
reference sits at zero regret by construction. The foils never approach it
(Fig.~\ref{fig:foils}b): the \emph{marginal} predictor (output
frequencies with the inputs ignored, the natural formalisation of pattern
completion) still pays $\approx\!2.2$ bits of excess loss at twelve
examples and spikes to a $17.2$-bit regret at one example by confidently
committing to an output the algorithmic map does not support; the
\emph{nearest-input} predictor (surface analogy) stays $\approx\!13$ bits
above the \Fmac{} reference; the \emph{uniform} predictor pays a constant
$\log_2|\mathcal O|\approx 7.6$ bits. The separation is large and stable.

\begin{figure}[!htbp]
  \centering
  \includegraphics[width=0.72\linewidth]{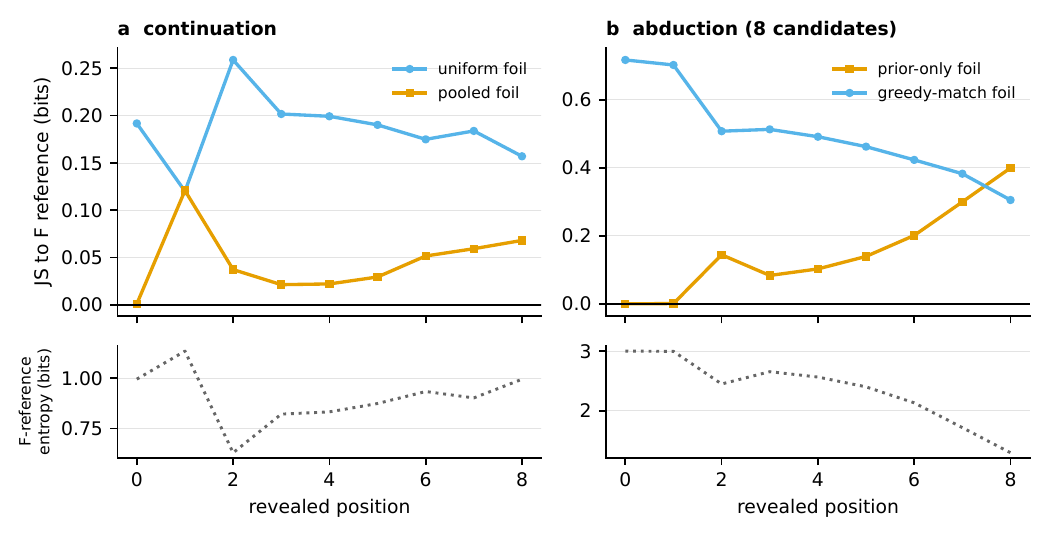}
  \caption{\textbf{Accumulating-evidence task families at scale}
  ($L\!\le\!13$, $M\!\le\!8$, $500$ length-balanced tasks each under the
  \sFmac{} mixture; output revealed token by
  token). \textbf{a}, Continuation: Jensen--Shannon divergence of the
  uniform and input-agnostic ``pooled'' foils to the \Fmac{} next-token
  distribution (upper panel); the \Fmac{} next-token entropy (lower panel,
  dotted) is non-monotonic and peaks near $1.1$ bits. Continuation is a
  sensitive fidelity probe but only weakly separates these foils.
  \textbf{b}, Abduction over eight candidate inputs: the \Fmac{} posterior
  entropy (lower panel, dotted) collapses from $3$ bits toward $\approx\!1.3$
  bits as evidence accumulates; the prior-only foil drifts \emph{away} from
  the \Fmac{} reference (it ignores evidence while the \Fmac{} reference sharpens) and the
  greedy-match foil improves but never reaches the \Fmac{} reference (upper panel).}
  \label{fig:families}
\end{figure}

At scale, the production backend ($L\!\le\!13$, $M\!\le\!8$; $500$
length-balanced tasks per family, Methods) instantiates two
accumulating-evidence families whose \Fmac{} references sharpen as evidence is
revealed token by token (Fig.~\ref{fig:families}): \emph{continuation}
(next-token prediction over $\{\code{0},\code{1},\textsc{end}\}$ against
the exact Bayesian continuation) and \emph{abduction} (inferring which of
eight candidate inputs produced the revealed output; the exact posterior's
entropy collapses from $3$ toward $\approx\!1.3$ bits). Their foil
separation differs informatively. In continuation, even an input-agnostic
``pooled'' predictor stays within $0.12$ bits (Jensen--Shannon) of the
\Fmac{} reference (typically below $0.07$) and a uniform next-token guess within
$\approx\!0.26$ bits (Fig.~\ref{fig:families}a); the family is a sensitive
fidelity probe but a weak foil separator. Abduction is far
more discriminative (Fig.~\ref{fig:families}b): a \emph{prior-only}
predictor drifts \emph{away} from the sharpening \Fmac{} reference (to
$\approx\!0.40$ bits) and a \emph{greedy-match} heuristic improves (from
$0.72$ to $\approx\!0.31$ bits) yet never reaches it. Continuation tests
distributional fidelity; abduction, Bayesian integration of
evidence.

In the panel of \S\ref{sec:eval-panel}, abduction accordingly serves as a
\emph{validation} family: models are scored on induction and continuation,
whose metrics read directly off token log-probabilities; scoring abduction
requires the model's distribution over a candidate \emph{input} set, which
current serving interfaces do not expose cleanly (Discussion).

One scope condition should be carried forward. The separation demonstrated
here is established on \emph{realisable} families drawn directly from the
\sFmac{} prior, whereas the panel is scored on curated task sets selected
for learnability and length balance (Methods), for which the \Fmac{} reference is not
the best predictor of the realised sample (\S\ref{sec:audit}). We take the
foil separation as evidence that distance to the \Fmac{} reference can diagnose
updating relative to that declared mixture, not as a measurement transferred
cell by cell to the curated sets or as a test of rationality under an
undisclosed prior; the classical nulls of \S\ref{sec:eval-panel},
which are scored on the curated sets themselves through the identical
pipeline, are what bracket the panel there.

\subsection{Evaluating a panel of language models}\label{sec:eval-panel}

We evaluated \textbf{105} model runs\footnote{Missing calls are dropped from
the affected means, never scored. One API model
(\code{mistral-large-3-675b}) was retired by the provider mid-evaluation, its
continuation phase permanently unservable (HTTP 410, end-of-life); its row is
reported from the completed induction family only. The episode is itself a
datum for the reproducibility discussion: benchmark rows measured through
commercial APIs can become unrepeatable overnight.}: an API panel of
\textbf{49} rows spanning NVIDIA NIM ($30$ open models), OpenAI (a dated
generational ladder from GPT-4o May 2024 to GPT-5.6, nine models), Google
(Gemini 2.5/3.x through Vertex and the developer API, plus gemma-4 through
the developer API), xAI (Grok 4.20 non-reasoning and Grok 4.5 at low
reasoning effort) and Anthropic (Claude Fable~5, Opus~4.8,
and the Fable-with-Opus-fallback stack), and
a local panel of
\textbf{56} \code{llama.cpp} GGUF builds spanning quantisation ladders, a
parameter-scale ladder, and base/instruction/reasoning triads (Methods;
serving routes per provider are documented there).

Counts below refer to four nested populations, named here once and used
consistently thereafter. The \emph{panel} is all $105$ runs. The
\emph{log-probability runs} are the $81$ of these, over $46$ distinct
models, whose endpoint exposes next-token distributions; only these carry a
fidelity score. The \emph{macro-scored runs} are the $80$ of those $81$ that
completed both scored families, the exception being the retired
\code{mistral-large-3-675b} row, which has induction only and is reported
separately wherever a macro score is required. Answer-level statistics
(accuracy, agreement, alignment) use the larger set of runs that returned
answers, which does not require log-probabilities and is stated at each use.
The headline below-anchor count is over the log-probability runs ($78$ of
$81$); the robustness re-checks of \S\ref{sec:audit}, which require a macro
score, report the macro-scored count ($77$ of $80$). Continuation analyses quote two
cell counts, which differ by bookkeeping granularity rather than by dataset
or route: the frozen family contains $1{,}950$ (task, position) cells (one
per output bit plus the terminator, over $300$ tasks of output length
$3$--$8$), but a cell's prompt and \Fmac{} reference depend only on its (input,
prefix) pair, and because tasks share inputs ($204$ distinct inputs across
the $300$ tasks) and overlapping output prefixes, $1{,}736$ of those cells
are distinct measurements; the harness issues each distinct prompt once and
reuses the result. Task-level scoring iterates all $1{,}950$ cells, while
analyses that operate on distinct measurements (the read-out audit and the
tokenisation check) quote $1{,}736$. The
model-level statement in the abstract counts distinct models rather than
runs.

The
panel was scored against the \emph{pipeline-scale build} of the
benchmark: the continuation family at the full $L\le13$ budget (identical, by
construction, to the extended build) and an induction family served from the
exact small-program joint (hidden programs $L\le6$, mass--time weighting);
an extended build, released alongside it and used for the validation
figures, scales induction to hidden programs of $L=5$--$9$ under the
complete $L\le13$ mixture with length-only weights served by program-set
intersection (Methods, Reproducibility). The two instruments are equivalent on these
tasks, and for a principled reason: as program length grows, the bounded
mixture thins (ever fewer programs realise new behaviour per unit of
prior mass, the same collapse that degrades the correlation between
mixture complexity and minimal program length in our companion
analysis\cite{fcomplexity2026}),
so short-program support carries the posterior either way. We verified the
equivalence exactly on the panel's own tasks by re-serving all $120$ induction tasks
from the complete $L\le13$ store under the frozen-release conventions
using the released robustness tool: across all $1{,}080$
(task, shot) cells the two \Fmac{} references differ by a mean Jensen--Shannon
divergence of $0.028$ bits with $98\%$ maximum-a-posteriori (MAP) agreement, and the difference
decomposes into a weighting-convention component ($0.027$ bits) and a
budget component of only $0.002$ bits, confirming that programs beyond
$L=6$ contribute negligible mass on these tasks. Models are tested
equivalently: rescoring every stored model distribution against the
swapped \Fmac{} references on identical candidate sets reproduces the per-run
\Fmac{}-ICL-A(induction) with rank correlation $0.975$ (mean absolute
change $0.015$, $n=81$ runs), and the headline count is unchanged, $77$ of
$80$ macro-scored runs below the keystroke reference under either instrument (one of
the $81$ log-probability runs, the retired mistral row, has no
continuation leg and so no macro score). What this equivalence does not
cover is the extended build's harder task regime (hidden programs of
length $7$--$9$, beyond the pipeline-scale build's $L\le6$), on which no
model has yet been scored; that regime is out of scope here.
Because the \Fmac{} reference is fixed, every run is measured against the same target, and
the same model served two ways can be compared directly.

The pipeline's missing-data guards proved load-bearing here: an initial local serving of the
gemma-4 family returned (near-)uniform log-probability read-outs (divergence
equal to the keystroke reference to within $2\times10^{-4}$), which the degeneracy
audit flagged as a serving artefact (a chat-template/token-mapping fault in the
serving stack) rather than a result. After the template fix the family was
re-evaluated cleanly and re-entered all analyses automatically; its true
continuation divergence places it among the families \emph{farthest} from the \Fmac{} reference posterior
in the panel, where the artefactual read-out would have placed one gemma
variant at the keystroke reference.

The audit has an answer-side twin: six API rows
returned empty replies on the majority of answer calls ($53$--$99.9\%$;
broken template or truncation at the serving layer, not models scoring
zero), and their accuracy facets are excluded with rates disclosed
under the released answer audit. Where a provider delivers no
answer, we report no answer.

A third audit closes the remaining gap on the log-probability side. The uniform test above detects read-outs that
collapse toward the anchor, but not the opposite failure, in which the
served distribution collapses onto a single class because the others never
appeared in the endpoint's top-$k$ and were filled by the smoothing floor.
Its signature is exact: two of the three classes carrying bitwise
identical probability, which cannot happen when a model genuinely resolves
them. Four locally served rows failed this test on a majority of cells
under the released read-out audit; two of them, both base
checkpoints, failed on $100\%$ of cells and consequently scored
\emph{numerically identically} to one another despite different
architectures, sizes and tokenisers: the instrument was reading its own
floor back rather than resolving the model. The cause is a harvesting
limit rather than a property of the models, so it is repairable: we
re-served nineteen local rows (the four failures and every row sharing
their serving configuration) requesting a larger top-$k$, after which all
nineteen resolve all three classes on all $1{,}736$ cells and the two base
checkpoints separate. The repair moves fourteen of the nineteen rows by
less than $0.01$ bits; the largest move is Qwen3-30B-A3B-Instruct, whose continuation
divergence rises by $0.096$ bits and whose \Fmac{}-ICL-A falls from
$0.343$ to $0.308$, which was $94\%$
tie-degenerate and is the middle rung of one of the two post-training
triads, so the repair \emph{enlarges} that triad's base-to-instruct drop
rather than creating it. The repaired read-outs are what the reported
panel uses; the originals are retained in the release as the evidence for
this audit. Two caveats follow from repairing rather than discarding.
The re-serve was scoped to the affected serving configuration, so
nineteen of the fifty-six local rows were harvested at the larger
top-$k$ and thirty-seven at the original depth; the base-to-instruct
comparisons above are configuration-matched, but the third triad rung and
part of the gpt-oss quantisation ladder are not, and harvesting depth is
recorded per row in the registry. And the threshold is a threshold, not a
guarantee: after the repair every local row resolves three classes on
every cell, while every API row still shows some epsilon ties, fourteen of the
twenty-four on $5\%$ or more of cells and none above $17.2\%$, which we cannot repair because the
providers cap the harvest. That residue flatters the affected rows
slightly (across API rows the tie fraction is negatively rank-correlated
with divergence), so it is conservative for the below-anchor headline but
not for orderings: within the OpenAI ladder the oldest and most faithful
row carries $10.9\%$ ties against $0.3\%$ for the next, and the ladder
claim should be read with that asymmetry in mind. Two lessons generalise: a degeneracy guard written for one
failure mode (near-uniform) is blind to its mirror image (near-point-mass),
and two rows scoring identically to six significant figures is a
diagnostic worth testing for automatically.

\begin{figure}[!htbp]
  \centering
  \includegraphics[width=0.85\linewidth]{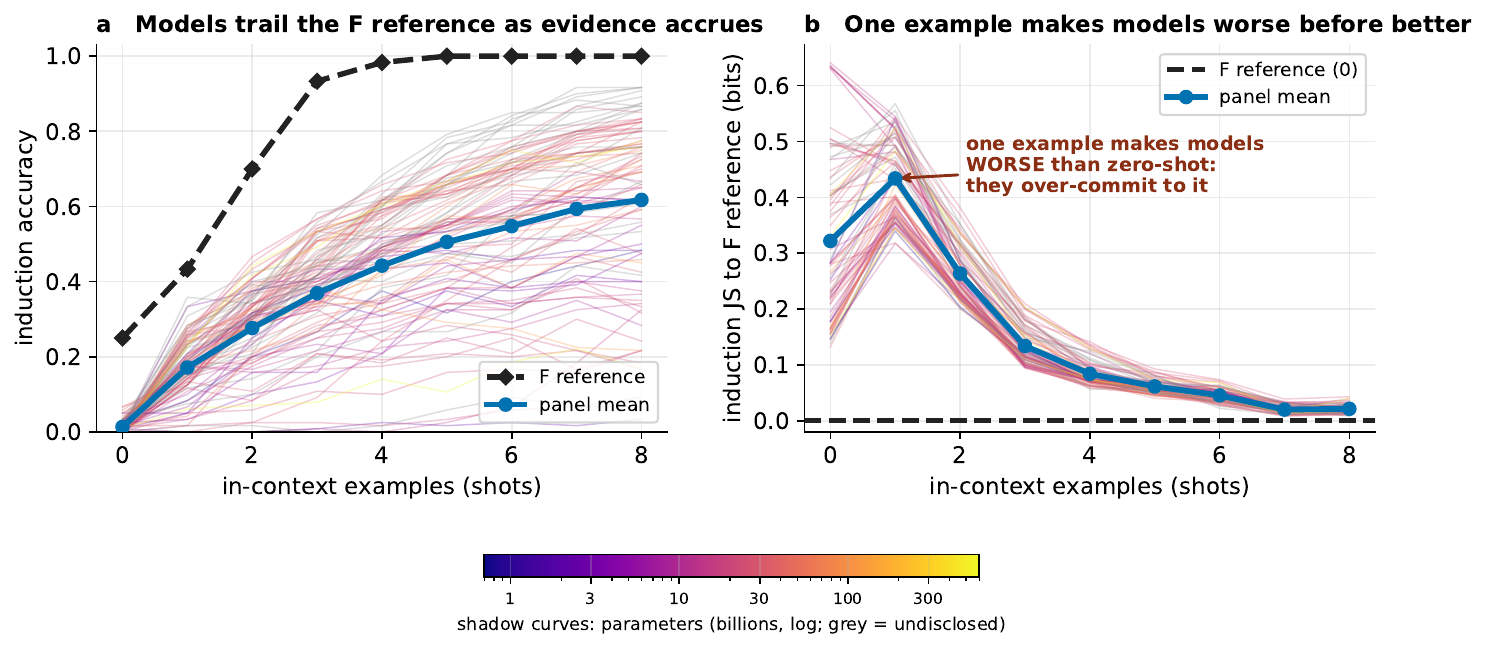}
  \caption{\textbf{Approach to the \Fmac{} reference.} Faint lines are
  individual models, coloured by parameter count (log scale, as in
  Fig.~\ref{fig:gap}; grey = undisclosed API size); the absence of a colour
  gradient previews the scale-invariance result. The bold line is the panel
  mean. \textbf{a}, Induction
  accuracy versus in-context examples: the \Fmac{} reference (black) reaches $1.0$ by five
  examples while the panel plateaus near $0.62$. \textbf{b}, Posterior Jensen--Shannon divergence to the \Fmac{} reference:
  one example \emph{increases} the mean divergence (models over-commit to a
  single datum) before it slowly declines, never matching the \Fmac{} reference's zero.}
  \label{fig:approach}
\end{figure}

\paragraph{The measurement, made concrete.}
The benchmark asks how fast a model's in-context inference approaches the
\Fmac{} reference as examples accumulate (Fig.~\ref{fig:approach}). The
\Fmac{} reference locks onto the truth quickly: its induction accuracy is $0.98$ by four
examples and exact by five. The panel, in contrast, rises slowly and plateaus
far short (mean accuracy $0.62$ at eight examples; Fig.~\ref{fig:approach}a),
leaving a large and persistent shortfall. The posterior-divergence view is even
more telling (Fig.~\ref{fig:approach}b): a \emph{single} example moves the
panel mean \emph{away} from the \Fmac{} reference, as models over-commit to one datum
before slowly recovering, exactly the failure the pattern-matching foils
exhibit (\S\ref{sec:foils}). This one-example over-commitment is nearly
universal: $69$ of the $81$ runs with usable log-probabilities, frontier
rows included, have a larger divergence after one example than after none
(sign test $p<10^{-9}$). The
shortfall is \emph{not}, however, an over-commitment to simplicity: panel
agreement with the Bayes-MAP answer tracks truth-accuracy to within $\pm0.02$
at every shot, so models favour neither the Occam answer nor the truth; they
are simply far from both. The remaining paragraphs quantify where this gap sits
and what does (and does not) move it.

\begin{figure}[!htbp]
  \centering
  \includegraphics[width=0.8\linewidth]{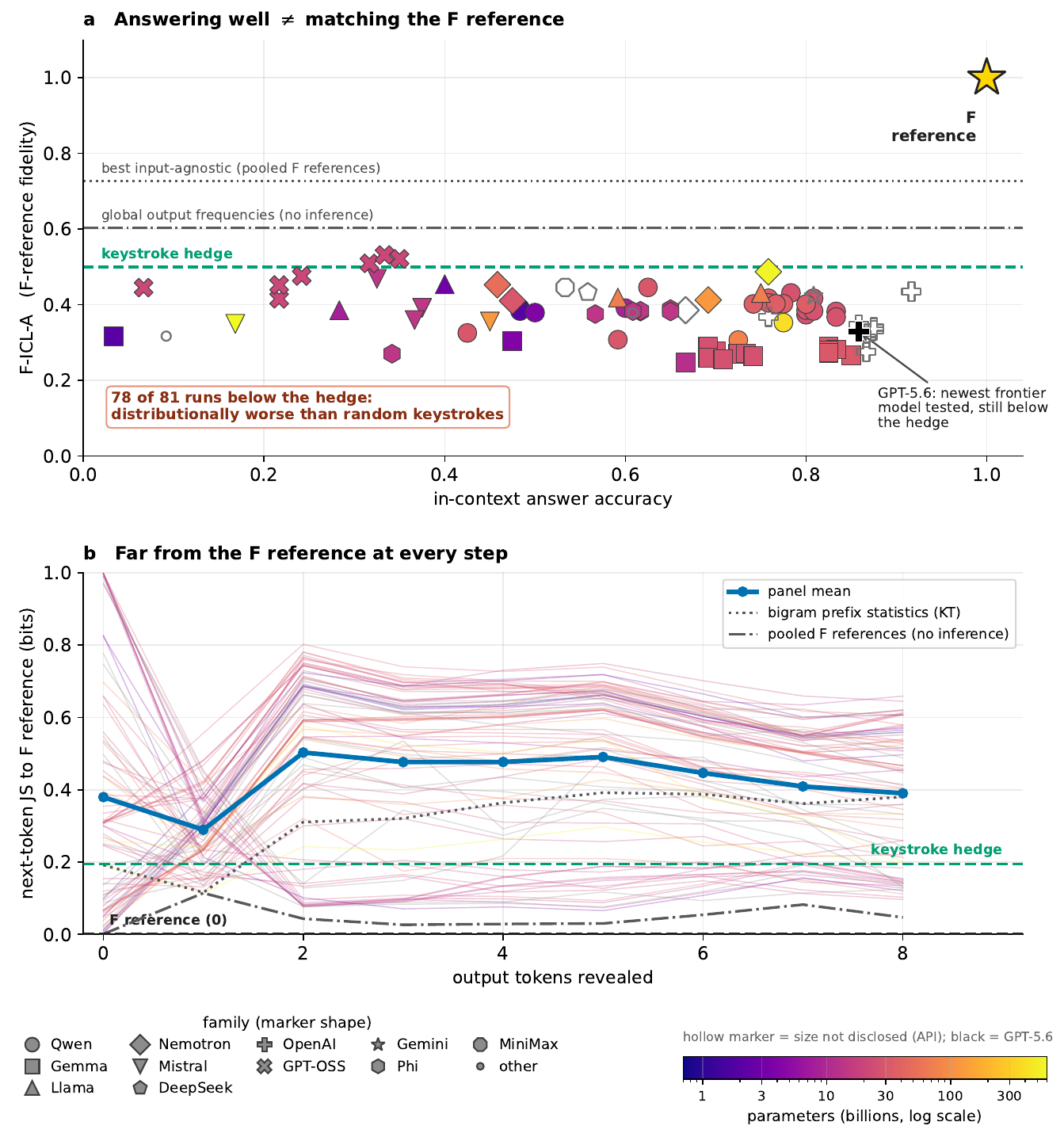}
  \caption{\textbf{The algorithmic-reasoning gap.}
  \textbf{a}, Every model's in-context answer accuracy (final shot) against its
  anchored \Fmac{}-ICL-A posterior fidelity to the algorithmic \Fmac{} reference; marker shape
  encodes the model family and colour the parameter count (log scale), with
  hollow markers for undisclosed API sizes. The dashed line is the keystroke reference and the star the \Fmac{} reference; grey reference lines are the stronger
  classical nulls (Methods): the dataset's global output frequencies ($0.60$)
  and the pooled per-shot \Fmac{} references ($0.73$), both above every model in the
  panel. Models reach up to $92\%$ accuracy yet
  cluster below the keystroke reference ($78$ of $81$ runs; two-sided sign test
  $p<10^{-9}$): answering well does not imply reasoning like the \Fmac{} reference. \textbf{b}, Faint lines are individual models
  (coloured by parameter count as in (a); grey = undisclosed API size) and
  the bold line the panel mean ($n=80$ runs with usable continuation) of the continuation
  Jensen--Shannon divergence to the \Fmac{} reference, which exceeds the keystroke reference ($0.19$ bits) at every one of
  the nine revealed positions and never reaches the \Fmac{} reference's zero; the dotted
  grey curve is the order-1 Krichevsky--Trofimov prefix-statistics null, which
  tracks the panel band, and the dash-dotted curve the pooled \Fmac{} references.}
  \label{fig:gap}
\end{figure}

\begin{figure}[!htbp]
  \centering
  \includegraphics[width=0.95\linewidth]{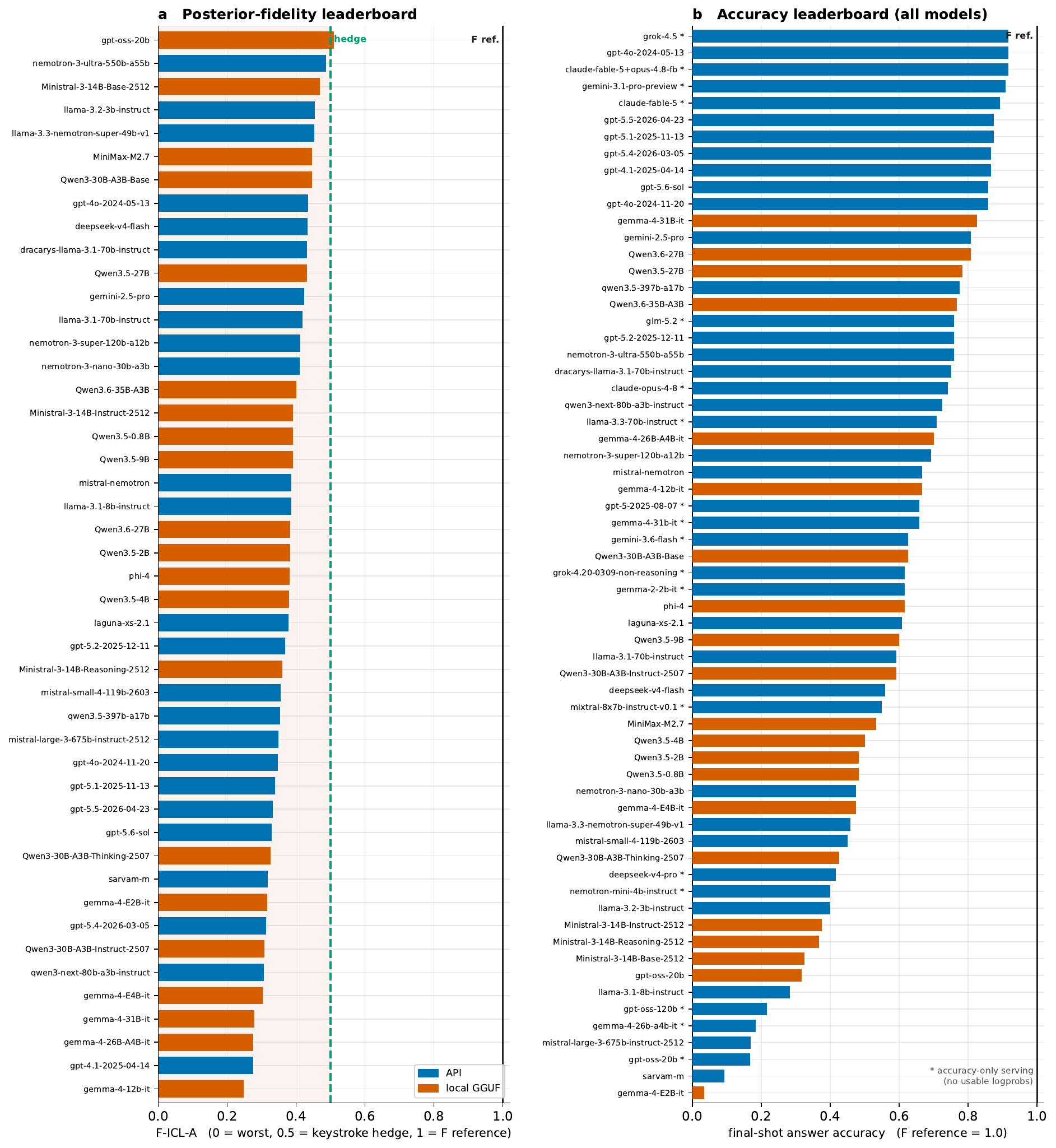}
  \caption{\textbf{Panel leaderboards.} Best-scoring precision per model
  ($n=46$ models with usable log-probabilities in (a), $64$ models with
  non-degenerate answer serving in (b)); blue = API, orange = local GGUF. \textbf{a}, The anchored \Fmac{}-ICL-A scale (models
  with usable log-probabilities). The dashed line is the keystroke reference and
  the shaded band the sub-anchor region; the solid line is the Bayes
  \Fmac{} reference. All models but one score below the keystroke reference; the exception,
  locally served \code{gpt-oss-20b} (plotted at its best-scoring
  precision, \textsc{q6}, $0.53$), is the only bar to cross it. \textbf{b}, Final-shot
  answer accuracy for \emph{every} model, including the accuracy-only API
  rows (marked $*$) whose serving exposes no usable log-probabilities and
  which therefore cannot appear in (a); the fidelity and accuracy
  orderings are visibly unrelated.}
  \label{fig:leaderboard}
\end{figure}

\paragraph{Models track accuracy but are confidently far from the posterior.}
The headline is a dissociation the exact \Fmac{} reference makes visible and accuracy
alone conceals (Fig.~\ref{fig:gap}). Of the \textbf{81} runs of \textbf{46}
distinct models with usable log-probabilities, $78$ runs ($45$ of the $46$
models) score \emph{below} the keystroke reference on the
anchored \Fmac{}-ICL-A scale, i.e.\ distributionally further from the posterior than the
keystroke reference on the algorithmic-reasoning axis. The single exception is locally served
\code{gpt-oss-20b}, which sits \emph{at} the keystroke reference (full-precision $0.51$,
task-bootstrap $95\%$ CI $[0.50,0.52]$; its
best single run, at \textsc{q6}, reaches $0.53$)
(Fig.~\ref{fig:gap}a; the full named ranking is
Fig.~\ref{fig:leaderboard}). The frontier is no exception: every proprietary full row, from OpenAI's
GPT-4o through GPT-5.6 and Google's Gemini 2.5 Pro ($0.42$), lands below
the anchor, and the current flagship \code{gpt-5.6-sol} pairs near-ceiling
answer accuracy ($0.86$) with an \Fmac{}-ICL-A of $0.33$, below the panel
median: whatever frontier training improves, it is not fidelity to the
algorithmic posterior. Point-prediction accuracy and posterior
fidelity are also uncorrelated (base-model Spearman
$\rho=-0.19$, $p=0.21$), so models answering up to $92\%$ of
in-context queries correctly still sit in the sub-reference zone, far from the Bayes
\Fmac{} reference. The gap is also an efficiency statement: the panel's median
sample-complexity efficiency is $0.40$, i.e.\ the typical model consumes about
two and a half times the information-theoretically minimal number of examples on the
tasks it
does solve. The continuation family localises the failure
(Fig.~\ref{fig:gap}b): the panel-mean next-token divergence exceeds the keystroke reference
($0.19$ bits) at \emph{every one} of the nine revealed positions
($0.29$--$0.50$ bits). The models are confidently wrong about the algorithmic
continuation, not just uncertain about it. The result does not depend on the
part of the measurement the sampling design can distort. Continuation
divergence splits into a termination coordinate (\textsc{end} versus not) and a
bit coordinate ($\{\code0,\code1\}$ renormalised), and the length quota bites
almost entirely on the first: because the shortest sampled output has length
three, a string cannot end at positions~1 or~2, yet the \Fmac{} reference still
places mean \textsc{end} mass there. Scoring the bit coordinate alone, the half
the quota cannot reach, still leaves $69$ of $80$ runs below the corresponding
anchor ($0.121$ against $0.057$ bits), with the panel's sole at-anchor
exception falling below it (\S\ref{sec:audit}). What the two coordinates do not
share is the ranking, so every ordering result below is reported as a
termination-coordinate finding; it is the \emph{level} that survives on the
bits, and the level is what the sampling objection concerns. This shortfall is
difficulty-wide, not a hard-task tail: panel-mean induction accuracy is $0.69$
even on tasks the \Fmac{} reference solves from the \emph{prior alone}
($\mathrm{BSC}{=}0$), drops to $0.53$ on tasks the \Fmac{} reference solves from a single
example, and stays at $0.53$--$0.58$ across the harder buckets, against the
\Fmac{} reference's $1.0$ (per-family learning curves and difficulty buckets are provided in the
released analysis notebook).

\paragraph{Classical nulls bracket the panel from both sides.}
The keystroke reference is deliberately the \emph{weakest} evidence-free
predictor, so we scored stronger classical nulls through the identical
pipeline (Methods; horizontal references in Fig.~\ref{fig:gap}). The
dataset's \emph{global output frequencies} (a single distribution ignoring
the query, the examples and the position) reach \Fmac{}-ICL-A $0.60$, and
the best input-agnostic predictor (the per-shot pooled \Fmac{} reference) reaches
$0.73$: every model in the panel is outscored on posterior fidelity by a
lookup table of output frequencies that performs no inference at all. Both
of these nulls are constructed from the frozen \Fmac{} references themselves (Methods),
so neither is achievable by a predictor that lacks the target; the
comparison bounds how much task-independent structure the models fail to
recover, and is not the claim that a deployable predictor beats them. From
the other side, Krichevsky--Trofimov add-$\tfrac12$ estimators fitted on the
visible prefix, the natural ``sequence statistics'' predictors, land
\emph{below} the keystroke reference ($0.41$ memoryless, $0.45$ bigram), and their
per-position continuation divergences ($0.38$ and $0.29$ bits) bracket the
panel's band ($0.29$--$0.50$ bits; Fig.~\ref{fig:gap}b). The panel's
distributional behaviour is thus indistinguishable in level from low-order
prefix statistics. This bracketing also gives the result its most
defensible form, and one that does not depend on the identity of the
target. Our anchor is itself an algorithmic-probability mixture, namely
the one induced by a \emph{print-only} machine with no loops; the
\Fmac{} reference is the mixture induced by a loop-bearing machine. Models sit
between the two, at the level of low-order prefix statistics. The claim
we are entitled to is therefore about the \emph{shape} of the measure
these models bring to such strings: it resembles a loop-free algorithmic
mixture or a Markov measure more than a loop-bearing one. That is a
structural distinction, and unlike the absolute distance to any one
\Fmac{} reference it is not at the mercy of a translation constant of the same order as our
entire description budget
(Methods).

\paragraph{Error anatomy: what models do when they are wrong.}
Because every task carries its exact posterior, each wrong answer can be
classified rather than only counted (answer-level records cover the $42$
complete API runs; empty replies are \emph{non-answers} (refusals or
delivery failures, $1{,}269$ calls) and are excluded from the error
taxonomy, though they still score as wrong in the accuracy facets). Restricted to task--shot cells
whose posterior support is non-degenerate, $31\%$ of wrong answers
carry exactly zero posterior mass under the evaluated build's truncated,
halting-conditioned support: no program in that support both matches the
shown examples and produces them. Because that support is bounded in
length and run time, and the prompt discloses neither bound, this is a
statement about the evaluated hypothesis class rather than a deductive
error by the model: an answer outside the support may still be realisable
by a longer or slower program. What it does show is that these answers
cannot be reached by any hypothesis the reference itself entertains, so
they are not near-misses within the scored class. A blunter symptom needs
no support argument at all: $46\%$ of errors do not even have the right
output length. The errors' positive structure is the pattern-completion
foil's: $41\%$ are verbatim copies of a shown example output, rising to
$70\%$ of errors by eight examples, with a recency preference for the
most-recent example. Only $2.8\%$ of errors are the Occam (Bayes-MAP)
answer, confirming at the item level that the shortfall is not
over-simplification. The served distributions show the complementary
failure: on the consistent support, models \emph{under}-commit. By eight
examples they
assign the truth $0.58$ where the \Fmac{} reference assigns $0.999$, and their
entropy after a single example is $1.46$ bits against the true
posterior's $0.93$. Models thus pay the penalty twice:
too flat where the posterior is sharp, and confidently wrong off its
support.

\paragraph{Termination is where model and target disagree most, and both
are miscalibrated.}
The continuation infidelity concentrates on the \emph{termination} symbol,
which carries $89\%$ of the divergence (\S\ref{sec:audit}). The
disagreement is genuine: across the $24$ API models with stored next-token
distributions ($46{,}800$ position-calls) the model's \code{END} mass is
\emph{anti}-correlated with the \Fmac{} reference's (Pearson $-0.16$, Spearman
$-0.11$). But the realised data show the error is two-sided rather than
the models'. Where the \Fmac{} reference is ${\ge}50\%$ sure the output ends, models
assign termination $0.24$ and the output actually ends $0.20$ of the time,
so the models are the better calibrated of the two; where the \Fmac{} reference is
${<}10\%$ sure, the true rate is $0.007$ and models assign $0.43$, so here
the models are badly wrong in the opposite direction. The one unambiguous
model failure is structural: at the first output position, where
termination is impossible and the \Fmac{} reference's \code{END} mass is exactly
zero, the median model still assigns $0.44$ and some assign $0.99$, which
is consistent with chat-format end-of-sequence leakage rather than with a
claim about programs, though we do not test that mechanism directly.
The honest summary is that models over-predict termination where it cannot
or does not happen, while the target over-predicts it wherever the length
quota has removed short outputs from the sample; we therefore report the
bit coordinate separately (\S\ref{sec:audit}) and rest no conclusion on
the termination coordinate alone.

\paragraph{In-context learning is non-monotone: models un-solve tasks.}
In this realisable, noiseless setting, consistent evidence never lowers
the posterior mass of the truth under any prior, and once an agent's
surviving hypotheses agree on the query its answer is frozen
(Appendix~\ref{app:priorfree}); a Bayesian agent can abandon a
correct answer only while an impostor hypothesis it prefers is still alive,
never after identification. The \Fmac{} reference shows exactly this lawful
profile on the frozen induction set: $46$ revisions against $406$ learnings
(one per nine), every one at the first or second example, none after its
candidate set collapses, and every one repaired by the final shot. The
panel's profile is quantitatively different: across the $98$ complete runs, additional examples produce
$6{,}545$ solved$\to$unsolved transitions against $13{,}702$ gains (one
forgetting event per two learnings, a median of $66$ per run on $120$
tasks), and $24\%$ of ever-solved (run, task) pairs end \emph{wrong} at
eight examples. This instability is uncorrelated with accuracy ($|\rho|<0.05$),
so it is not an artefact of weak models, and it tracks post-training
in both triads (Ministral-3-14B base$\to$instruct$\to$reasoning:
$9\to39\to54$ events; Qwen3-30B-A3B: $35\to60\to69$), giving the
post-training result a purely behavioural, distribution-free counterpart:
we read this as alignment tuning destabilising hypothesis maintenance.

\begin{figure}[!htbp]
  \centering
  \includegraphics[width=0.95\linewidth]{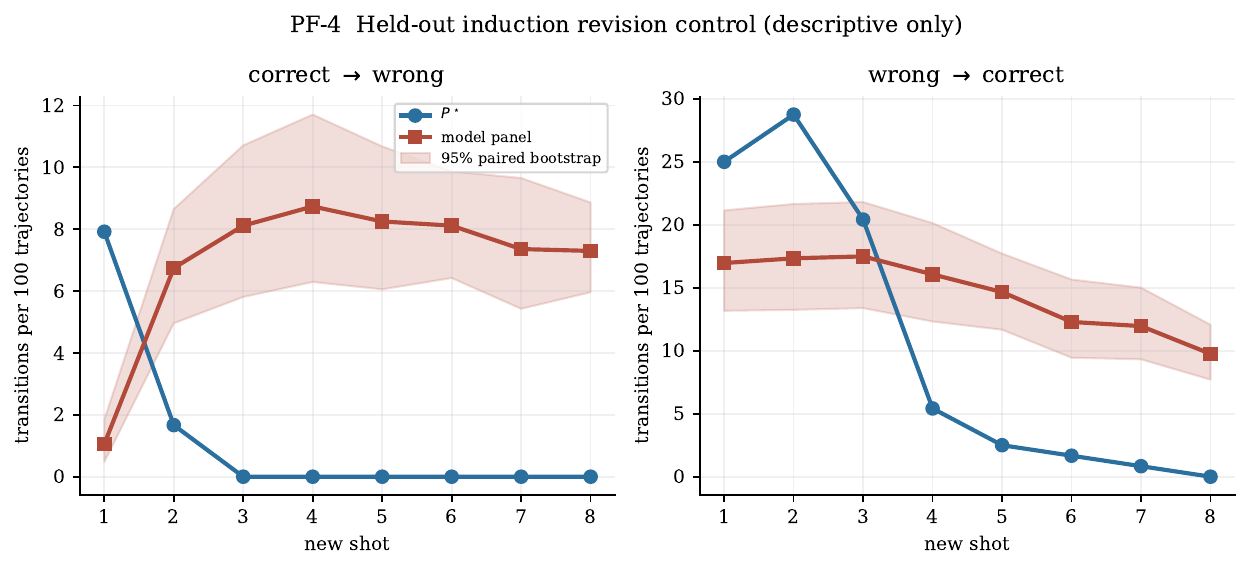}
  \caption{\textbf{Held-out induction revision control (descriptive).}
  Transition frequencies per 100 held-out trajectories by newly added shot.
  The model band is a 95\% bootstrap interval resampling complete runs and
  base-task clusters while keeping complement twins together. The reference
  posterior can revise correct$\to$wrong before identification; reversal by
  itself is therefore not evidence against every Bayesian prior. Counts are
  regenerated from the frozen data.}
  \label{fig:pf4-revision}
\end{figure}

\paragraph{Failures are idiosyncratic; difficulty is shared but not Bayesian.}
Item-level errors correlate only weakly across the panel: mean
pairwise Cohen's $\kappa$ on final-shot correctness is $0.17$, and even
serving and quantisation replicas of the \emph{same} gpt-oss-20b weights
agree only at
$\kappa=0.34$, the item-level face of the serving-reproducibility result
below. The union of runs solves every task, and a simple plurality vote
over the API panel scores $0.93$, above every single run in the panel
(the best three are tied at $0.92$): the
panel's answer knowledge is complementary, yet no aggregation touches the
distributional gap. At the same time, \emph{which} tasks are hard is a
stable property of models as a class (split-half reliability $0.90$), and
it is mostly orthogonal to Bayesian difficulty: Bayes sample complexity,
program length and prior mass of the truth jointly explain only $13\%$ of
the variance, while surface copyability does better (tasks whose truth
coincides with the \emph{modal} example output score $0.68$ versus $0.59$
otherwise, $p=0.014$; the six tasks whose truth never appears among the
example outputs are the hardest of all, at $0.42$). Models share a
difficulty axis, but it is the pattern-completion axis, not the
algorithmic one.

\begin{figure}[!htbp]
  \centering
  \includegraphics[width=0.9\linewidth]{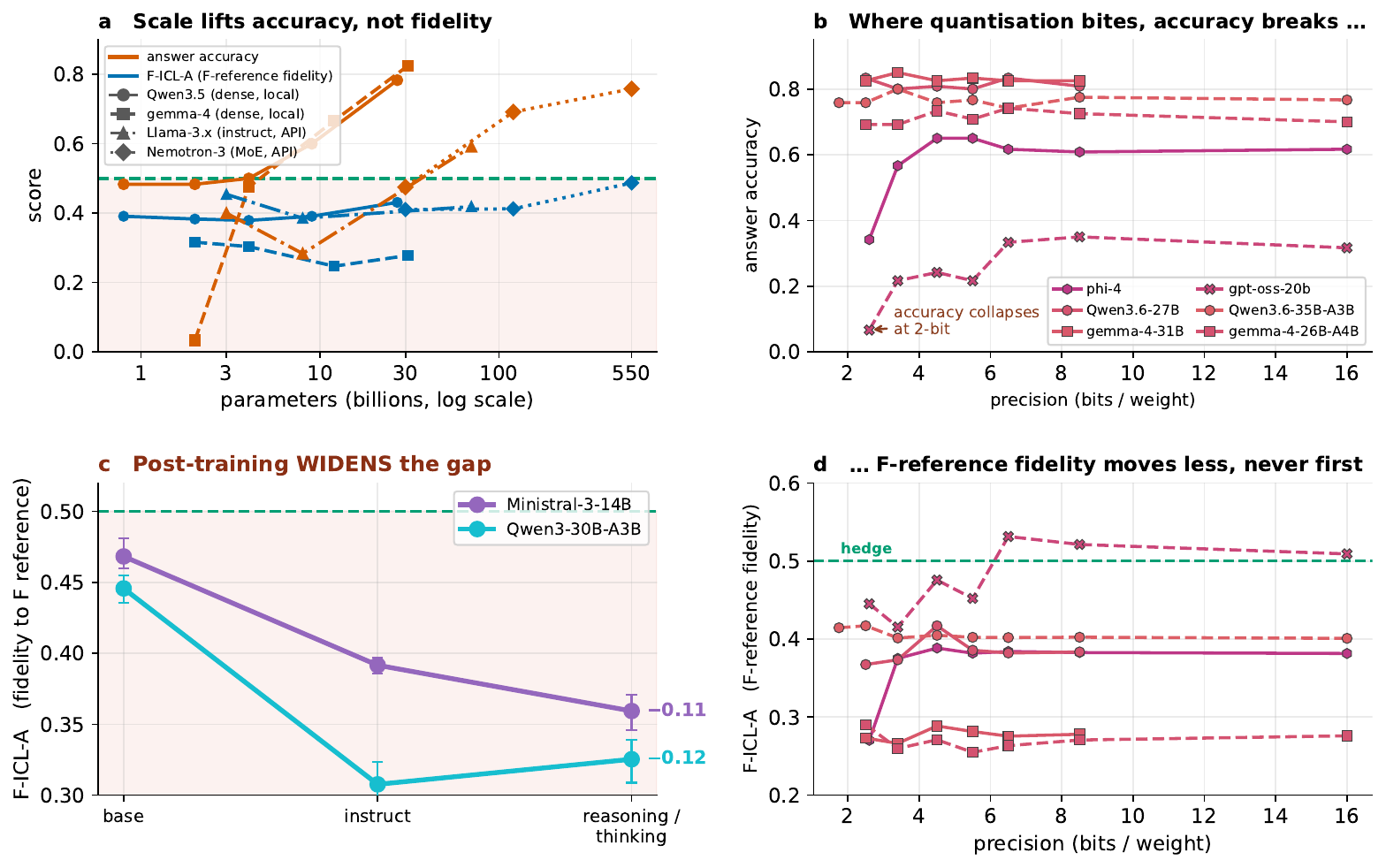}
  \caption{\textbf{What moves the gap.} The keystroke reference is marked in every
  \Fmac{}-ICL-A panel.
  \textbf{a}, Scale: the four within-family ladders the panel supports, two
  local dense families (Qwen3.5, circles; gemma-4, squares) and two API
  ladders held to one generation and post-training stage (Llama-3.x
  instruct, triangles; Nemotron-3 mixture-of-experts (MoE), diamonds). Answer accuracy (orange)
  climbs with parameters in every family, e.g.\ $+0.30$ across Qwen3.5's
  $34\times$ range, while posterior fidelity (blue) moves little and stays
  at or below the keystroke reference throughout.
  \textbf{b}, \textbf{d}, Quantisation, all six ladders (marker = family,
  colour = parameter count as elsewhere, dashed = MoE). Four ladders are
  flat in both facets down to the lowest precision served; in the two that
  degrade (gpt-oss-20b, phi-4), answer accuracy (\textbf{b}) breaks earlier
  and harder than \Fmac{}-ICL-A (\textbf{d}), and in no ladder does
  fidelity degrade first.
  \textbf{c}, Post-training: \Fmac{}-ICL-A falls sharply from base to
  instruct in both families and then continues down in one and edges back
  in the other in the only two open families that
  ship all three stages at matched size (change annotated; error bars are
  $95\%$ task-bootstrap CIs, twins resampled with their base task);
  alignment widens the gap.}
  \label{fig:moves}
\end{figure}

\paragraph{Scale, quantisation, post-training: what moves the gap.}
The controlled comparisons show that the interventions which most improve
conventional benchmarks do not close the fidelity gap, and one widens it
(Fig.~\ref{fig:moves}). These orderings are measured on the full divergence, whose ranking is
carried by the termination coordinate (Spearman $+0.97$) and not reproduced
by the bit coordinate ($+0.18$, $p=0.12$); they are termination-coordinate
findings (\S\ref{sec:audit}). \emph{Quantisation} turns out to be
benign for most of the panel: four of the six ladders are flat in both
facets down to the lowest precision served ($0.76$ accuracy at $1.75$
bits per weight for Qwen3.6-35B-A3B). Where quantisation does bite, it is
accuracy that breaks: \code{gpt-oss-20b} collapses from $0.32$ to $0.07$
and phi-4 halves from $0.62$ to $0.34$ at $2$-bit, while their fidelity
declines far less ($0.51\to0.45$ and $0.38\to0.27$, the latter only at the
extreme $2$-bit point; Fig.~\ref{fig:moves}b,d). The a-priori expectation
was the opposite, that the log-probability metrics would erode first; in no
ladder does fidelity degrade before accuracy. A distribution already far
from the \Fmac{} reference has little further to fall, whereas the argmax that
accuracy reads off is exactly what low-bit quantisation perturbs. \emph{Scale} raises accuracy but barely moves posterior fidelity
(Qwen3.5 $0.8$B$\to$$27$B: accuracy $+0.30$, \Fmac{}-ICL-A $+0.04$, still below
the keystroke reference; Fig.~\ref{fig:moves}a). \emph{Post-training} moves it the wrong way: in both
families the \emph{base} model has the highest fidelity and instruction tuning
lowers \Fmac{}-ICL-A (Ministral-3-14B $0.47\to0.39\to0.36$; Qwen3-30B-A3B
$0.45\to0.31\to0.33$; the base and instruct $95\%$ CIs do not overlap in
either family, and the base$\to$instruct drops exceed the CI half-widths
several-fold; Fig.~\ref{fig:moves}c) even as instruction tuning
holds accuracy. The base$\to$instruct step is the large and consistent one;
the further step to explicit reasoning continues the decline in one family
($0.39\to0.36$) and reverses slightly in the other ($0.31\to0.33$), so we
claim the instruction-tuning effect and not a monotone ordering across all
three stages (Fig.~\ref{fig:moves}c). Post-training
sharpens a model's distributions toward its aligned answer format and away from
the algorithmic prior, the dissociation the benchmark was built to detect.

\paragraph{Spend does not buy fidelity either.}
Across the seventeen metered frontier rows, cost trends with answer accuracy
(Spearman $\rho=+0.51$, $p=0.036$) but shows no detectable association with
posterior fidelity ($\rho=+0.30$, $p=0.43$, $n=9$ rows with a usable
distributional read-out); the panel's most faithful row costs \$$4.54$
against \$$8.80$ for the most expensive row that yields a fidelity score.
Only the accuracy leg reaches significance, and at this sample size we report
both as description rather than inference.

\begin{figure}[!htbp]
  \centering
  \includegraphics[width=0.7\linewidth]{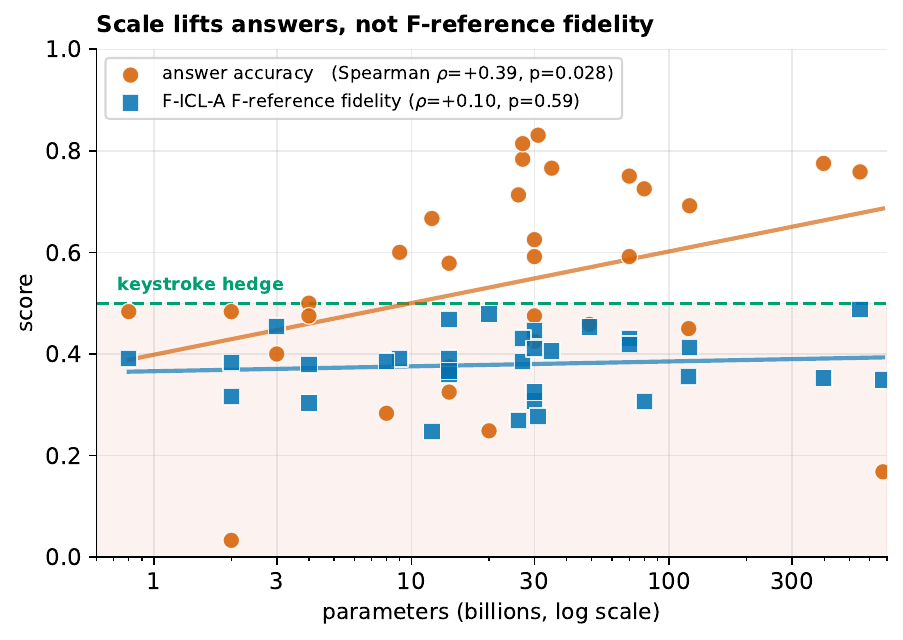}
  \caption{\textbf{Scale lifts answers, not posterior fidelity.} Base-model
  means for the $32$ panel members with disclosed parameter counts. Answer
  accuracy (orange) rises significantly across three decades of scale (Spearman
  $\rho=+0.39$, $p=0.028$); \Fmac{}-ICL-A posterior fidelity (blue) shows no
  detectable association ($\rho=+0.10$, $p=0.59$; $95\%$ CI $-0.26$ to $+0.43$)
  and remains below the keystroke reference at every size. Both are two-sided
  Spearman rank correlations against $\log_{10}$ parameters on the same $32$
  models, so the two tests have identical power. Lines are least-squares fits in
  $\log$-parameters and are shown for orientation only; the reported statistics
  are rank-based and unaffected by the log transform.}
  \label{fig:scaleinv}
\end{figure}

\paragraph{Posterior fidelity is scale-invariant where accuracy is not.}
The within-family scale ladder generalises to the whole panel as a double
dissociation (Fig.~\ref{fig:scaleinv}). These orderings are measured on the full divergence, whose ranking is
carried by the termination coordinate (Spearman $+0.97$) and not reproduced
by the bit coordinate ($+0.18$, $p=0.12$); they are termination-coordinate
findings (\S\ref{sec:audit}). Across the $32$ base models with
disclosed parameter counts, spanning $0.8$B to $675$B, nearly three orders of
magnitude, answer accuracy rises significantly with scale (Spearman
$\rho=+0.39$, $p=0.028$) while \Fmac{}-ICL-A shows no detectable association
($\rho=+0.10$, $p=0.59$), pinned just below the anchor at every size. Both
coefficients are Spearman rank correlations against $\log_{10}$ parameters,
two-sided, estimated on the \emph{same} $32$ models and therefore at identical
power; that shared power is what makes the contrast informative rather than an
artefact of sample size. We state the null with the caution it deserves: at
$n=32$ the smallest detectable $|\rho|$ at $\alpha=0.05$ is $0.35$, so the
fidelity result excludes a large positive association but not a moderate one
($95\%$ CI $-0.26$ to $+0.43$). What it does establish is that the two axes
move differently under the same test on the same models. On this evidence,
scale does not deliver fidelity to the \Fmac{} inductive reference over
the range accessible to open models today: whatever additional capability
parameters buy, we detect none of it on the fidelity axis.

\begin{figure}[htbp]
  \centering
  \includegraphics[width=0.95\linewidth]{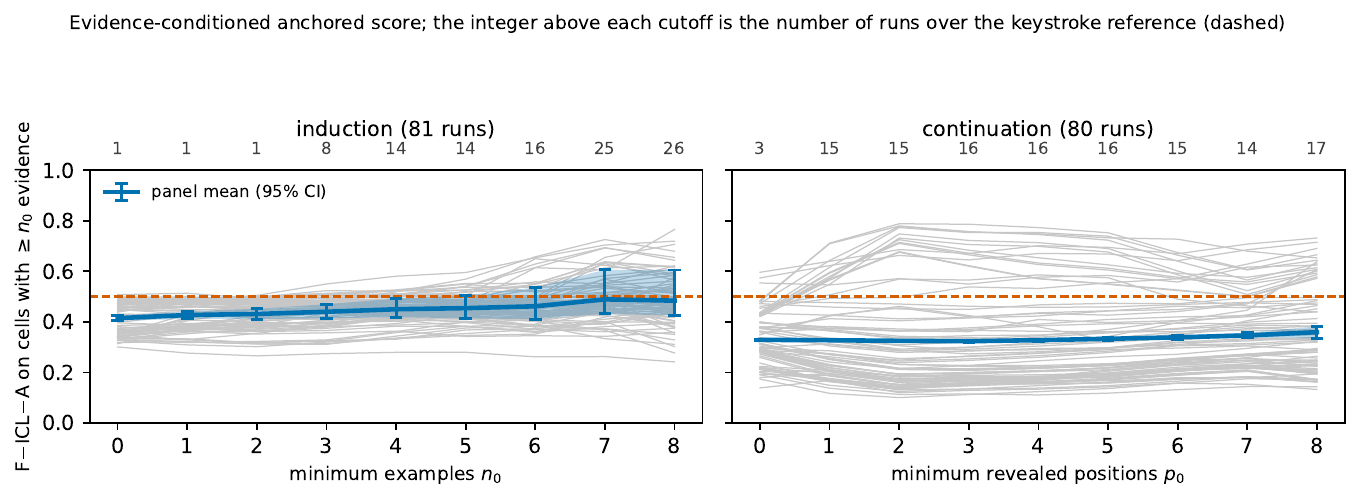}
  \caption{\textbf{The evidence-conditioned anchored score.} \Fmac{}-ICL-A
  recomputed on the cells with at least $n_0$ examples (induction, left) or
  $p_0$ revealed positions (continuation, right); thin lines are the $81$
  (respectively $80$) runs with a distributional read-out, the bold line is
  the panel mean with a task-bootstrap $95\%$ band, and the count above each
  cutoff is the number of runs over the keystroke reference (dashed).
  Induction rises with evidence but its mean stays below $\tfrac12$ at every
  cutoff; continuation is flat at every depth: the termination
  miscalibration does not fade with evidence. Computed entirely from the
  frozen artefacts; the driver is released with the code.}
  \label{fig:evidence}
\end{figure}

\paragraph{Evidence narrows the induction gap where scale does not; it does
not close it.}  The headline aggregates over all evidence levels, so we also
report the anchored score conditioned on minimum evidence
(Fig.~\ref{fig:evidence}): \Fmac{}-ICL-A recomputed on the cells with at
least $n_0$ examples, with the per-cell hedge and worst-case anchors
recomputed on the same cells so both branches of Eq.~\eqref{eq:skill} remain
exact.  On induction, whose read-out has no termination coordinate, the
panel mean rises from $0.414$ over all shots to $0.489$ at $n_0{=}7$
(task-bootstrap $95\%$ CI $[0.43,0.61]$), and the number of runs above the
keystroke reference rises from $1$ of $81$ to $25$ of $81$; the crossers
concentrate in the newest model families (the Qwen3.6 ladders and
Gemini~2.5~Pro from $n_0{=}3$--$4$, GPT-5.6 from $n_0{=}5$).  Two bounds on
the reading: the late-cutoff estimate rests on the cells where the posterior
has not yet collapsed (\S\ref{sec:standard}), so the intervals widen
accordingly; and the panel mean stays below $\tfrac12$ at every cutoff, so
the sub-reference headline is not an artefact of averaging over the
low-evidence regime.  On continuation the picture is unchanged at every
depth (mean $0.329$ to $0.359$): the termination miscalibration does not
fade with evidence, although excluding the first revealed position alone
moves $12$ of $80$ runs above the keystroke reference on this facet,
quantifying that
single structural cell's contribution (\S\ref{sec:audit}).  Evidence is thus
the one lever in this paper that \emph{improves} posterior fidelity, and it
improves only the bit-level facet.

\paragraph{Polarity twins expose a model bias the yardstick lacks.}
Because every task appears with its complemented twin and the \sFmac{}
\Fmac{} reference is provably identical on the pair (Methods~\ref{sec:methods-sf}),
the polarity gap $\Delta_{\mathrm{pol}}$ (Eq.~\eqref{eq:polgap}) is exactly
zero for the \Fmac{} reference and for \emph{any} complement-equivariant predictor, so
a nonzero value is attributable to the model alone. Quantisation ladders are correlated
replicas of one model, so the unit of analysis is the \emph{base model} (cluster
mean over its runs). Across the $46$ base models the mean gap is small and
positive but sensitive to panel composition: with the open panel alone the
clustered mean is $+0.0030$ ($t(36)=2.10$, $p=0.043$; $24$ of $37$ positive), while the frontier
rows, several of which carry comparatively large gaps of either sign (the
frontier GPT-5.6's is $-0.023$), move the full panel to $+0.0020$
($t(45)=1.40$, $p=0.17$; Wilcoxon signed-rank $p=0.037$; $28$ of $46$
positive). We read this as a per-model bias that is real and measurable on
a provably neutral instrument, with no panel-level sign consensus. On the
accuracy route the per-model bias is larger and two-signed (absolute mean
$0.06$), a confound invisible to any benchmark whose targets are not
polarity-balanced. The small mean, however, hides an item-level symmetry
violation: any solver that operates at the level of \emph{rules} solves a
task if and only if it solves its complement twin, yet within-run
original--twin correctness correlation is only $\phi=0.29$, and the median
model produces the bitwise-complemented answer on its twin task just $52\%$
of the time (range $20$--$72\%$; $40\%$ even restricting to pairs where both
answers are wrong, a pure style probe with accuracy removed).
Complement-equivariance, which the symmetrised machine grants the \Fmac{} reference
by construction, is itself a model capability the twins measure, and most
of the panel lacks it.

\paragraph{Serving method is not neutral: it changes delivery, content, and
even the sign of comparisons.}
Three models served two ways each (identical weights, identical prompts,
temperature $0$; calls paired exactly through the content-addressed cache)
decompose the serving effect into two mechanisms. The first is
\emph{delivery}: on the API side, \code{gpt-oss-20b} returned an empty
reply on $52\%$ of paired answer calls and MiniMax-M2.7 on $45\%$, while
the local builds always answered; scored naively over all paired cells, empties made the local build look
better for gpt-oss ($0.27$ against $0.24$), though the paired difference
is not significant on its own (McNemar exact $p=0.085$). The second is \emph{content}: restricted to pairs where both
stacks deliver, only $28\%$ ($34\%$) of greedy outputs are
byte-identical, and the accuracy comparison \emph{inverts}: the API side
is right on $0.50$ ($0.65$) of delivered pairs against the local $0.24$
($0.36$). The gemma-4 pair isolates content cleanly: served by Google's
own API with zero empty replies on either side, the same checkpoints agree
on only $18\%$ of outputs and the direction flips again, local $0.47$
against API $0.18$, and the 31B pair the same way ($0.57$ against
$0.38$, with $36\%$ of outputs identical). Serving thus
affects whether an answer arrives at all, what the answer is when it does,
and which side of a comparison wins, model-dependently. A leaderboard that
mixes serving providers inherits all three effects; \Fmac{}-ICL reports
the serving path per row and audits reply delivery explicitly. No
same-model serving pair exposes usable API log-probabilities, so the
analogous comparison for the \emph{distributional} readout remains open.

\subsection{Robustness of the measurement, and what the metric
licenses}\label{sec:audit}

A benchmark whose target is a designated distribution invites a specific
objection: the score measures agreement with that target, and any normative
reading of it depends on the target being the right object for the data
actually evaluated. Ours is not, in one respect that we now quantify.
Candidate tasks originate in bounded \Fmac{} program behaviours and are then
\emph{selected}: the continuation set uses equal quotas over output length and
complexity tercile, while induction adds learnability and difficulty gates
(Methods). If $S(z)$ denotes the probability that a generated task $z$ passes
that selection, then the evaluated law is
$q_{\rm eval}(z)\propto P^{F}(z)S(z)$, not $P^{F}(z)$. Because $S$ is deliberately
nonconstant, Eq.~\eqref{eq:posterior} is not the Bayes predictor for the
curated law. The
measurements below are all computed from the frozen artefacts
(the full audit is released with the code).

\paragraph{The score measures posterior fidelity, not predictive skill on
the sample.}
On the $1{,}950$ revealed continuation positions the \Fmac{} reference's mean
log-loss on the \emph{realised} token is $2.65$ bits against the keystroke reference's $1.59$, and it is the worse of the two on $57\%$ of cells; $39$ of
the $80$ scored runs predict the realised data better than the \Fmac{} reference
does. On induction the two are indistinguishable ($0.93$ bits each,
averaged over the nine shot counts; the keystroke reference is ahead at zero and two
examples). This is what curation does: quotas over length and complexity
deliberately over-sample outcomes the prior finds improbable, and a
predictor faithful to the prior is penalised on them. It follows that
\Fmac{}-ICL-A measures agreement with the algorithmic posterior and
\emph{not} predictive skill on this sample, and that ``below the keystroke reference''
cannot be read as ``worse than making no inference''. We use the weaker
and correct reading throughout: a model below the anchor is further from
the algorithmic posterior than a fixed length-prior null is.

\paragraph{Most of the raw divergence is the termination coordinate, where
the target is miscalibrated by construction.} Splitting the continuation
divergence into a termination coordinate (\textsc{end} versus not) and a
bit coordinate ($\{\code0,\code1\}$ renormalised) shows the termination
coordinate carries $89\%$ of it (mean $0.387$ of $0.435$ bits) and
reproduces the full ranking at Spearman $+0.97$. That coordinate is also
where the curation bites hardest: the shortest sampled output has length
three, so a string \emph{cannot} end at positions~1 or~2, yet the \Fmac{} reference
places mean \textsc{end} mass $0.37$ and $0.82$ there. A model that
correctly refuses to terminate early is penalised at exactly the positions
where the target is wrong about the evaluated sample.

\paragraph{The headline survives both corrections; its magnitude does
not.} Re-weighting cells to undo the length quota leaves the count
unchanged at $77$ of $80$ runs below the anchor. Discarding the
termination coordinate entirely and scoring only the bit coordinate, the
half of the measurement the quota cannot distort, leaves $69$ of $80$
runs below the corresponding anchor ($0.121$ against $0.057$ bits). The two sets are not nested: eleven runs leave and three enter, the three
being \code{gpt-oss-20b} replicas, so on the bit coordinate the panel's
sole at-anchor exception falls below it with the rest. The effect is
roughly a third of its raw size once termination is removed, and the
qualitative result is not an artefact of either the sampling design or the
termination miscalibration. But the two coordinates do not order the
panel alike. Spearman between the full metric and the bit coordinate is
$+0.18$ ($p=0.12$), against $+0.97$ between the full metric and
termination. Every ordering result in this paper (ladders, triads,
quantisation, clustering) is therefore measurably a
termination-coordinate finding, and we label them as such rather than as
statements about the bit-level posterior. Predictive regret tells the same story in
absolute terms: the panel pays $+1.86$ bits per token against the anchor
and $+0.79$ against the \Fmac{} reference.

\paragraph{What does and does not survive a different prior.}
Absolute distances, anchored scores and model orderings are properties of the
chosen \Fmac{} reference and need not survive an arbitrary change of universal
machine or prior. We do not claim otherwise. Three increasingly strong checks
separate the dependence. First, the following budget control varies the
finite approximation while holding the machine family fixed. Second, the
bit-only and curation-reweighted controls above change the evaluation
coordinate and law. Third, Appendix~\ref{app:priorfree} removes the numerical
choice of rival prior: on a genuine reveal stream, every fixed alternative
mixture that gives the computable \Fmac{} reference positive weight (equivalently,
finitely covers it) has bounded pathwise excess log-loss, and every
deterministic Bayesian mixture assigning positive mass to the realised truth
has bounded cumulative truth-loss. It further proves that two mixtures
assigning positive mass to the same computable deterministic truth have
bounded pathwise JS disagreement and hence a dyadic JS index tending to zero,
without sampling histories from $P^F$. PF-1 and PF-2 estimate the minimum finite
budgets already required, both for the 80-run operational read-out and for a
six-run unsmoothed projected validation. At nine positions they are lower
bounds, not a proof that no larger constant exists; this coverage-conditional
statement is therefore weaker than invariance of the headline score but
stronger in scope.

\paragraph{A budget change does not manufacture the result.} The
mismatch objection has a computable special case: if ``below the anchor''
were an artefact of the particular bounded prior, a different budget
should move it. Re-serving the induction \Fmac{} references from the complete
$L\le13$ store and rescoring every cached model distribution against them
(\S\ref{sec:eval-panel}) leaves the per-run ordering at Spearman
$0.975$ across a budget range of $L\le6$ to $L\le13$ and a change of
weighting convention. The count is not quite unchanged: on the induction
facet $80$ of $81$ runs are below the anchor under the pipeline-scale build \Fmac{} references and
$76$ of $81$ under the $L\le13$ store, so four runs cross when the budget
changes, all of them within $0.02$ of the anchor to begin with. The macro
count is stable at $77$ of $80$. That is a control over
\emph{budget}, not over the reference machine itself. The machine axis is
covered separately and externally: the certified description lengths of
\sFmac{} agree with published Coding Theorem Method values, computed over a
different machine family by output frequency, at Pearson $0.91$--$0.92$
(Methods). The two controls therefore bracket the objection, one varying
the machine at fixed object and the other the object at fixed machine, while
their conjunction, a conditional posterior served from another family,
remains untested and unclaimed.

\paragraph{The anchor is a length-prior null, not an uninformed one.} The
induction anchor renormalises $c(s)=(1/3)^{|s|+1}$ over the candidate set,
and that measure is length-decreasing exactly as the posterior is: it sits
closer to the \Fmac{} reference ($0.110$ bits) than a flat distribution over the
same candidates does ($0.122$). It is therefore not a ``no inference''
point, and our classical nulls say the same thing from the other side, the
frequency table scoring $0.60$ and the pooled predictor $0.73$ while
performing no inference at all. The anchor's role is to be a fixed,
computable, model-independent reference with an exact value of
$\tfrac12$; we name it the keystroke (print-prior) reference and no longer
read its sign as the presence or absence of inference.

\paragraph{What the model is told.}
The standard induction prompt presents input--output rows from ``one fixed
hidden program''; continuation says that ``a fixed program over bit strings''
produced the prefix (verbatim prompts in Appendix~\ref{app:prompts}). It does
not say that the program is written for \Fmac{} or Brainfuck, that programs are
weighted by description length, that $L\le13$, how nonhalting mass is treated,
or that the benchmark was quota-selected. Thus the experiment measures the
intersection between an LLM's default interpretation of ``program'' and the
declared \Fmac{} measure. It does not test whether an informed agent can compute
the \Fmac{} posterior, and it does not license ``the model should have used
$P^{F}$''. The present prompt ablation did not test such disclosure; any earlier
description of it as doing so was too strong. A decisive follow-up requires at
least three pre-registered arms: the current minimal prompt, a full
machine/prior/budget disclosure, and the same disclosure plus the curation law.

\paragraph{Prompt format moves the absolute score, and the direction is
diagnostic.} (The unmodified prompts are in
Appendix~\ref{app:prompts}.) The existing five-model ablation varies wording, templating, the grammar
constraint, example order and the terminator glyph, run on the same pipeline-scale build
the panel is scored against and with each model served through the template
its panel row uses, is released with the code (Qwen3.5-4B, Qwen3.5-9B,
Ministral-3-14B-Reasoning, gemma-4-12b-it, phi-4; $50$ continuation tasks per
arm). Two quantities appear below and must not be conflated: the anchored
score \Fmac{}-ICL-A (dimensionless, higher is better, Eq.~\eqref{eq:skill}) and
the raw Jensen--Shannon divergence to the \Fmac{} reference (in bits, lower is
better). Movements are reported in the score unless bits are named.
Wording, the grammar constraint and example order are immaterial:
rewording the instruction moves the score by at most $0.041$, removing the
grammar by at most $0.002$, and shuffling the example order moves full-shot
induction divergence by at most $0.017$ bits, against an \Fmac{} reference that is
order-invariant by construction.

Changing the \emph{template family} is not immaterial, but the benchmark's own
geometry identifies most of that movement as a read-out failure rather than a
fidelity gain. Serving under a bare chat template raises \Fmac{}-ICL-A by as much as $0.26$
in the four models whose template actually changes, while driving their answer
accuracy from $0.26$--$0.46$ down to $0.00$--$0.11$, and the resulting continuation
divergences collapse onto a single value: $0.189$ bits for Ministral-3,
Qwen3.5-9B and gemma-4-12b alike. That number is not a property of any model.
It is $\mathrm{JS}(P^{F}\,\|\,U)$ on exactly these $50$ tasks, $0.1888$
bits, the value returned when the endpoint resolves none of the three classes,
and it is the same signature the degeneracy audit of \S\ref{sec:eval-panel}
uses. phi-4 supplies the internal control: its panel row already uses the
default chat template, so this arm is a no-op for it, and its score moves by
$+0.001$ with accuracy unchanged. Two arms are genuine format effects rather
than breakage, and we report them as such: Ministral-3 under a raw completion
wrapper gains $0.090$ while its accuracy \emph{rises} ($0.260\to0.284$) and its
divergence stays well clear of the floor at $0.331$, and phi-4 gains $0.023$ at
accuracy $0.362$. The terminator glyph is likewise not immaterial: replacing
the full stop with \code{x} moves continuation divergence by up to $0.24$ bits
(Qwen3.5-4B, $0.469\to0.229$) in four of five models. Adding an explicit
chain-of-thought instruction without suppression costs score in four of the five
(up to $-0.136$) and is neutral in the fifth, which is the behaviour the
no-think wrappers exist to prevent.

Because a uniform read-out scores $0.189$ while
these models score $0.333$--$0.668$ when they are following the format,
\emph{breaking a model's output format improves its \Fmac{}-ICL-A score}: the
numbers we report are what models achieve when the instrument is working, and
this failure mode inflates rather than deflates them. We therefore treat the \emph{relative} comparisons at fixed
format (every ladder, triad, quantisation and panel ordering in this paper) as
the load-bearing results, and read any single row's absolute distance from the
anchor as conditional on the serving format, of which a template change can
consume a large fraction.

\subsection{Continuation improves late but does not converge}
\label{sec:sequential-results}

\begin{figure}[!htbp]
  \centering
  \includegraphics[width=\linewidth]{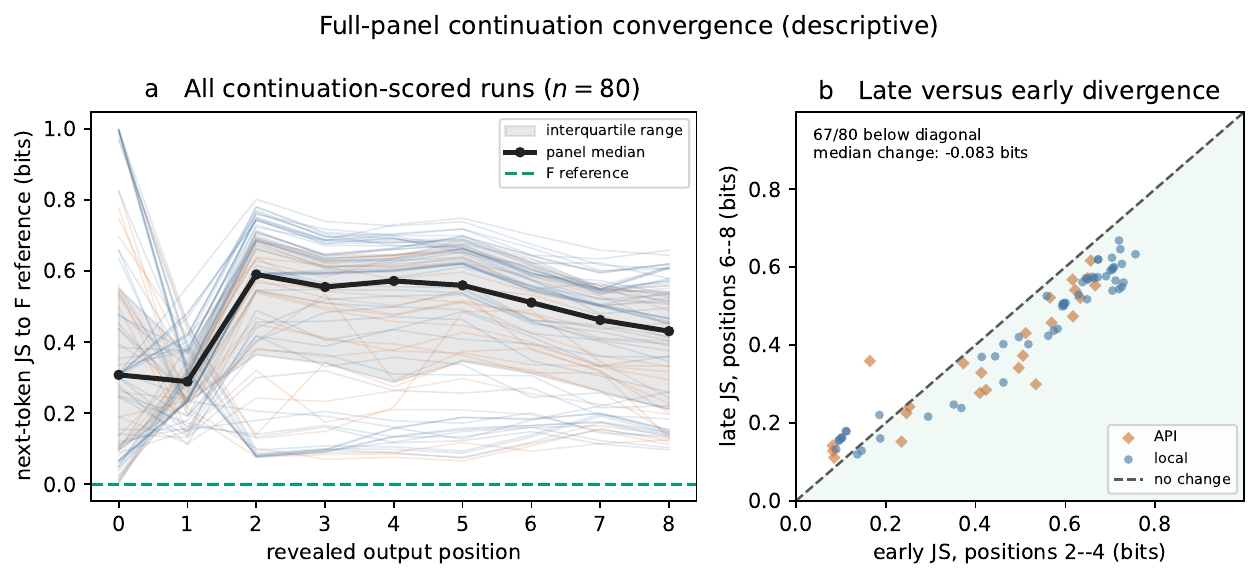}
  \caption{\textbf{Full-panel continuation convergence (descriptive).}
  \textbf{a}, Every usable serving-configuration curve; blue denotes local
  runs, orange API runs, the black line the cross-run median and the band the
  interquartile range. \textbf{b}, Each run's mean divergence at positions
  2--4 against positions 6--8; points below the diagonal improve. The
  job-level distributions use the legacy top-$k$ fold and epsilon smoothing;
  this figure tests the empirical divergence direction, while
  Fig.~\ref{fig:pf6-operational} applies the pathwise identities to that fixed
  operational read-out.}
  \label{fig:pf0-all}
\end{figure}

The continuation family lets us examine how every served distribution moves
as the true output is revealed, rather than reducing each run to one mean
score. Of the 105 serving configurations, 80 contain finite continuation
distributions at all positions 0--8; 24 are induction/accuracy-only and one
continuation record is entirely nonfinite. Figure~\ref{fig:pf0-all}a draws
every usable curve. The position-wise median rises sharply to $0.590$ bits at
position two and then declines, non-monotonically, to $0.431$ bits at position
eight. Comparing
blocks within each run, 67 of 80 configurations have lower mean JS at
positions 6--8 than at positions 2--4, and the median change is $-0.083$ bits
(Fig.~\ref{fig:pf0-all}b). This direction is not an artefact of counting
quantisations as independent models: after grouping all quantisations of a
checkpoint, 38 of 45 checkpoint families improve (one-sided exact sign test,
$p=1.56\times10^{-6}$), and a checkpoint-cluster bootstrap places the mean
change at $-0.066$ bits (95\% CI $[-0.087,-0.043]$). The test concerns the
\emph{direction} of the late--early change, not convergence to zero; its null
treats checkpoint-family direction signs as exchangeable and does not treat
the model panel as a random draw from all LLMs. There is therefore broad
late-horizon movement
toward the designated posterior, but not convergence to it within the tested
horizon: the median late-block divergence remains $0.465$ bits rather than
approaching zero.

This descriptive result is deliberately separated from the stronger
finite-budget question. Position-wise means condition on outputs long enough
to reach each position, and the panel was length--complexity curated rather
than sampled from $P^F$. Moreover, the full-panel job files store the legacy
top-$k$ fold after epsilon smoothing. This blocks recovery of each provider's
native full-vocabulary probability, but it does not make the stored
distribution mathematically unusable. Below we define that fixed
transformation itself as the operational three-class predictor
$\widetilde Q$. Appendix~\ref{app:priorfree} proves the pathwise and
deterministic-truth bounds that apply to $\widetilde Q$, and separately states
why the curated average cannot be substituted into an \emph{expected}
$P^F$-sampling theorem.

\paragraph{All-panel pathwise test.}
For every history, let $\widetilde Q$ be exactly the distribution recorded by
the benchmark: top-$k$ token masses are folded onto
$\{\code0,\code1,\textsc{end}\}$, $10^{-9}$ is added to each class, and the
three values are normalised. This deterministic history-to-distribution map is
a coherent predictor and induces a joint distribution by the chain product.
It is the operational read-out actually scored in the paper, not a claim about
the provider's latent native vocabulary. We can therefore compute the
pathwise identities for all 80 configurations and 23,998 of 24,000
configuration--trajectory pairs without any sampling-law assumption.

Figure~\ref{fig:pf6-operational}a--b shows every configuration rather than a
selected model subset. At the stopped endpoint, $29.6\%$ of operational paths
already require a finite-cover penalty exceeding 10 bits and $18.6\%$ exceed 20
bits (simultaneous 95\% lower bounds $11.7\%$ and $0.65\%$). The reference-free
truth budget is larger: $86.7\%$ and $39.3\%$ of paths exceed 10 and 20 bits
(lower bounds $68.3\%$ and $20.9\%$). At a 40-bit ceiling the point estimates
are $3.50\%$ and $13.0\%$, but both simultaneous lower bounds are $0\%$: the
panel does not resolve that tail at cluster level, and we rest no claim on it. Panels c--d report the full threshold-survival curves, so no
single ceiling is selected after viewing the data. Their simultaneous 95\%
bands use crossed resampling of 45 checkpoint families and 150 complement-twin
task clusters. They are a cluster-stability analysis of the curated panel,
not a claim that its tasks were sampled from $P^F$. The bands are wide and narrow the
claim accordingly: the simultaneous lower bound stays positive at a 20-bit
operational coverage ceiling (observed $18.6\%$, lower bound $0.65\%$) but
is already zero at the next tabulated ceiling, 22 bits, so only
the 10- and 20-bit statements survive cluster resampling.

\begin{figure}[!htbp]
  \centering
  \includegraphics[width=\linewidth]{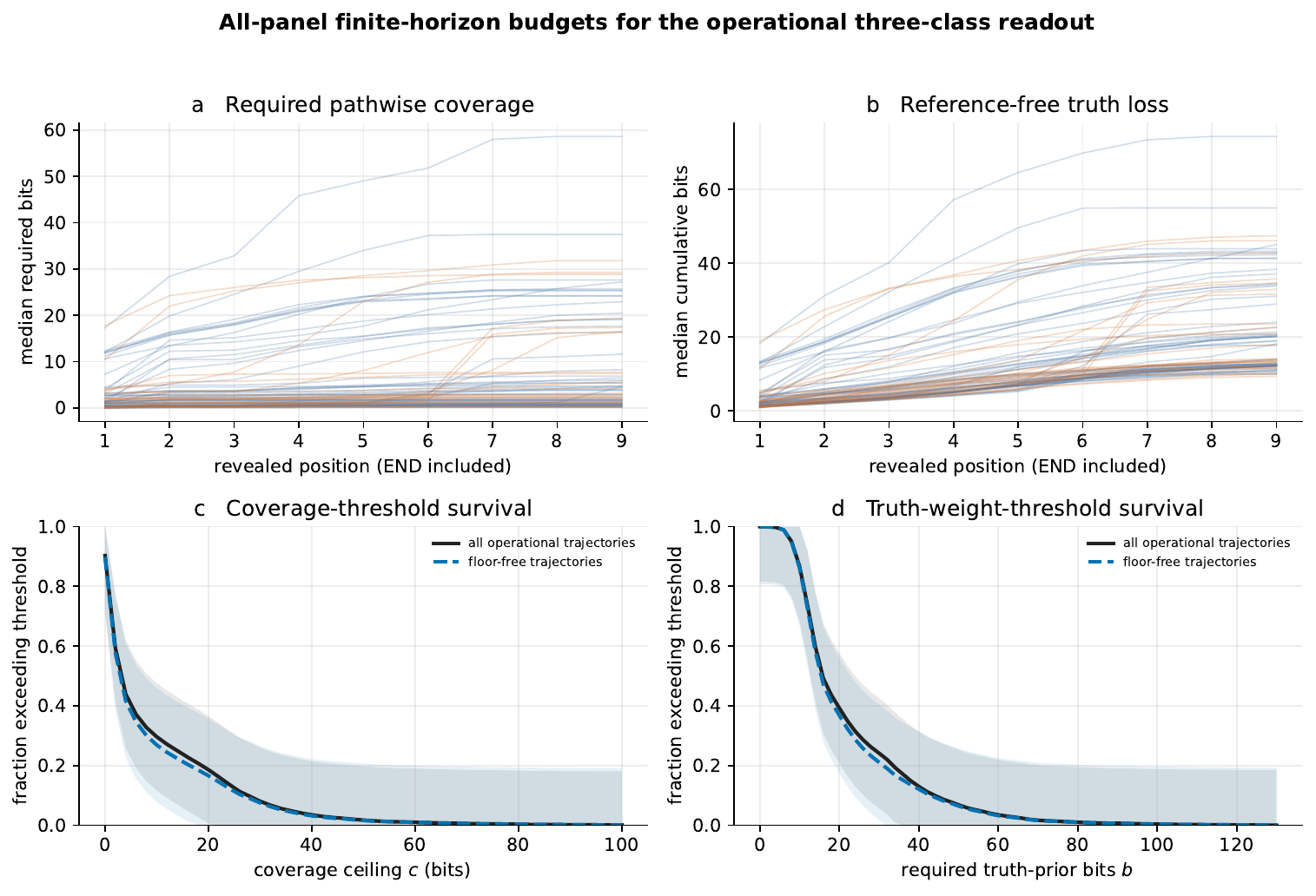}
  \caption{\textbf{All-panel finite-horizon budgets for the operational
  three-class read-out.} \textbf{a--b}, Per-configuration medians of the
  required pathwise coverage and cumulative realised-symbol truth loss; blue
  denotes local and orange API configurations. Values are stopped after END.
  \textbf{c--d}, Exact fractions of configuration--trajectory pairs exceeding
  every displayed budget, with simultaneous 95\% crossed-cluster bootstrap
  bands. Dashed curves remove an entire trajectory if any realised symbol was
  assigned the $10^{-9}$ floor. These are finite-horizon lower-budget and
  sensitivity curves, not rejection of every possible finite constant.}
  \label{fig:pf6-operational}
\end{figure}

For six frontier runs the endpoint caches retain the folded masses of $0$,
$1$, and END before smoothing. Normalising those masses gives an exact,
no-clipping top-$k$-projected scoring rule; trajectories whose realised class
is absent are censored rather than assigned artificial mass. The retained
counts are 298, 299, 222, 49, 75, and 126 of 300 for Gemini~2.5~Pro and
GPT-5.1 through GPT-5.6-sol, respectively. Figure~\ref{fig:pf1-coverage}
reports the resulting minimum pathwise coverage budget. By position nine the
largest observed retained path requires between $2.88$ bits (GPT-5.5, 75 of
300 trajectories retained) and $95.22$ bits (GPT-5.1, 299 of 300), while run
medians range from $0.29$ to $31.78$ bits. This range must not be read as a
comparison between models. Censoring is not neutral: it removes exactly the
trajectories whose realised class had zero recorded mass, which are the
trajectories with the largest budgets, so a run's retained maximum falls as
its censoring rises and the least surprised-looking runs here are the most
heavily censored ones. The retained values are therefore lower bounds within
each run, biased downward by an amount that grows with censoring. They are
finite-panel lower bounds on any covering constant, not estimates of one
universal ceiling and not a rejection of every larger finite constant. The
all-panel operational estimand above avoids this selection entirely, at the
cost of scoring the smoothed read-out rather than the raw one.

\begin{figure}[!htbp]
  \centering
  \includegraphics[width=\linewidth]{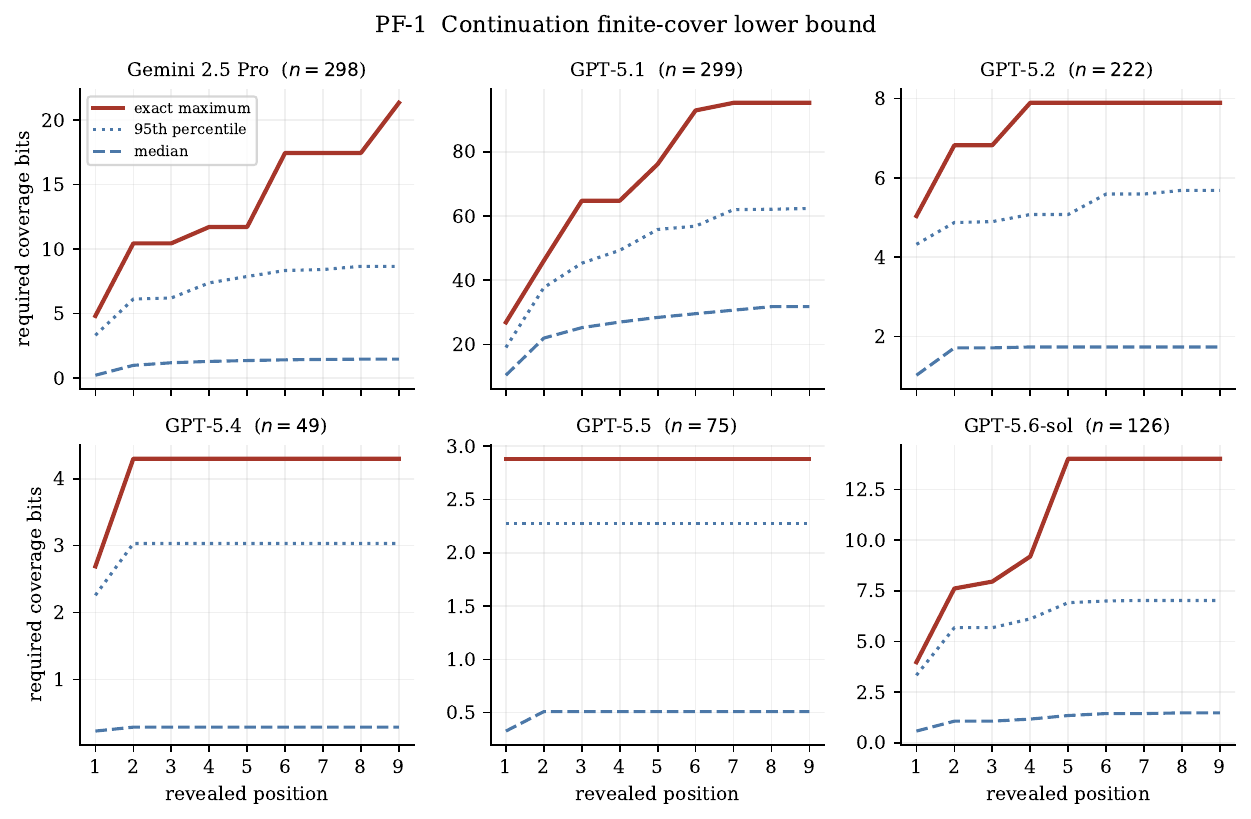}
  \caption{\textbf{Continuation finite-cover lower bound (PF-1).}
  Required pathwise coverage bits for the exact, no-clipping top-$k$-projected
  scoring rule on complete retained trajectories. The solid curve is the
  maximum over trajectories and therefore a lower bound on every covering
  constant for the tested domain; the 95th percentile and median describe the
  retained panel. Cumulative values are held fixed after END.}
  \label{fig:pf1-coverage}
\end{figure}

\begin{figure}[!htbp]
  \centering
  \includegraphics[width=\linewidth]{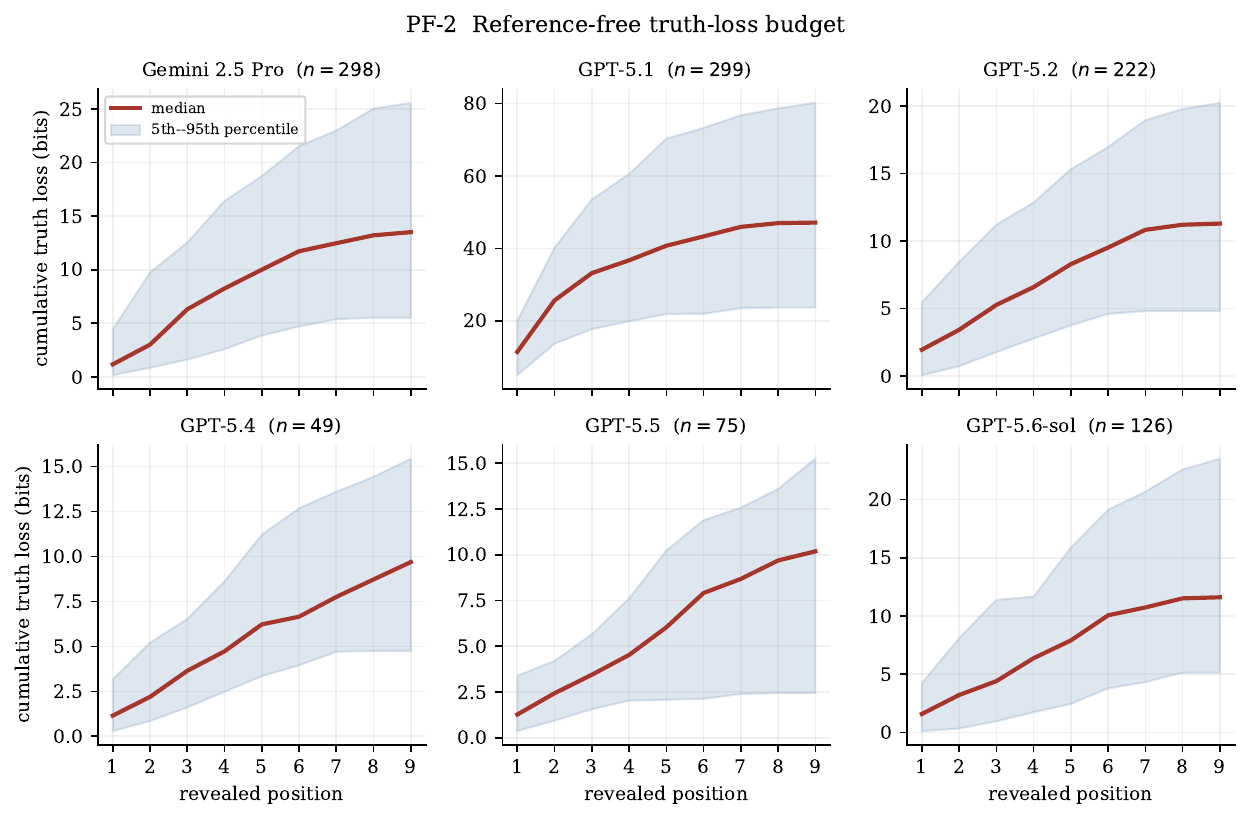}
  \caption{\textbf{Reference-free continuation truth-loss budget (PF-2).}
  Cumulative realised-symbol log-loss for the same projected scoring rule and
  retained trajectories. Lines are medians and bands the 5th--95th
  percentiles; values remain fixed after END. The finite curves measure the
  budget already consumed, while the asymptotic boundedness claim is proved
  and scoped in Appendix~\ref{app:priorfree}.}
  \label{fig:pf2-truth}
\end{figure}

The reference-free view leads to the same qualitative caution. Median
cumulative realised-symbol log-loss at position nine ranges from $9.69$ to
$47.13$ bits across the six runs (Fig.~\ref{fig:pf2-truth}). Any fixed
Bayesian mixture over deterministic hypotheses must pay for the realised
truth from a finite lifetime budget determined by its initial truth weight;
the observed curves quantify how much of that budget is already required.
They do not establish divergence on strings that end after at most nine
symbols. The coexistence of falling late-position JS in the full panel and
growing cumulative truth loss is not contradictory: the former is a local,
cross-sectional divergence, whereas the latter adds realised-symbol loss
along each stopped trajectory.

\begin{figure}[!htbp]
  \centering
  \includegraphics[width=\linewidth]{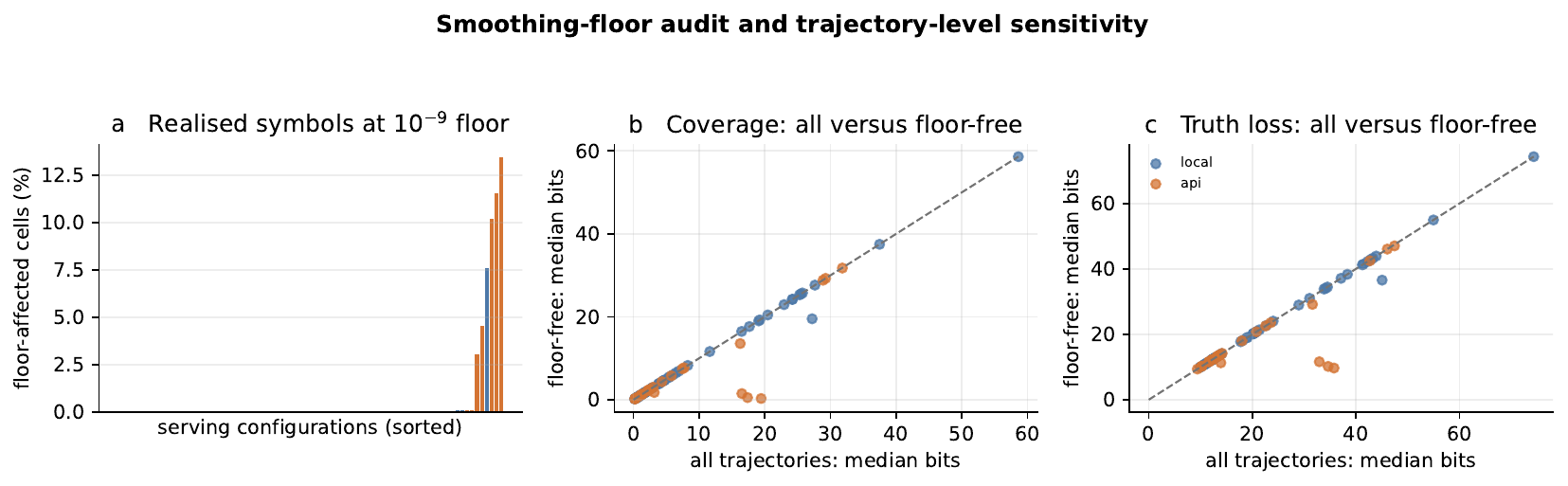}
  \caption{\textbf{Smoothing-floor audit.} \textbf{a}, Fraction of realised
  symbol cells at the operational $10^{-9}$ floor in each configuration,
  sorted by rate. \textbf{b--c}, Endpoint per-run medians with all trajectories
  against medians after removing every trajectory touched by the floor, for
  pathwise coverage and truth loss. The diagonal denotes equality. The
  concentration of departures in a few runs motivates reporting both
  estimands rather than discarding the other 74 configurations.}
  \label{fig:pf7-smoothing}
\end{figure}

The raw-cache analysis is an unsmoothed validation, not the only theorem-valid
use of the caches. Across the full operational panel, only 991 of 155,984
realised-symbol cells lie at the $10^{-9}$ floor ($0.635\%$), and 3,313
($2.12\%$) are at or below $10^{-6}$. Sixty-eight of 80 runs contain no floor
event at all and 74 contain at most two; the floor is concentrated in six
runs. Removing every floor-affected trajectory leaves 23,069 endpoint
observations and closely follows the all-trajectory threshold curves
(Fig.~\ref{fig:pf7-smoothing}). On this stricter subset, $26.9\%$ of paths
still exceed 10 coverage bits (simultaneous 95\% lower bound $7.86\%$), and
$36.8\%$ still consume more than 20 truth-loss bits (lower bound $17.6\%$).
Thus smoothing materially affects a small set of configurations but does not
generate the panel-wide budget relation. The six pre-smoothing caches remain
valuable because they identify a separate projected-native estimand and bound
the unreported mass.

\paragraph{Finite-scale longitudinal signature.}
The sharper unknown-$C$ question is whether successive evidence blocks enter
the vanishing-tail regime of Eq.~\eqref{eq:jstailzero} or retain the
positive-floor signature of Eqs.~\eqref{eq:jsrichardson}--\eqref{eq:jsindex}.
For $T\in\{1,2,4\}$ we therefore compute the mean contribution in positions
$T{+}1{:}2T$ and its ratio to the cumulative contribution in positions
$1{:}T$. Every trajectory is stopped after END, contributing zero thereafter,
and checkpoint quantisations and complement twins remain clustered.

At the largest available scale, $T=4$, the JS tail still contributes $0.271$
bits per step (simultaneous crossed-cluster 95\% interval
$[0.208,0.333]$), and the dyadic index is $0.666$. Thirty-seven of 45
checkpoint-family indices lie closer to the positive-floor benchmark $1$ than
to the finite-budget limit $0$ (one-sided exact sign test against the natural
midpoint $1/2$, $p=7.69\times10^{-6}$). This is not a flat plateau: the
$T=4$ tail is lower than the $T=2$ tail in 44 of 45 checkpoint families
($p=1.31\times10^{-12}$). The longitudinal conclusion is therefore two-sided
in substance: improvement is real, but by the largest observable block the
panel has not entered a near-zero tail regime and remains closer to the
positive-floor geometry.

\begin{figure}[!htbp]
  \centering
  \includegraphics[width=\linewidth]{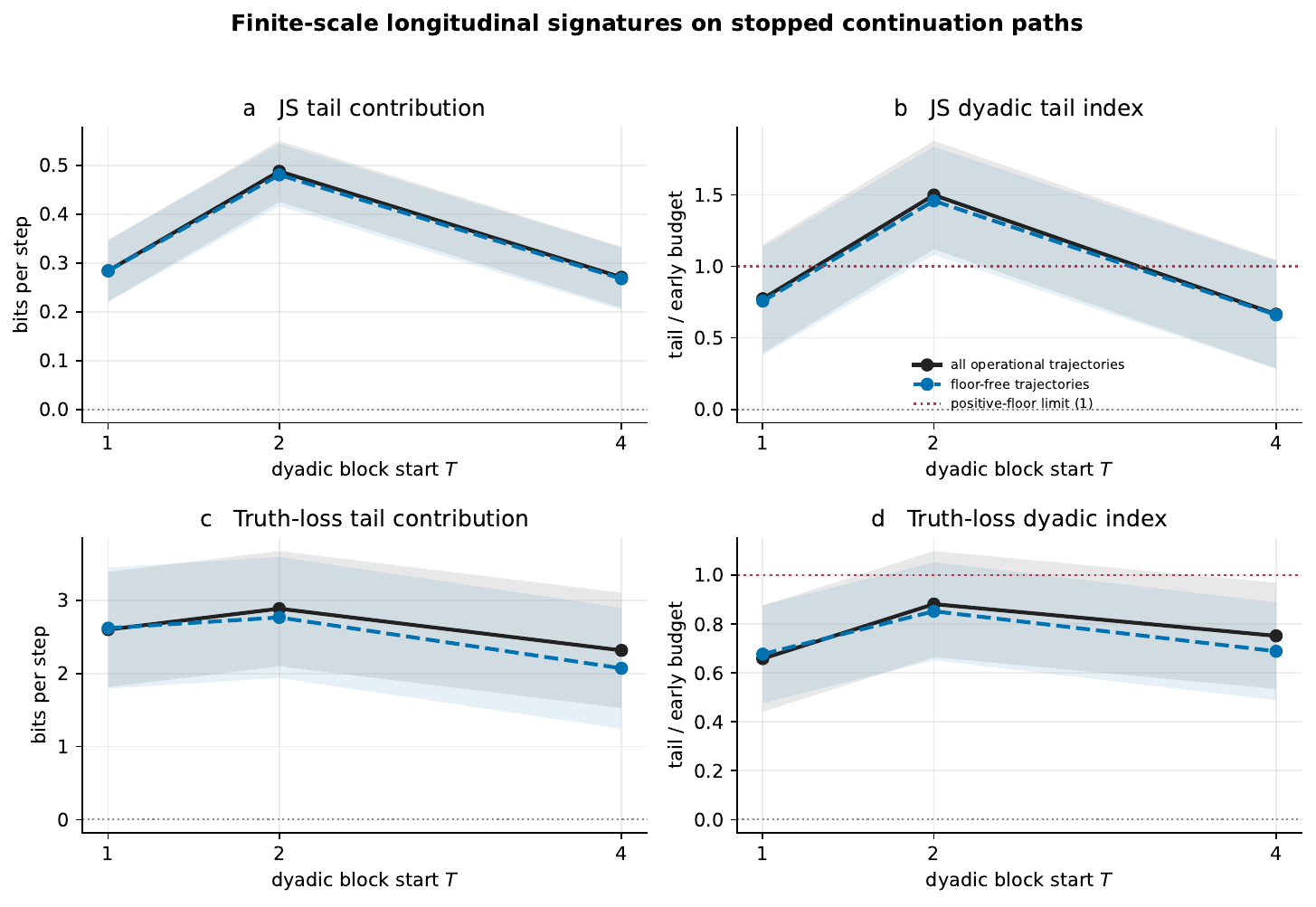}
  \caption{\textbf{Finite-scale longitudinal tail signatures.}
  \textbf{a--b}, The stopped dyadic JS tail contribution
  $D^{\rm tail}_{\rm JS}(T,2T)$ and tail index
  $I^{\rm tail}_{\rm JS}(T)$ at every identifiable dyadic scale.
  \textbf{c--d}, The corresponding reference-free realised-symbol truth-loss
  contribution and ratio. Lines give equal-checkpoint/equal-twin estimates and
  bands simultaneous 95\% crossed-cluster intervals; dashed blue removes a
  whole trajectory after any smoothing-floor event. The dotted red line marks
  the positive-floor ratio $1$ and grey line the finite-budget limit $0$.
  These are scale-local signatures under the curated evaluation law, not
  estimates of a $P^F$-sampled infinite-horizon limit.}
  \label{fig:pf8-longitudinal}
\end{figure}

The reference-free truth-loss calculation is stronger with respect to the
choice of machine. Positions 5--8 still cost $2.315$ realised-symbol bits per
step (simultaneous 95\% interval $[1.527,3.104]$), and their aggregate dyadic
index is $0.752$. At checkpoint-family level, 39 of 45 indices are closer to
$1$ than $0$ ($p=2.71\times10^{-7}$). Removing every floor-affected trajectory
leaves the JS contribution and index at $0.268$ and $0.661$, and the truth-loss
values at $2.070$ bits per step and $0.689$, respectively
(Fig.~\ref{fig:pf8-longitudinal}). Thus neither result is generated by the
$10^{-9}$ replacement.

The commitment of \S\ref{sec:audit} applies here as well, and it changes what
this result may be said to show. Splitting the realised-symbol loss by the
chain rule into a termination coordinate and a bit coordinate, the tail
signature is carried by termination: the \textsc{end} coordinate contributes
$1.726$ of the $2.315$ bits per step, with index $1.032$ and 39 of 45 families
closer to $1$, whereas the bit coordinate alone contributes $0.590$ bits per
step with index $0.419$, and only 3 of 45 families lie closer to $1$. The sign
test therefore reverses on the bit coordinate. Within nine positions, the
panel's residual positive-floor geometry in the reference-free budget is a
statement about \emph{when models think the output stops}, not about the bits
they emit while it continues; on the bit coordinate the observed trajectories
are already in the decaying-tail regime and supply no evidence against a
bounded budget. The endpoint budgets decompose the same way, though less
starkly: of the paths exceeding 10 and 20 truth-loss bits ($86.7\%$ and
$39.3\%$), $26.1\%$ and $8.4\%$ do so on the bit coordinate alone, and the mean
required coverage falls from $9.39$ to $4.27$ bits. Both coordinates are real
costs a covering predictor must pay, and termination is a genuine part of the
prediction problem rather than a formatting artefact; but the asymptotic
reading rests on the coordinate this paper has undertaken not to rest
conclusions on by itself.

\begin{figure}[!htbp]
  \centering
  \includegraphics[width=\linewidth]{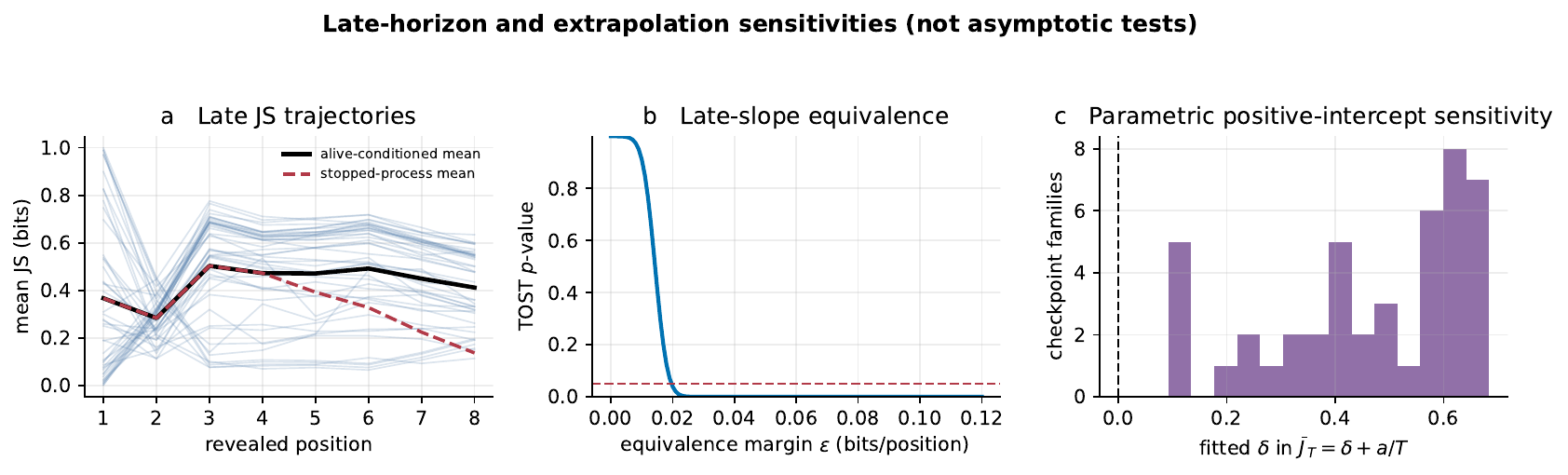}
  \caption{\textbf{Late-horizon and extrapolation sensitivities.}
  \textbf{a}, Alive-conditioned checkpoint-family curves, their mean, and the
  stopped-process mean that falls as finite episodes end. \textbf{b}, TOST
  $p$-value for equivalence of the alive-conditioned late slope to zero over
  every displayed practical margin. \textbf{c}, Checkpoint-family intercepts
  from the explicitly parametric finite-range fit
  $\bar J_T=\delta+a/T$. The panels test current-scale geometry and modelling
  sensitivity; none supplies observations beyond position eight.}
  \label{fig:pf9-extrapolation}
\end{figure}

Two extrapolative sensitivities make that final limitation visible rather
than hiding it (Fig.~\ref{fig:pf9-extrapolation}). Among trajectories still at
risk, the checkpoint-cluster mean JS slope over positions 4--8 is
$-0.0143$ bits per position (90\% $t$ interval
$[-0.0196,-0.0090]$); a TOST analysis establishes equivalence to zero only
after the tolerance reaches $0.020$ bits per position. A deliberately
parametric fit of $\bar J_T=\delta+a/T$ over $T=4,\ldots,8$ gives positive
$\delta$ in all 45 checkpoint families (mean $0.464$ bits; sign-test
$p=2.84\times10^{-14}$), and BIC favours the intercept model in all 45.
This fit is evidence for a positive finite-range intercept \emph{conditional
on that functional form}; Eq.~\eqref{eq:cesarorate} does not require the $a/T$
approximation to be accurate this early, so we do not promote it to an
asymptotic rejection.

\paragraph{What this does, and does not, establish across priors.}
Computing a dyadic contribution or ratio does not itself require random
sampling: it is an algebraic summary of the observed sequence of conditional
losses. Its inferential meaning does depend on the loss and averaging law.
For JS, Eqs.~\eqref{eq:jstail}--\eqref{eq:jsindex} concern
$j_t=\mathbb E_{H_t\sim P^F}\mathrm{JS}(P^F_t\|Q_t)$, whereas the plotted
mean uses the curated evaluation law. Figure~\ref{fig:pf8-longitudinal}
therefore establishes that the tested cases have not displayed the
finite-scale merging signature; it is not a population estimate of the
$P^F$-expected theorem.

There is nevertheless a separate pathwise result that requires no $P^F$
sampling. Every benchmark story is the computable deterministic output of an
enumerated program. If both $P^F$ and an alternative Bayesian mixture $Q$
assign that truth positive mass, their cumulative truth losses are bounded;
the inequalities
$\mathrm{JS}(P^F_t\|Q_t)\le\operatorname{TV}(P^F_t,Q_t)
\le[1-P^F_t(\sigma_t)]+[1-Q_t(\sigma_t)]$
then make the pathwise cumulative JS bounded as well. Hence both the pathwise
JS and reference-free truth-loss dyadic indices must eventually tend to zero
for another universal Bayesian prior. Appendix~\ref{app:priorfree} gives the
proof. The observed ratios are accordingly finite-scale, prior-robust
diagnostics on the curated computable stories, although the short terminating
horizon cannot exclude a larger finite prior penalty or convergence beginning
later.

This yields a deliberately finite-horizon, prior-robust conclusion. Exact
agreement with the \Fmac{} posterior is not required of a predictor based on
another universal prior: on a genuine $P^F$ reveal stream, finite coverage
predicts eventual merging. Across the complete observable horizon, however,
the model panel does not exhibit the corresponding finite-scale convergence
geometry, including under the reference-free truth diagnostic, whose residual
tail signature is carried by the termination coordinate: on the bit
coordinate alone the trajectories are already in the decaying-tail regime
(\S\ref{sec:sequential-results}). Thus the
observed failure cannot be dismissed merely as numerical disagreement with
the exact \Fmac{} prior. We do \emph{not} infer that the reference-machine
choice is irrelevant, that no other universal prior could generate the
finite responses, or that language-independent asymptotic non-convergence
has been proved: an alternative prior with a sufficiently large constant or
delayed transient is not excluded by sequences ending at position nine.

The statistical statement is equally scale-specific. Inverting the
simultaneous 95\% intervals rejects a position-5--8 tail no larger than any
prespecified $\varepsilon<0.208$ JS bits per step, or
$\varepsilon<1.527$ realised-symbol bits per step, under the crossed curated
panel law. The exact index tests reject the null that no more than half of
checkpoint families lie on the persistent side of the natural midpoint
$I=1/2$. No finite test can reject the unrestricted union of all proper
priors: their truth weight can be arbitrarily small, allowing an arbitrarily
large finite budget and arbitrarily delayed convergence. The confidence claim
is therefore that the tested panel has not entered any practically near-zero
tail regime at the observed scale, not that every possible proper prior is
excluded.

These casewise statements should not be conflated with PF-3. The pathwise
truth-loss inequality holds for each curated sequence and hence for every
plotted case. PF-3 asks a different population question: the expected JS
budget when histories themselves are drawn from $P^F$. Averaging the present
JS statistics estimates the curated evaluation law, not that expectation;
this precludes the PF-3 population interpretation, not calculation of the
scale-local statistic or the pathwise truth-loss test above.
Nor can this be repaired by multiplying each case by $2^{-\text{size}}$. The
correct importance ratio is the $P^F$ prefix mass divided by the actual
history-inclusion probability. The latter includes equal length and
complexity-tercile quotas, probability-proportional sampling without
replacement, duplicate rejection and forced twins; moreover it is zero for
output-length strata outside 3--8 and for unqueried branches. Appendix
\ref{app:priorfree} gives the exact distinction.
Likewise, PF-5 cannot be reconstructed from the current induction cache:
those calls repeatedly predict one held-out query, whereas PF-5 must predict
each next demonstration before revealing it. The audit finds 3,422 distinct
prequential prompt keys but only two matches in representative complete
caches, so new model calls really are required. We do not require longer
sequences or new calls for the finite-horizon conclusions reported here.

\subsection{Models cluster in distribution space by
family}\label{sec:modelspace}

\begin{figure}[!htbp]
  \centering
  \includegraphics[width=\linewidth]{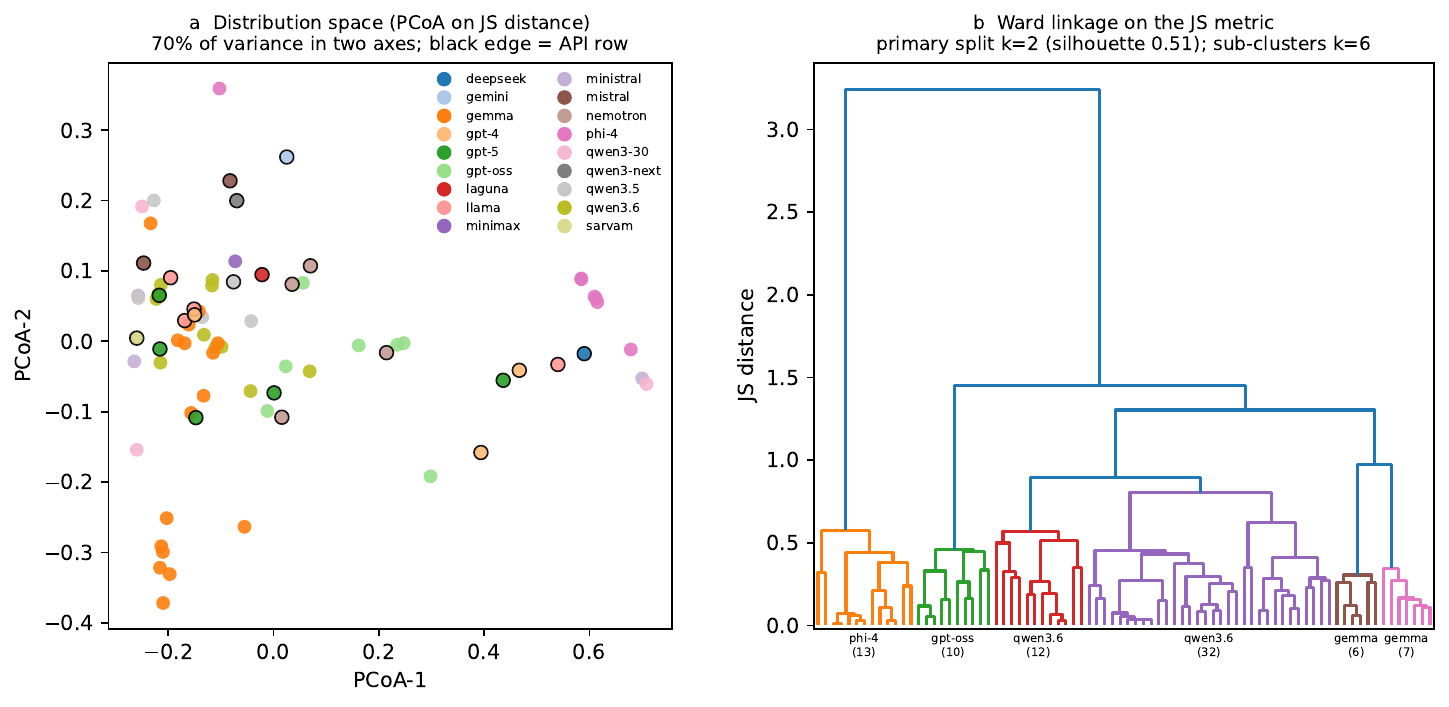}
  \caption{\textbf{The panel in distribution space.} \textbf{a}, Principal
  coordinate analysis of the pairwise Jensen--Shannon distance between the
  served continuation distributions of the $80$ runs with usable
  read-outs (two axes carry $70\%$ of the positive eigenvalue mass);
  colour is model family, black edges mark API rows. \textbf{b}, Ward
  linkage on the same metric: the silhouette-optimal split is $k=2$
  (silhouette $0.51$), and the finer $k=6$ cut recovers quantisation
  ladders as clusters (labels give each sub-cluster's dominant family and
  size). The dominant split separates runs by how they spread mass rather than
  by how well they score, and replicas of one model remain adjacent; the
  matched-ensemble decomposition is reported in the text.}
  \label{fig:modelspace}
\end{figure}

The panel's runs are not just points on a score scale: each one is a
\emph{function} from context to distribution, so the served distributions
themselves define a geometry. For the $80$ runs with usable continuation
read-outs we computed the pairwise mean Jensen--Shannon \emph{distance}
(the square root of the divergence, which unlike the divergence satisfies
the triangle inequality, so the embedding, linkage and silhouette
computations below are metric-valid) over their shared
$(\text{task},\text{position})$ cells, and embedded the resulting matrix by
principal coordinate analysis (Fig.~\ref{fig:modelspace}a). Classical PCoA
is used rather than a neighbour-embedding method such as $t$-SNE
deliberately: our claims are about \emph{global} distances, how far
apart two families sit, which neighbour embeddings distort by
construction. Two axes carry $70\%$ of the positive eigenvalue mass with a
non-Euclidean residue of $2\%$, so the picture is a faithful one. The
geometry passes the obvious sanity check: quantisation replicas of one
model sit at mean distance $0.20$ from each other against $0.51$ between
different models, a factor of $2.5$.

Structure appears at two scales. The silhouette-optimal partition is a
single clean split ($k=2$, silhouette $0.51$) that separates $13$ runs
from the remaining $67$, and it is a split in \emph{style}, not in
quality: the two groups have indistinguishable answer accuracy ($0.617$
versus $0.616$) and similar posterior fidelity ($0.40$ versus $0.37$).
They differ in how they spread probability mass, not in how well they
score. Cutting finer ($k=6$, Fig.~\ref{fig:modelspace}b) recovers the
quantisation ladders as clusters in their own right (phi-4's, gpt-oss's,
gemma-4's and Qwen's ladders each hold together), confirming at the
level of served distributions what the score ladders showed pointwise:
compression barely moves a model. A permutational multivariate analysis of
variance on the distance matrix confirms that model family explains
distributional position ($F=3.93$, $p=0.0002$, $4{,}999$ permutations),
whereas the API-versus-local contrast returns no detectable effect
($F=1.02$, $p=0.33$). That second contrast is \emph{not identified} as a
test of the serving route: no model in this panel is served both ways with
a usable distributional read-out (\S\ref{sec:eval-panel}), so route is
perfectly confounded with model identity and the family factor above
absorbs it. We therefore report the null as uninformative about serving
rather than as evidence that the geometry is free of it. What does bound
measurement noise is the within-model quantisation replica above, which
places replicas of one model adjacent under the same read-out path; and
serving does change greedy \emph{answers} substantially (above), so the two
read-outs are affected differently. Position is only weakly tied to
score: Mantel tests relate distributional distance to differences in
fidelity ($\rho=0.27$, $p=0.001$) and in accuracy ($\rho=0.12$,
$p=0.008$, $n=80$), both significant but far from determinative. Models can be
distributionally remote and still score alike.

That last fact is what makes the panel's aggregate behaviour intelligible.
Clustering the API rows by answer disagreement and drawing matched
four-model ensembles, a vote among models from a \emph{single} cluster
scores $0.60$, while a vote among models drawn from \emph{different}
clusters scores $0.76$. The ensemble
advantage reported above is therefore a diversity effect and not a
head-count effect: what the panel supplies collectively is not more of the
same signal but complementary errors. The limit of that logic is equally
clear. Aggregation operates on
\emph{answers}; the deficiency we measure is in \emph{distributions}. A
committee of predictors that individually sit below the keystroke reference
can vote its way to the panel's best accuracy, but no vote converts
sub-reference posteriors into a calibrated one: the ensemble inherits the
axis it was built on.

\subsection{The frontier under an exact lens: no better, and harder to
measure}\label{sec:frontier}

The panel's frontier rows support a three-part finding that the exact
instrument makes visible: frontier releases have not improved on the
fidelity axis, the distributional readout needed to measure it is being
withdrawn, and what remains measurable is increasingly decided by the
serving and safety layers rather than by the models.

\begin{figure}[!htbp]
  \centering
  \includegraphics[width=0.8\linewidth]{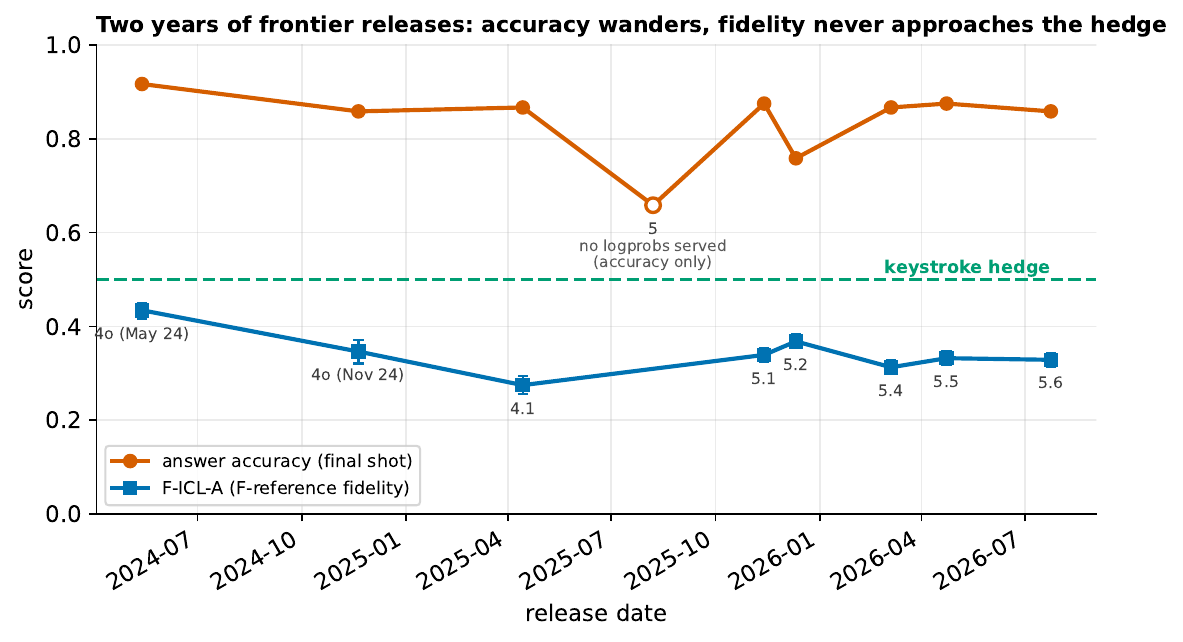}
  \caption{\textbf{Two years of frontier releases at one laboratory.}
  Final-shot answer accuracy (orange) and \Fmac{}-ICL-A posterior fidelity
  (blue) for nine dated OpenAI releases evaluated through the identical
  harness (chat-logprob route for the GPT-4 era, Responses route with
  reasoning effort none for GPT-5.1 and later). Error bars on the fidelity
  series are $95\%$ task-bootstrap CIs (twins resampled with their base
  task; $n=60$ induction and $150$ continuation base tasks, twins resampled with their base). Fidelity is
  highest for the oldest model, and no release approaches the keystroke reference; the
  open marker is the original GPT-5, which exposed no log-probabilities in
  any mode and is therefore accuracy-only.}
  \label{fig:timeladder}
\end{figure}

\paragraph{Fidelity has not improved across two years of frontier releases.} The
per-run ordering behind this comparison is shown in
Fig.~\ref{fig:leaderboard}.
Parameter count is one axis along which capability grows; release date at a
frontier laboratory is another, and it folds in every ingredient of
frontier progress at once (data, architecture, post-training, scale).
Because OpenAI still exposes token log-probabilities (see below), we
evaluated a dated generational ladder of nine models from GPT-4o (May
2024) to GPT-5.6 (July 2026) through the identical harness
(Fig.~\ref{fig:timeladder}). Fidelity does not improve with frontier
time. The oldest model is the best: May-2024 GPT-4o reaches
\Fmac{}-ICL-A $0.43$ (task-bootstrap $95\%$ CI $[0.42, 0.45]$), together
with the highest answer accuracy in the entire panel ($0.92$, later
matched by the two-model Fable-with-fallback stack and by Grok~4.5 at
low reasoning effort, below), and every
later release sits in a $0.27$--$0.37$ band whose CIs never reach $0.39$,
so the separation of the 2024 point from the entire later era is itself
significant ($0.35\to0.27$ over 2024--25, then $0.31$--$0.37$ across the five
GPT-5.x releases), echoing at release timescale the within-family result
that post-training moves models away from the posterior. Two caveats
bound the reading: nine releases support no trend statistic, and the
GPT-4 era is measured over a different serving route (Chat Completions)
than the GPT-5.x era (Responses), a difference we cannot control for, and our
local-versus-API comparison shows that serving route is not generally
neutral; and every frontier row is served with
deliberation disabled or minimised, because log-probabilities are exposed
only in those modes (Methods), so the comparison speaks to the serving
configurations that expose distributions rather than to the systems at full
deliberation. The conservative statement is that two years of frontier
development produced no improvement in fidelity as served through the
distribution-exposing modes. The pattern is not
ours alone: a compression-based test of frontier models over a different
measurement axis independently reports that the newest releases often
regress\cite{superarc2026}, so two unrelated algorithmic-information
instruments agree that recency is not fidelity. Accuracy
meanwhile wanders non-monotonically ($0.66$ at the original GPT-5, $0.76$
at GPT-5.2, near $0.86$ elsewhere). The ladder also carries a serving
datum in its interior: the original GPT-5 release served no
log-probabilities in any mode, a one-generation gap in the distributional
readout that later releases reversed.

\paragraph{The distributional readout is disappearing from frontier APIs.}
A benchmark that scores served distributions depends on providers exposing
them, and direct probes of the four frontier laboratories (July 2026; probe
scripts in the repository) show that this dependency is now precarious.
OpenAI is the only laboratory that still returns token log-probabilities
for its newest models: the GPT-4-era models expose them through Chat
Completions, the GPT-5.1 line and later through the Responses API with
reasoning effort \code{none}, and the original GPT-5 release exposed them
through neither, a one-generation gap later reversed. Google has withdrawn
them at the frontier: on Vertex the native API returns HTTP 400
(``Logprobs is not supported for this model'') for every Gemini generation
after 2.5, while the 2.5 generation, which still serves them and provides
our full Gemini row, has already been retired from the consumer developer
API. Google's OpenAI-compatibility layer strips log-probabilities from
\emph{all} models it hosts, including third-party models whose native APIs
nominally support them. xAI accepts the log-probability parameters in its
request schema and silently returns none for every current Grok model; its
completions endpoint refuses outright (``Raw sampling is not supported for
reasoning models''), and Anthropic's Messages API has never exposed
them (no such parameter exists; probed directly, July 2026). Each withdrawal
has a distinct signature (explicit refusal, silent omission, schema-level
removal), which matters practically: silent omission is exactly the failure
mode our degeneracy audit exists to catch. Distribution-level measurement of frontier models
currently runs through a single provider, benchmark rows can become
unrepeatable when a provider retires a model (the mistral-large episode
of the panel footnote), and open-weight local serving is the only path
whose distributional readout is fully under the evaluator's control. Sampling-based approaches could recover distributional features of these outputs, but they are out of scope here.

\paragraph{The safety layer is part of the serving stack.}
Anthropic's Claude rows make the point most directly. Claude
enters the panel accuracy-only (no log-probabilities, as above); what
the probes additionally revealed is that a \emph{safety classifier} can set a
frontier model's benchmark score, and we measured this as a controlled
three-configuration experiment on identical prompts. \emph{Pure
Fable~5}: the cyber-content filter refuses $16\%$ of our plain
binary-string induction queries (\code{stop\_reason: refusal},
category \code{cyber}; false positives on strings that encode no
code and no instructions), peaking exactly in the ambiguous-evidence
regime ($35$ of $120$ tasks at four examples) and falling to $12$ by
eight; scored as served it reaches $0.89$ final-shot accuracy, but on the
queries it answers it reaches $0.99$, and the comparison survives
matching: restricted to the same $108$ answered cells, May-2024 GPT-4o
scores $0.95$ and GPT-5.6 scores $0.91$, so the margin is not an
artefact of the classifier removing hard cells. \emph{Pure Opus~4.8}: one refusal in
$1{,}080$ cells but far weaker reasoning, $0.74$, so the two
Anthropic failure modes, capability and classifier taxation, are
cleanly dissociated. \emph{Fable~5 with the server-side Opus~4.8
refusal fallback} (the provider's recommended production
configuration): $0.92$ as served, tying May-2024 GPT-4o and Grok~4.5
for the best accuracy in the entire panel (all three at $110/120$).
Yet $4\%$ of cells \emph{still}
return nothing, because on $38$ of those $43$ cells the fallback chain
refuses as well. A same-lab sibling, \code{claude-opus-5}, is stricter
still: its classifier refuses \emph{every} induction query (pilot
$27/27$, confirmed by replay), and a system prompt declaring the
benchmark context does not unblock it, so the model cannot be measured
on this benchmark at all. Two further serving facts fall out of the
experiment. Re-running Fable~5 on byte-identical prompts (the fallback row repeats
every non-refused call; $n=898$ paired calls, hours apart) reproduces
its answers only $86\%$ of the time. The \code{temperature} parameter
is \emph{rejected} on current Claude models (``deprecated for this
model'', HTTP 400), so no client-side setting can remove this
variation; the design cannot separate sampling nondeterminism from
silent server-side updates, and either way the $86\%$ is a floor on
single-run reproducibility for every accuracy-only Claude row. And an offline merge of the two pure rows
(Fable's answers, Opus's on refused cells) predicts the fallback
stack's score to within one task ($0.91$ vs $0.92$), confirming the
in-call fallback behaves as documented. Whether a frontier system can
be benchmarked, and what score it gets, can here be decided by
its safety layer rather than its reasoning: entirely so for
\code{claude-opus-5}, by a sixteen-percent tax for Fable~5, and not at
all for Opus~4.8.

\subsection*{Summary}
Across a heterogeneous panel the findings are consistent. Models select
answers well above chance, but their predictive distributions sit at or
below the keystroke reference, below a frequency lookup table, and at the
level of low-order prefix statistics. A single example makes nearly every
model worse before it gets better; a large share of errors is
inconsistent with the benchmark's support for the shown evidence, and
halting predictions anti-correlate with the truth. None of the levers
that improve conventional benchmarks close the gap: not scale, not two
years of frontier releases, and not post-training, which widens it. Within a task, added evidence narrows the induction gap without closing it, and leaves the termination miscalibration unmoved. Continuation likewise improves late for most scored configurations but remains far from zero; its six-run cache audit consumes substantial pathwise coverage and truth-loss budgets without resolving the finite-horizon impossibility. Two
conditions bound the reading. The gap is distance from the algorithmic
posterior, not a claim that models predict the curated sample poorly: on those
sets they often predict the realised data better than the \Fmac{} reference
does (\S\ref{sec:audit}); and the orderings above are carried by the termination
coordinate, which the bit coordinate does not reproduce
(\S\ref{sec:audit}).

\section{Discussion}

\Fmac{}-ICL asks where a model's in-context predictive distribution lies
relative to one exact, computable algorithmic reference rather than only
relative to other models or human answers. Its defining feature is
computational exactness: the target is the posterior induced by a declared
bounded prior, computed in closed form from an exhaustively enumerated and
independently certified reference machine. Exactness here means that distance
to $P^F$ has no Monte Carlo or baseline-estimation error. It does not make
$P^F$ the uniquely correct posterior for the model's epistemic state.

Three properties make the benchmark practical. It is \emph{exact}: every
target is an exact posterior, not an estimate, so a nonzero gap is a real
distance from a designated posterior rather than baseline noise. That is a
weaker statement than predictive inferiority, and deliberately so: on the
curated evaluation sets the \Fmac{} reference is not itself the best predictor of the
realised tokens (\S\ref{sec:audit}), so the gap measures departure from the
algorithmic posterior, not failure to predict the data. It is \emph{cheap to run}: the
expensive enumeration is frozen once into a self-contained dataset, after
which evaluating a model costs only its own inference. It is also
\emph{difficulty-aware}: tagging tasks by the sample complexity of the
\Fmac{} reference makes it a per-task reference scale, so a model's effective
sample complexity can be compared with the evidence required by that
particular mixture.
The ordering results reported above (scale, generational ladder, post-training)
are measured on the full divergence, which \S\ref{sec:audit} shows is
carried mainly by the termination coordinate; they should be read as
statements about that coordinate rather than about the bit-level
posterior. Across $37$ open models and frontier systems from four
proprietary laboratories the panel returns a consistent result: in-context
answer selection well above chance, yet predictive distributions at or
below the keystroke reference relative to the algorithmic \Fmac{} reference, at a level
indistinguishable from low-order prefix statistics. The framework states
this in exact, reproducible numbers: posterior fidelity to the \Fmac{} reference
shows no detectable improvement across three orders of magnitude of scale
or across two years of frontier releases in the serving modes that expose distributions, is degraded rather than improved
by post-training, is
invisible to accuracy-only evaluation, and is not delivered by frontier
capability: the most faithful frontier model in the panel is the oldest
one evaluated. Every occurrence of ``fidelity'' in these claims therefore
means fidelity to $P^F$, not context-free rationality or predictive
optimality on the curated sample.

\subsection*{Why compare an uninformed text model with the \Fmac{} reference?}
The comparison is useful as a controlled coordinate, not as a demand that the
model should already believe the \Fmac{} prior. A text model necessarily brings
some default distribution to the underspecified phrase ``a fixed program''.
Holding prompts and tasks fixed while changing the evaluated model reveals how
that default distribution and its subsequent updating align with a loop-bearing
algorithmic mixture. The print-only and low-order Markov references then locate
the same model distributions relative to simpler, named measures. This supports
the structural statement that the panel resembles those simpler measures more
than $P^F$ on these prompts. It does \emph{not} show that a rational text model,
given only the prompt, ought to output $P^F$.

The distinction also separates two experiments that the present paper had
previously allowed to blur. Our minimal-prompt experiment probes the model's
\emph{implicit prior plus updating}. A full-disclosure experiment would probe
whether the model can approximate a specified Bayesian computation after being
told the machine, code, prior, resource bounds and selection law. We ran the
former only. The wording/template ablation is not the latter and cannot be used
as evidence that generator disclosure fails. Reporting this missing control is
more accurate than treating ordinary prompt robustness as an answer to the
epistemic objection.

\subsection*{What survives changing the prior?}
The absolute JS values, anchored scores and most model rankings are
reference-relative. An arbitrary alternative universal machine can change
them, especially at descriptions of only tens of bits where an invariance
constant is not negligible. Our evidence supports a hierarchy rather than a
blanket invariance claim. The $L\le6$--$L\le13$ rescore shows stability to a
budget and weighting change within the \Fmac{} family; curation-reweighting and
bit-only scoring preserve the qualitative below-reference majority while
changing its magnitude and membership; complement twins remove one known
machine asymmetry exactly. None is a direct second-machine posterior test.

The continuation bounds in Appendix~\ref{app:priorfree} establish a different,
coverage-conditional statement. Any fixed alternative mixture that assigns
positive mass to the computable \Fmac{} environment incurs only a finite
pathwise excess-log-loss budget relative to $P^F$, with the ceiling allowed to
depend on that alternative prior; any deterministic Bayesian mixture assigning
positive mass to the realised truth likewise has finite cumulative truth loss.
PF-1 and PF-2 report how large those unknown budgets must already be on the
observed paths for all 80 operational read-outs, with a floor-free sensitivity
and a separate six-cache unsmoothed projection. Because the horizon is finite,
they cannot rule out every larger finite constant, and their residual
positive-floor geometry is a termination-coordinate finding: restricted to the
bit coordinate the dyadic index is $0.419$ and the sign test reverses
(\S\ref{sec:sequential-results}). Thus the paper's strongest
machine-independent result is a testable bounded-budget constraint, reported
per coordinate, not invariance of the \Fmac{} score.

\paragraph{Sampled universal-machine data and exhaustive enumeration.}
The closest neighbour to this construction is the line of work that
meta-trains networks on data generated by a universal Turing
machine\cite{grau2024universal}. The reference machines are close relatives:
their BrainPhoque is Brainfuck-equivalent and \Fmac{} is a minimal
Turing-complete tape automaton in the same family, so the universality
argument is common to both, and in both cases it is a property of the
unbounded machine rather than of the instrument actually run. Their
instrument is bounded at $1{,}000$ steps, $200$ memory cells and $256$ output
symbols; ours at $1{,}024$ steps, $128$ cells and $L\le13$. Neither instrument samples
or enumerates anything universal, a bound we state plainly for our own budget
above.

The difference that matters is therefore not fidelity of the machine but
sampling versus exhaustive enumeration, and it decides what can be measured.
Sampling yields an unlimited training stream, which is exactly what
meta-training needs, but it leaves the Solomonoff target out of reach: the
optimal baseline in that work is by necessity a (rather loose, but
non-trivial) upper bound on the log-loss computed from the length of the
shortened generating program, which presumes knowledge of the program that
produced the sequence and bounds a loss rather than supplying a posterior.
Sampling also perturbs the object it approximates, since imbalanced brackets
are skipped to keep every sampled program valid, which the authors note
changes the program distribution. Exhaustive enumeration gives up unlimited
data and program depth (a sampled program may run to the step budget, whereas
ours is capped at thirteen instructions), and buys the posterior
itself in closed form, together with three properties that follow only from
completeness: a difficulty scale from Bayes sample complexity, an exact
accounting of the missing mass rather than its absorption into a normaliser,
and the exact removal of a known machine bias by symmetrisation. The two
designs are complementary rather than competing: one is built to train a
predictor, the other to score one.

One empirical point of contact deserves recording, because it is not
specific to either design. The probability gap left by non-halting programs
is handled there by adopting the normalised Solomonoff prior and filling the
gap with absorbing tokens; here the same phenomenon surfaces as the
termination coordinate, which carries $89\%$ of the continuation divergence
(\S\ref{sec:audit}) and which we therefore report separately from the bit
coordinate. Sequence termination is the dominant nuisance coordinate in both
constructions, and we suggest that any comparison of a served distribution
against an algorithmic target report it separately rather than folded into a
single divergence.

\subsection*{Limitations}
The construction has clear boundaries, which are part of its design. The
bounded prior at this budget reaches counting, arithmetic and bounded
recursion, not the full richness of universal computation; \Fmac{}-ICL is a
\emph{clean} probe of algorithmic reasoning at small description length, not
a proxy for open-ended reasoning. The target is the posterior conditional on
the \Fmac{} prior; because the tested tasks are subsequently selected, it is
not the Bayes predictor for their evaluation law, and another reference
machine would produce different numbers. The symmetrisation
\Fmac{}$\to$\sFmac{} shows that a known structural bias of the reference
machine can be removed \emph{exactly} rather than merely noted.
Continuation, as we noted, separates posterior fidelity well but the chosen foils
weakly; abduction is the sharper separator, and evaluating models on it
awaits a harness that can elicit a distribution over candidate inputs
(today's serving interfaces expose next-token log-probabilities, which
serve induction and continuation directly).

On the measurement side, scores
are conditional on a prompt format, which we hold fixed across every row
rather than eliminate; a five-model prompt ablation (wording, templating, the
grammar constraint, example order and the terminator glyph, plus a
chain-of-thought arm) is released with the code and discussed in
\S\ref{sec:audit}; it is run on the pipeline-scale build the panel uses, and finds
wording, the grammar constraint and example order immaterial while a
template-family change can consume a large fraction of the anchor distance,
mostly by degrading the read-out rather than the model, and models served through commercial
APIs are measured as served: their quantisation and decoding stack are not
observable, so API rows are annotated and, where log-probabilities are not
exposed, restricted to accuracy facets. This restriction is tightening: our
provider probes (\S\ref{sec:eval-panel}) find token log-probabilities
withdrawn or absent at the frontier for every laboratory except OpenAI, so
the distributional facets of future panels may be measurable only for
open-weight models served locally.

The distributional readout also inherits a tokenisation confound\cite{pimentel2024word,oh2024whitespace}: the
harness folds next-token log-probabilities over the exact tokens \code{0},
\code{1} and the terminator, so mass that a vocabulary routes through merged
multi-digit tokens is not credited, by an amount that can differ across
tokenisers; the accuracy facets are immune by construction. A
first-character folding readout that recovers merged-token mass bounds the
effect directly: recomputing the continuation family under both foldings
leaves the scores statistically unchanged (GPT-4o Nov-2024: mean JS $0.399$
exact vs $0.402$ first-character; GPT-5.5: $0.481$ vs $0.480$; Gemini 2.5 Pro,
the panel's sole non-OpenAI frontier fidelity row and a distinct
SentencePiece-family tokeniser: $0.334$ vs $0.338$, anchored score
$0.398$ vs $0.395$; Llama-3.1-70B, a third vocabulary family, $0.408$ vs
$0.413$ over all $1{,}736$ cells; every difference is an order of
magnitude smaller than the divergences being compared, and the largest of
them, Llama's, has a task-bootstrap $95\%$ CI of
$[+0.004,+0.008]$ bits against a panel range of $0.29$--$0.51$), so the
sub-anchor result is not a token-boundary artefact for the four tokeniser
families validated,
including the digit-merging o200k family that motivated the concern; the
induction readout is immune by construction, since candidates are
teacher-forced bit by bit.

Training-set contamination is
unlikely for the current frozen tasks (the realised strings are new), but
the machine and dataset are public, so future training runs could target
them; the dataset is versioned and the pipeline regenerates fresh task sets
cheaply, which is the standing remedy. Finally, the dumped program joint at full length is
storage- and bandwidth-bound: the index for $L=13$, $M=8$ is large, and its
grouping is a (vectorised) host-side \code{lexsort} after transfer.
Regenerating dumps at still larger budgets would therefore benefit from
moving the output keying onto the GPU, a planned extension; the present index already
serves the complete $L\le13$ budget for every posterior we evaluate.

Three boundaries of interpretation deserve to be stated explicitly. First,
the \Fmac{} reference is conditional on conventions the prompt never discloses: the
reference machine, the length budget $L\le13$, the $1024$-step and
$128$-cell bounds, the halting normalisation, the Levin-style
halt-time weighting and the \sFmac{} polarity mixture
(Table~\ref{tab:conventions}). A model is therefore not asked to be
Bayesian with respect to a prior it has been shown; distance to the
\Fmac{} reference confounds inferential competence with knowledge of an unshared
convention. Whether disclosing the machine, prior, budget and curation law
in-context closes any part of the gap is untested here and is the first
control we intend to run.
The bracketing of \S\ref{sec:eval-panel} is the reading least exposed to
this concern, since it compares the panel against two \emph{named} mixtures
rather than against an absolute distance.

Second, the terminator glyph is not a neutral implementation choice.
Replacing the full stop with \code{x} moves continuation divergence by up to
$0.24$ bits in four of five models (\S\ref{sec:audit}), a swing comparable
to the panel's own spread. The direction is toward the uniform read-out
floor of $0.189$, which is the behaviour the degeneracy signature is
designed to detect and which we exclude row by row; but we have not re-served
the full panel under a non-terminator-adjacent glyph, so the headline count
is established under one terminator convention rather than shown invariant
to it.

Third, no evidence-only predictor in this work exceeds the keystroke
reference: the Krichevsky--Trofimov estimators reach $0.41$ and $0.45$,
while the two nulls that do exceed it are constructed from the frozen
\Fmac{} references. The benchmark is therefore demonstrated to be
\emph{discriminative}, so it separates the declared reference update from pattern completion
(\S\ref{sec:foils}) and orders predictors sensibly, but we do not
exhibit a system without access to the target that attains a high score.
Establishing that such a system exists, for instance a network meta-trained
toward the universal predictor, is a natural next step and would convert
the present comparative claim into an absolute one.

The most direct extension is the prequential induction audit (PF-5) that
Appendix~\ref{app:priorfree} specifies but this paper leaves as future work.
The pathwise and truth-loss theorems require a genuine reveal, so they apply
to continuation but not to the present held-out induction protocol, in which
the same query is re-asked as demonstrations accumulate and its target is
never revealed. Scoring each next demonstration before revealing it would
make induction prequential and bring it under
Eqs.~\eqref{eq:pathregret}--\eqref{eq:truthdiagnostics}, supplying for
induction the reference-machine robustness that repeated held-out probes
cannot. The exact \Fmac{} targets are available from the same
program-set backend, but the model calls are not: the audit finds $3{,}422$
distinct prequential prompt keys with only two matches in representative
complete caches, so the missing probabilities cannot be recovered by
rearranging existing responses and the audit needs a fresh evaluation pass.
Until it is run, the observed solved-set instability stands as
a behavioural finding about held-out updating, not as a distribution-free
impossibility result.

Several further directions follow naturally. The same pipeline supports larger
length budgets and alternative reference machines, allowing the difficulty
ceiling to be raised or the prior to be varied as a robustness study, and
connecting the difficulty scale to the length-generalisation behaviour of
transformers on algorithmic tasks\cite{anil2022length,zhou2024rasp}. The
per-token and per-example divergence curves are a fine-grained behavioural
signal that could be related to model scale, training data and inference
strategy, and the certified witness programs make every task
\emph{exhibitable}: one can show the exact program a model failed to infer.
Because the \Fmac{} reference is explicit, \Fmac{}-ICL can also serve as a target for
training, not only evaluation, in the spirit of neural program induction
and synthesis\cite{graves2014ntm,ellis2021dreamcoder}. By giving in-context
learning an exact declared reference, \Fmac{}-ICL offers a precise,
reproducible instrument for tracking one form of algorithmic alignment as
models improve.

\section{Methods}

\small

\subsection{The reference machine \Fmac{}}
A program is a length-$L$ string over the five instructions of
Table~\ref{tab:instructions}, indexed as a base-5 integer
$\mathrm{rule}\in[0,5^{L})$. The tape has a fixed width (default $128$ bits)
with the head initialised at the centre; the input $x$ is written so that
$x[0]$ occupies the centre cell and $x$ extends rightward. Execution reads
the bit under the head, applies the instruction, and updates the program
counter; $\code{[}$ jumps to the instruction past its matching $\code{]}$
when the current bit is $0$, and $\code{]}$ jumps to the instruction past
its matching $\code{[}$ when the current bit is $1$, so
$\code{[}\dots\code{]}$ is a while-bit-set loop. A run \emph{halts} when the
program counter advances past the last instruction; moving the head off
either tape edge aborts the run (out-of-bounds); a step budget (default
$1024$) bounds run time; and, for non-halting detection only, a repeated
full machine state $(\text{counter},\text{head},\text{tape})$ certifies a
cycle. Bracket matching is the sole validity test; invalid (mismatched)
strings are excluded from the enumeration universe, and the non-halting
outcome $\NONHALT$ is reserved for \emph{valid} programs that exhaust the
step budget or provably cycle. The output of a halting run is the bit
string on the head-excursion window $[\min_{\text{head}},\max_{\text{head}}]$;
the empty string is a valid output, and non-halting runs are assigned the
sentinel $\NONHALT$. With an empty input this convention reproduces the
blank-tape behaviour exactly, which anchors the conditional pipeline to the
unconditional one. A step-by-step trace is given in Appendix~\ref{app:trace}.

\subsection{Bounded program prior and algorithmic probability}
A length-$L$ program is encoded in a halt-terminated six-symbol code (five
instructions plus a halt symbol, read until halt), so its description length
in bits is
\begin{equation}
  \ell(L)=\big\lceil (L{+}1)\,\log_2 6\big\rceil ,
  \label{eq:plen}
\end{equation}
which is constant across the $5^{L}$ programs of one length. The bounded,
Solomonoff-style program prior weights a program $p$ by
\begin{equation}
  w(p)=2^{-\ell(|p|)},
  \label{eq:prior}
\end{equation}
restricted to a bounded sub-probability measure by the length budget
$L_{\max}$ (the semimeasure deficiency proper comes from non-halting; both
deficits are quantified below); shorter
programs dominate, with the per-length prior mass $5^{L}2^{-\ell(L)}$ falling
geometrically in the mean (the ceiling in Eq.~\eqref{eq:plen} makes it
non-monotone from length to length; Appendix~\ref{app:plen}) and falling
geometrically.
Enumerating every program of length $1..L_{\max}$ on a fixed ordered set of
inputs and recording each program's \emph{behaviour}
$b=\big(b(x_1),\dots,b(x_K)\big)$ yields the map
\begin{equation}
  b\;\longmapsto\;\textstyle\sum_{p:\,\mathrm{beh}(p)=b} w(p),
  \label{eq:table}
\end{equation}
the total prior mass of each distinct behaviour, the sufficient statistic on
which every \Fmac{} reference in this paper is computed. Halt times are coded with the Elias-$\delta$ code (worked example in
Appendix~\ref{app:elias}), whose
bit-length we write $\delta(\cdot)$; with $\beta(n)=\lfloor\log_2 n\rfloor+1$ the
number of bits in $n$, and Elias-$\gamma$ bit-length
$\gamma(n)=2\beta(n)-1$,
\begin{equation}
  \gamma(n)=2\lfloor\log_2 n\rfloor+1,\qquad
  \delta(n)=\gamma\!\big(\beta(n)\big)+\beta(n)-1=2\lfloor\log_2 \beta(n)\rfloor+\beta(n).
  \label{eq:elias}
\end{equation}

For each input $x$ and output $s$ the pipeline records both a length-only
mass (Solomonoff weighting) and a time-penalised algorithmic probability
(Levin weighting), and the cheapest certified description length:
\begin{align}
  m(s\mid x)      &= \sum_{p:\,p(x)=s} 2^{-\ell(|p|)}, \label{eq:mass}\\
  \hat m(s\mid x) &= \sum_{p:\,p(x)=s} 2^{-\ell(|p|)-\delta(t_p)}, \label{eq:masst}\\
  \hat K(s\mid x) &= \min_{p:\,p(x)=s}\big[\ell(|p|)+\delta(t_p)\big],
  \label{eq:kcost}
\end{align}
where $t_p$ is the halt time of $p$ on $x$ and the witness attaining
\eqref{eq:kcost} is stored. The normalised conditional algorithmic
probability used by the marginal families is
\begin{equation}
  P(s\mid x)=\frac{\hat m(s\mid x)}{Z(x)},\qquad
  Z(x)=\sum_{s'}\hat m(s'\mid x),
  \label{eq:condprob}
\end{equation}
the sum running over halting outputs. All served \Fmac{} references are computed over
valid programs only, with an explicit per-family weighting
(Table~\ref{tab:conventions}). The induction family uses the length-only
weights of Eq.~\eqref{eq:mass} (Solomonoff): its posterior is served by
program-\emph{set} intersection (Eq.~\eqref{eq:serve}), where length-only
weighting reduces serving to exact index arithmetic; this is what makes
the full length budget servable with no truncation anywhere,
$0$-shot included. The accumulating-evidence families use the
time-penalised weights of Eq.~\eqref{eq:masst} (Levin
weighting\cite{levin1973,schmidhuber2002speed}), which the enumeration
stores per (input, output). The weighting convention is explicit and fixed per family
(Table~\ref{tab:conventions}). The cross-check below
compares program \emph{counts} $\#\{p:p(x)=s\}$, which are
weighting-independent and so certify the machine under either choice. Worked
numerical values for Eqs.~\eqref{eq:plen}--\eqref{eq:condprob} are given in
Appendices~\ref{app:plen}--\ref{app:cont}.

\begin{table}[t]
\centering
\caption{\textbf{Serving conventions of the frozen benchmark.} All
families use \sFmac{} \Fmac{} references (Eq.~\eqref{eq:sfid}; one polarity bit, then
the \Fmac{} program) over valid programs only. Induction is served by
program-set intersection (Eq.~\eqref{eq:serve}), which is weighted by
program length alone (Solomonoff) and therefore serves the \emph{full}
length budget exactly, with no truncation; the marginal families
use the time-penalised (Levin) weights of Eq.~\eqref{eq:masst}, whose halt
certificates the enumeration stores. Every induction and continuation task
is emitted in both polarities; abduction is a validation family (foil
separation), not part of the model score.}
\label{tab:conventions}
\small
\begin{tabular}{@{}llllll@{}}
\toprule
Family & Backend & Lengths & Weights & Role & Twins\\
\midrule
Induction & program index (joint) & $1..13$ (never truncated) & $2^{-\ell(L)}$ & scored & yes\\
Continuation & full-budget conditionals & $1..13$ & $2^{-\ell(L)-\delta(t)}$ & scored & yes\\
Abduction & full-budget conditionals & $1..13$ & $2^{-\ell(L)-\delta(t)}$ & validation & --\\
\bottomrule
\end{tabular}
\end{table}

\subsection{The symmetrised machine \sFmac{}, twin tasks and the polarity
gap}\label{sec:methods-sf}
A worked numerical example is given in Appendix~\ref{app:sf}.

An \sFmac{} program is a polarity bit $\beta$ followed by an \Fmac{}
program $p$; the machine reads $\beta$, runs $p$ on the input $x$ if
$\beta=0$ or on the complemented input $\bar x$ if $\beta=1$, and
complements the output in the latter case ($\NONHALT$ maps to itself). Its
description length is $1+\ell(|p|)$ bits (plus the halt-time code where
charged), and its conditional algorithmic probability satisfies the exact
identity
\begin{equation}
  \sP(s\mid x)=\tfrac12\big[P(s\mid x)+P(\bar s\mid \bar x)\big],
  \label{eq:sfid}
\end{equation}
each branch carrying the polarity bit's factor $\tfrac12$. \sFmac{} is an
ordinary prefix machine, so Eq.~\eqref{eq:sfid} is a change of reference
machine, not an ad-hoc average; it is complement-invariant by construction
($\sP(s\mid x)=\sP(\bar s\mid\bar x)$, and the first-bit law equals
$\tfrac12$ exactly), and the certified minimum tightens to
$\hat K_{\sFmac}(s\mid x)=\min\big(\hat K(s\mid x),\hat K(\bar s\mid\bar
x)\big)+1$, never more than one bit above the raw machine's, and often
below it. Legitimacy, exact invariance and this tightening are proved in
\S\ref{app:sfproofs}; the companion paper\cite{cdm2026} builds its
estimator on the same machine, so using \sFmac{} here keeps the series'
reference machine uniform.

For few-shot posteriors the polarity bit flips the \emph{whole} task: an
\sFmac{} program $\beta p$ is consistent with an example $(x_i,o_i)$ iff
$p(x_i)=o_i$ when $\beta=0$, or $p(\bar x_i)=\bar o_i$ when $\beta=1$.
Both routes are served from the same \Fmac{} enumeration (route $\beta=1$
performs the identical lookups at the complemented keys and complements the
outcome), so the \sFmac{} posterior costs two \Fmac{} queries, each at
weight $\tfrac12$:
\begin{equation}
  \sP(s\mid x_q,\mathcal{E})
  =\tfrac12\,P(s\mid x_q,\mathcal{E})
  +\tfrac12\,P(\bar s\mid \bar x_q,\bar{\mathcal{E}}),
  \qquad
  \bar{\mathcal{E}}=\{(\bar x_i,\bar o_i)\},
  \label{eq:sfpost}
\end{equation}
normalised over outcomes. Under Eq.~\eqref{eq:sfpost} the posterior for a
complemented task is exactly the complemented posterior for the original.

This invariance is enforced, not assumed. Every induction and continuation
task is emitted in both polarities (twin ids share the base id), and the
build asserts, for every pair: the twin's stored \Fmac{} reference equals the flipped
original to $10^{-9}$; every stored distribution normalises; and across
paired first positions $P(\code0)=\tfrac12$ exactly. The evaluation then
scores a model on the union of both polarities and reports the polarity gap
as a facet,
\begin{equation}
  \Delta_{\mathrm{pol}}
  \;=\;
  \mathrm{FICL\text{-}A}\big(\text{original tasks}\big)
  \;-\;
  \mathrm{FICL\text{-}A}\big(\text{twin tasks}\big),
  \label{eq:polgap}
\end{equation}
the score difference between original and twin tasks. For the \Fmac{} reference, and for any
model that treats $\code0$ and $\code1$ equivalently,
$\Delta_{\mathrm{pol}}=0$; a deviation measures the model's own polarity
preference on an instrument that provably has none.

\subsection{GPU enumeration pipeline}
The enumeration is implemented in Taichi/CUDA. The unconditional pass
streams every length-$L$ program through the machine on a blank tape and
reduces, per unique output, the count, halt-time statistics, minimal
certified cost and prior masses, using a GPU hash aggregator with an exact
minimum-rule tie-break so that the recorded witness is deterministic. The
conditional pass performs the same enumeration with the tape preloaded with
each input $x$ and aggregates outputs \emph{per input}; it reuses the
validated aggregator verbatim, overriding only the tape-initialisation and
output-span steps. The reduction is associative, which we exploit for
parallelism: the multi-GPU unconditional driver stripes rule batches across
devices and merges the partial dictionaries through a single coalescing path
that preserves the minimum-rule tie-break, while the conditional driver
parallelises over \emph{inputs} (whose per-input HDF5 groups are
independent, so multi-device results are stitched by a pure group copy with
no array-level merge). A dense-prefix copy moves only the halting rows
across the PCIe bus each batch, which is what makes $L=13$ feasible. A small
micro-benchmark detects power-of-two tape-stride partition camping on the
target GPU and selects an odd tape padding when needed.

\subsection{Budget adequacy, merge and certification}
Because a fixed step and tape budget can silently drop programs that halt
only after more steps or a wider head excursion, a coverage scan, coupled
bit-for-bit to the simulator's semantics and using Brent cycle
detection, captures, per input, the exact set of halts the base budget
would miss. These ``outliers'' are re-run at the full budget on a faithful
CPU replica and merged into the base aggregation; per input, the partition
closes exactly (base halts $+$ re-run outliers $=$ coverage halts). The
pipeline orchestrates coverage $\rightarrow$ base $\rightarrow$ merge
$\rightarrow$ copy into a unified result, each stage in its own process so
each obtains a clean CUDA context. Certification (gate~2) samples stored
rows, decodes the stored witness, re-simulates it on the input with exact
CPU semantics and asserts that the halt time and output match the recorded
values; the count cross-check (gate~1) enumerates every length-$L$ program
with the independent reference implementation and asserts per-(input, output)
count equality with the GPU aggregation (verified at $L=5$, $M=8$ across all
$511$ inputs); and the dumped program index (below) is gated against the same
reference machine for per-(input, output) \emph{rule-set} identity (gate~3):
the exact \emph{set} of program identities behind each output, a strictly
stronger check than counts. Exact agreement on these checks is required for a
build to be accepted.

\subsection{The dumped program joint and exact few-shot serving}
The aggregate conditional pass records only per-(input, output) statistics
(counts, witnesses, masses), which yield the marginals $P(\cdot\mid x)$ but
discard \emph{which} program produced \emph{which} output, so they cannot
answer a query conditioned on several different inputs. The dump reuses the
\emph{validated} conditional kernels verbatim but keeps the per-halt
$(\text{rule},\text{output})$ records and writes the full joint as an inverted
index: one HDF5 shard per length $(L,M)$, and within it one group per input
$x$ holding the distinct clean outputs (CSR arrays \code{bit\_length},
\code{out\_word\_ptr}, \code{out\_words}) and, for each output, the complete
\emph{sorted} list of base-5 length-$L$ program identities realising it
(CSR arrays \code{rules\_ptr}, \code{rules}). Sharding by length is essential:
a bare rule index is undecodable without its $L$, and serving weights each
length by the prior $2^{-\ell(L)}$ (Eq.~\eqref{eq:prior}). Outputs are keyed
by the clean head-excursion window (stray packing bits masked), so the
per-output rule \emph{count} equals the marginal file's count exactly, tying
the two backends together.

Given examples $\mathcal{E}$ and a query $x_q$, the exact posterior of
Eq.~\eqref{eq:posterior} is served per length and summed:
\begin{equation}
  P(s\mid x_q,\mathcal{E})=
  \frac{\sum_{L}2^{-\ell(L)}\,\big|\{p\in C_L(\mathcal{E}):p(x_q)=s\}\big|}
       {\sum_{L}2^{-\ell(L)}\,\big|C_L(\mathcal{E})\big|},\qquad
  C_L(\mathcal{E})=\!\!\bigcap_{(x_i,o_i)\in\mathcal{E}^{+}}\!\!\mathrm{rules}_L(x_i,o_i),
  \label{eq:serve}
\end{equation}
where $\mathcal{E}^{+}$ are the positive (halting) examples and
$C_L(\mathcal{E})$ is the consistent set at length $L$, obtained by
sorted-array intersection of their rule lists (smallest first). Negative
examples (an input on which the program did \emph{not} halt) filter
$C_L$ by index membership at that input: a program absent from every output
list of $x_i$ is exactly one that does not halt on $x_i$. The
survivors' outcomes on the query are read from the query input's index by a
single sorted-array membership pass (equivalently, the survivors can be
re-run on the validated machine; the implementation cross-checks the two
routes for bit-exact agreement); survivors absent from the query's index
contribute to the non-halting outcome. Under the length-only weighting no
run-time quantity is needed at serving time, which is what lets every
posterior, at every shot count, mix over the \emph{complete} $L\le13$
budget. Indices are loaded lazily per (length, input) and cached
(example lists as contiguous slices; full query indexes in a bounded pool),
so a query touches only the inputs it mentions and the full joint is never
materialised. The same intersection gives the Bayesian evidence
$\sum_L 2^{-\ell(L)}|C_L(\mathcal{E})|$ (the marginal likelihood). One subtlety:
with \emph{no} positive example the consistent set is the complement of the
halting sets, which the index does not store; the $0$-shot prior is instead
served in closed form from the enumeration's per-length \emph{aggregate}
marginals: per (input, output) halting counts, plus the valid-program count
$\#\mathrm{valid}(L)$ (a bracket-matching dynamic program) for the
non-halting mass, again over the complete budget. The served posterior is bit-identical to the
small-program joint across $0$--$5$ examples (Appendix~\ref{app:index}), and it
is this backend that produces the frozen induction \Fmac{} references in the
evaluation dataset, mixing over the complete $L\le13$ budget with hidden
programs drawn from $L=5$--$9$. The \sFmac{} posterior of Eq.~\eqref{eq:sfpost} is served
by running the identical intersection twice, once on the examples as given
and once on their complements at the complemented query, at weight
$\tfrac12$ each; the index is never duplicated, since complemented inputs
are first-class inputs of the same dump.

Two engineering choices make the dump feasible at $L=13$. A dense-prefix copier
moves only the $\sim\!n_{\text{halts}}$ halting rows across the PCIe bus per
batch (rather than the whole batch buffer), and a single vectorised
\code{lexsort} (primary output bits, finest rule) groups equal-output halts and
sorts their rule lists in one pass, replacing a per-halt loop.

\subsection{Task families}
\textbf{Induction} (joint / index): a hidden program is shown by its
(input $\rightarrow$ output) behaviour on several distinct inputs; the model
predicts its output on a held-out query. Tasks are curated to be
non-degenerate (examples not all identical), learnable (the \Fmac{} reference
posterior converges to the truth by the final example) and non-trivial (the
zero-example \Fmac{} reference places at most $0.7$ on the truth), and each is tagged
with its Bayes sample complexity; the full per-example posterior is stored.
The task set is \emph{difficulty-balanced}: equal quotas per
$\mathrm{BSC}$ bucket (Eq.~\eqref{eq:bsc}), with the hidden program length
spread within each bucket, so free sampling cannot flood the benchmark with
tasks that Occam-guessing already solves.
\textbf{Continuation} (marginals): a program's output for an input is
revealed token by token, with the exact \Fmac{} next-token distribution over
$\{\code{0},\code{1},\textsc{end}\}$ stored at each position; the \Fmac{} reference
uses the full conditional and remains exact. Because the bounded program
prior concentrates on short, simple outputs, free sampling would starve both
the later reveal positions and the high-complexity regime; the task set is
therefore balanced on \emph{two} axes: equal quotas per final output length,
and, within each length, equal quotas per tercile of the task's conditional
codelength $\kappa=-\log_2 P(s\mid x)$ (the \Fmac{} reference's own total log-loss on
the reveal, the continuation analogue of $\mathrm{BSC}$; stored with every
task). Conditioning task \emph{selection} on $(|s|,\kappa)$ does not affect
exactness: each stored distribution remains the exact conditional given
$(x,\text{prefix})$. The hard tercile realises outputs that are correct but
deliberately \emph{not} low-complexity: the reveal walks the model through
atypical branches of the conditional, and the model is scored against the
exact conditional distribution at every step, never on reproducing the
realised bit. \textbf{Abduction}
(marginals): a hidden input from a candidate set produces an output revealed
token by token, with the exact \Fmac{} posterior over inputs computed from
the per-input marginals. Targets are sampled in proportion to prior mass, so
the \Fmac{} predictor's expected log-loss is the information-theoretic floor
under its stated sampling law.
Abduction is used to validate the benchmark's discriminative power
(Fig.~\ref{fig:families}b); model scoring uses induction and continuation
(Table~\ref{tab:conventions}).
Every induction and continuation task is emitted in both polarities
(Methods~\ref{sec:methods-sf}); twin pairs share their curation tags
($\mathrm{BSC}$, $\kappa$), and duplicate rejection during sampling treats a
task and any task's mirrored cell as the same cell, so the two polarity
halves are disjoint and balanced identically.

\subsection{Exact predictors for the accumulating-evidence families}
Write $\mathcal{A}=\{\code{0},\code{1},\textsc{end}\}$, let $r$ be a revealed
output prefix of length $k$, and let $\mathcal{S}_r=\{s:\ r\sqsubseteq s\}$ be
the outputs extending $r$. The \Fmac{} next-token predictor for
\emph{continuation} reads, from the conditional $P(\cdot\mid x)$ of
Eq.~\eqref{eq:condprob},
\begin{equation}
\begin{aligned}
  q_{\code 0}&=\!\!\sum_{\substack{s\in\mathcal{S}_r\\ |s|>k,\ s_{k}=\code0}}\!\!P(s\mid x),\quad
  q_{\code 1}=\!\!\sum_{\substack{s\in\mathcal{S}_r\\ |s|>k,\ s_{k}=\code1}}\!\!P(s\mid x),\quad
  q_{\textsc{end}}=\!\!\sum_{\substack{s\in\mathcal{S}_r\\ |s|=k}}\!\!P(s\mid x),\\[2pt]
  P^{F}_{\mathrm{cont}}(a\mid x,r)&=\frac{q_a}{q_{\code0}+q_{\code1}+q_{\textsc{end}}},\quad a\in\mathcal{A}
\end{aligned}
\label{eq:cont}
\end{equation}
(uniform $1/3$ if the denominator is $0$); this is exactly
$P(\text{next token}\mid x,\ \text{output starts with }r)$. The
\emph{pooled} foil replaces $P(\cdot\mid x)$ in Eq.~\eqref{eq:cont} by the
input-agnostic $\bar P(s)=\tfrac1{|\mathcal{X}|}\sum_{x\in\mathcal{X}}P(s\mid x)$;
the \emph{uniform} foil is $(\tfrac13,\tfrac13,\tfrac13)$.

For \emph{abduction} over a candidate input set $\mathcal{C}$ with uniform
prior $\pi(x)=1/|\mathcal{C}|$, define the prefix mass
$\mu(x,r)=\sum_{s\in\mathcal{S}_r}P(s\mid x)$ (so $\mu(x,\varepsilon)=1$). The
\Fmac{} posterior over the hidden cause is Bayes' rule,
\begin{equation}
  P^{F}_{\mathrm{abd}}(x\mid r)=
  \frac{\pi(x)\,\mu(x,r)}{\sum_{x'\in\mathcal{C}}\pi(x')\,\mu(x',r)} .
  \label{eq:abd}
\end{equation}
The \emph{prior-only} foil returns $\pi$ regardless of $r$; the
\emph{greedy-match} foil places all mass on the candidate $x$ maximising the
pair $\big(\textsc{agree}(\hat s_x,r),\ P(r\mid x)\big)$, where
$\hat s_x=\arg\max_s P(s\mid x)$ and $\textsc{agree}$ counts matching
leading characters, a surface match rather than Bayesian integration.

\subsection{Reference predictors for induction}
All induction predictors return a distribution over the outcome set
$\mathcal{O}$ (distinct outputs plus the non-halting outcome) and are
$\varepsilon$-smoothed, $S_\varepsilon(d)(s)\propto d(s)+\varepsilon$
($\varepsilon=10^{-9}$, renormalised over $\mathcal{O}$), so log-loss and KL
are finite. Given examples $\mathcal{E}_n=\{(x_{i_1},o_1),\dots,(x_{i_n},o_n)\}$
and query $x_q$:
\begin{align}
  \text{F reference:}\quad & S_\varepsilon\!\big(P(\cdot\mid x_q,\mathcal{E}_n)\big)\ \text{from Eq.~\eqref{eq:posterior}},\nonumber\\
  \text{uniform:}\quad & d(s)=1/|\mathcal{O}|,\nonumber\\
  \text{marginal (pattern completion):}\quad & d(s)\propto\big|\{j\le n: o_j=s\}\big|\ \ (\text{uniform if }n{=}0),\label{eq:foils}\\
  \text{nearest-input (surface analogy):}\quad & d=S_\varepsilon(\mathbf{1}_{o_{j^\ast}}),\ \
  j^\ast=\arg\min_{j\le n}\big(d_H(x_{i_j},x_q),\,\big||x_{i_j}|-|x_q|\big|\big),\nonumber
\end{align}
with $d_H$ the Hamming distance (length-padded) and nearest-input uniform at
$n{=}0$. The marginal foil tracks output frequencies while ignoring which
input produced them; nearest-input copies the most input-similar example
without any program inference.

\paragraph{Classical nulls on the panel axes.}
For the model panel we additionally score four evidence-free or
statistics-only predictors through the \emph{identical} pipeline as the
models (same candidate sets, same restricted and renormalised \Fmac{} references, same
JS and anchoring), so they land on the \Fmac{}-ICL-A axis of
Fig.~\ref{fig:gap}: (i) \emph{global output frequencies}, one fixed
distribution, the average of the exact \Fmac{} references over all tasks at zero
examples (induction) and over all tasks and positions (continuation);
(ii) the \emph{pooled \Fmac{} reference}, the per-shot (respectively per-position)
average of the exact \Fmac{} references, the best predictor that ignores the individual
task; (iii, iv) \emph{Krichevsky--Trofimov} add-$\tfrac12$ estimators of
order 0 and 1 fitted on the revealed prefix (continuation) or on the
concatenated example outputs (induction), the standard minimax sequence
predictors over a binary alphabet. Their scores are reported with the panel
(implementations released with the code).

\subsection{Losses, divergences and the \Fmac{}-ICL score}
Worked numerical examples of the divergence and of the anchored score are
given in Appendices~\ref{app:js} and~\ref{app:score}.
For a predictive distribution $p$ and realised outcome $y$ the log-loss is
$\mathrm{LL}(p,y)=-\log_2 p(y)$, the reference loss is $\mathrm{LL}(P^{F},y)$,
and the \emph{regret} is the (non-negative) excess
$R=\mathrm{LL}(p,y)-\mathrm{LL}(P^{F},y)$. We also report
$\mathrm{KL}(p\,\|\,q)=\sum_s p(s)\log_2\!\frac{p(s)}{q(s)}$. The primary
distribution-comparison metric is the base-2 Jensen--Shannon divergence,
\begin{equation}
  \mathrm{JS}(p\,\|\,q)=\tfrac12\mathrm{KL}(p\,\|\,m)+\tfrac12\mathrm{KL}(q\,\|\,m),
  \qquad m=\tfrac12(p+q),
  \label{eq:js}
\end{equation}
which is symmetric, finite without smoothing, and bounded in $[0,1]$. Per
shot/position $n$, every reported curve is the mean over tasks of the
corresponding quantity.

For the headline score, each evaluated family $f$ (here induction and
continuation) contributes its mean JS to the \Fmac{} reference over all of its
$(\text{task},\text{position})$ instances $I_f$,
\begin{equation}
  \bar D_f=\frac{1}{|I_f|}\sum_{i\in I_f}\mathrm{JS}\big(P^{F}_i\,\big\|\,P^{\mathrm{model}}_i\big),
  \label{eq:famjs}
\end{equation}
and the \Fmac{}-ICL score \emph{macro-averages} these over the family set
$\mathcal{F}$ (every family weighted equally, regardless of its task count):
\begin{equation}
  \mathrm{FICL}=1-\frac{1}{|\mathcal{F}|}\sum_{f\in\mathcal{F}}\bar D_f
  =\frac{1}{|\mathcal{F}|}\sum_{f\in\mathcal{F}}\big(1-\bar D_f\big)\ \in[0,1],
  \label{eq:ficl}
\end{equation}
so $\mathrm{FICL}=1$ iff the model matches the \Fmac{} reference on every family, and
the score equals the mean of the per-family facet axes $1-\bar D_f$.

\emph{Keystroke-hedge thresholds.} Emitting uniform random keystrokes over
$\mathcal{A}$ gives a whole output $s$ probability $c(s)=(1/3)^{|s|+1}$. Per
family we report a \emph{hedged} reference $\mathrm{JS}(P^{F},U)$ (against
the uniform next-token / candidate distribution $U$) and a \emph{committed}
reference $\sum_a U(a)\,\mathrm{JS}(P^{F},\mathbf 1_a)$; for induction over
a candidate set the candidate distribution is $c$ renormalised over
$\mathcal{C}$. Both are macro-averaged exactly as in Eq.~\eqref{eq:ficl}
($1-$ mean hedged / committed JS) and reported beside the score, never
blended in.

\emph{The normalised scale.} The raw score of Eq.~\eqref{eq:ficl} must be
read against its keystroke-hedge threshold, not against $0$: Jensen--Shannon
divergence is bounded, and on a three-symbol alphabet the uniform hedge is
never more than
$\mathrm{JS}(\mathbf 1_a\,\|\,U)=\tfrac12\log_2\!\tfrac32+\tfrac16\approx 0.459$
bits from \emph{any} distribution, so $1-\bar D$ compresses every
predictor, informed or not, into a narrow band near the top (the keystroke reference
typically scores $0.85$).  A bounded, task-intrinsic scale follows from
the same geometry: $\mathrm{JS}(P^{F}\,\|\,\cdot)$ is convex on the
simplex, so its maximum is attained at a vertex: the \emph{worst possible
predictor} for an instance is the point mass on the \Fmac{} reference's least likely
symbol (or candidate), and its divergence
$D^{\mathrm{w}}_i=\max_a \mathrm{JS}(P^{F}_i\,\|\,\mathbf 1_a)$ is
computable exactly from the stored \Fmac{} reference.  The headline scale anchors
\emph{three} exactly computable points (the worst predictor, the keystroke reference,
and Bayes) by the piecewise-linear map
\begin{equation}
  \mathrm{FICL\text{-}A}_f=
  \begin{cases}
    \tfrac12+\tfrac12\,\dfrac{\bar D^{\,U}_f-\bar D_f}{\bar D^{\,U}_f}
      & \bar D_f\le \bar D^{\,U}_f\ \ \text{(better than random)},\\[8pt]
    \tfrac12\,\dfrac{\bar D^{\mathrm{w}}_f-\bar D_f}{\bar D^{\mathrm{w}}_f-\bar D^{\,U}_f}
      & \bar D_f> \bar D^{\,U}_f\ \ \text{(worse than random)},
  \end{cases}
  \qquad
  \mathrm{FICL\text{-}A}=\frac{1}{|\mathcal{F}|}\sum_{f\in\mathcal{F}}\mathrm{FICL\text{-}A}_f,
  \label{eq:skill}
\end{equation}
with $\bar D^{\,U}_f$ the family's mean hedged keystroke JS and
$\bar D^{\mathrm{w}}_f$ its mean worst-case JS.  The scale is bounded in
$[0,1]$ by the convexity argument above, and its three anchors are fixed on
\emph{every} dataset: the worst predictor scores $0$, the keystroke reference
scores exactly $\tfrac12$, and Bayes scores $1$.  A single number
therefore carries both facts a reader needs: which side of the keystroke
(print-prior) reference the model sits on (the sign of
$\mathrm{FICL\text{-}A}-\tfrac12$), and how
much of the corresponding gap it closes. Above $\tfrac12$ this is the
fraction of the hedge-to-Bayes gap; below, the position between the worst
predictor and the hedge.  Distances are linear within each regime (the map has a kink at
$\tfrac12$ by construction).  Two diagnostics are retained alongside: the
worst-anchored $1-\bar D_f/\bar D^{\mathrm{w}}_f$ with the hedge reported as
a floating random marker, and the signed
$\mathrm{skill}_f=1-\bar D_f/\bar D^{\,U}_f$ (unbounded below; used for its
sign).

\emph{Difficulty and efficiency.} The Bayes sample complexity of an induction
task is the first shot at which the \Fmac{}-reference MAP equals the truth,
\begin{equation}
  \mathrm{BSC}=\min\big\{n\in\{0,\dots,N\}:\ \arg\max_s P^{F(n)}(s)=\text{truth}\big\}
  \ \ (\,=N\ \text{if never}),
  \label{eq:bsc}
\end{equation}
which makes difficulty explicit. The difficulty-robust \emph{BSC-macro} score
buckets tasks by $\mathrm{BSC}$ (into $\{0\}$, $\{1\}$, $\{2,3\}$,
$\{4{+}\}$), forms a
per-bucket skill $1-\overline{\mathrm{JS}}_{\text{bucket}}$, and averages the
buckets with equal weight. These scoring buckets are deliberately coarser
than the sampling strata: the frozen induction set is balanced $120/120/120/120$
over $\mathrm{BSC}\in\{0\},\{1\},\{2\},\{3{+}\}$, whereas the score pools
$\{2,3\}$ so that each bucket retains enough tasks for a stable per-bucket mean. The \emph{sample-complexity efficiency} compares
the examples a model needs to the Bayes minimum: writing $n^\ast$ for the
first shot at which the model solves the task,
\begin{equation}
  e=\begin{cases}1 & n^\ast=0\\[1pt] 0 & \text{never solved}\\[1pt]
  \min\!\big(1,\ \mathrm{BSC}/n^\ast\big) & \text{otherwise,}\end{cases}
  \qquad \text{SC-eff}=\operatorname{median}_{\text{tasks}} e
  \label{eq:sceff}
\end{equation}
($1$ = as few examples as the \Fmac{} reference). Single-answer induction axes (model answer
$a$, \Fmac{}-reference MAP $s^\ast$, posterior $P^{F(n)}$, each averaged over
$(\text{task},\text{shot})$) are accuracy $\mathbf 1[a{=}\text{truth}]$,
agreement $\mathbf 1[a{=}s^\ast]$, and Bayes-alignment $P^{\star(n)}(a)$.

\emph{The truth--complexity dissociation.} The true output is, by design,
\emph{not} systematically the lowest-complexity consistent answer: whenever
$\mathrm{BSC}>n$, the posterior MAP $s^\ast$ at shot $n$ (the
Occam-favoured, smallest-description candidate) differs from the truth, and
a model that answers $s^\ast$ is wrong on accuracy while performing exactly
the right inference. The metric set is built around this dissociation rather
than suffering from it: accuracy is truth-referenced, agreement and
alignment are Bayes-referenced, and their divergence on high-$\mathrm{BSC}$
tasks \emph{measures} how a model resolves the tension between simplicity
and evidence, which is the object of study rather than an artefact. Two safeguards keep the
comparison fair. First, the curation gates guarantee the tension resolves:
by the final shot the posterior MAP equals the truth, so final-shot accuracy
is a fair end-point accuracy check under the declared reference. Second, the distributional
induction metric scores the model over a candidate set consisting of the
\Fmac{} reference's top $k{-}1$ outputs \emph{plus the true output} ($k{=}6$): without
the truth's inclusion the JS facet would be computed only over the
Occam-favoured candidates and would be blind to the probability mass a model
assigns to a high-complexity truth precisely on the tasks where the
dissociation bites. The set is rebuilt at each shot from that shot's
posterior, so it contracts as evidence eliminates programs: it is the full
$k{=}6$ at $n{=}0$, but a single candidate in $63\%$ of (task, shot) cells
overall and in $93\%$ by the final shot. On those singleton cells the
divergence is identically zero for the model, the keystroke reference and the
worst-case predictor alike, so they dilute all three anchors equally and leave
the ratio the anchored score is built on intact; the induction divergence
facet accordingly carries its signal on the cells where the posterior has not
yet collapsed.

The
facet radar plots all axes on $[0,1]$ (higher is better); its area depends on
the arbitrary axis order and is \emph{not} a metric; ranking uses the scalar anchored
$\mathrm{FICL}$ score. Worked numerical values for
Eqs.~\eqref{eq:cont}--\eqref{eq:ficl} and for Eq.~\eqref{eq:sceff} appear in
Appendix~\ref{app:cont} onward; Appendix~\ref{app:score} also locates that
worked run on the anchored scale of Eq.~\eqref{eq:skill}.

\subsection{Language-model evaluation harness}
The exact prompts every score is read from are reproduced verbatim in
Appendix~\ref{app:prompts}.
Models are queried through an OpenAI-compatible endpoint, so the same code
drives local servers (\code{llama.cpp}, vLLM) and hosted APIs; requests are
submitted concurrently for throughput on batching servers. Few-shot prompts
present examples as \code{input -> output} lines and leave the query open.
For the accuracy metric the answer is constrained by a context-free grammar
to a non-empty bit string, which yields clean tokens and bypasses the
``thinking'' preambles of reasoning models; the grammar constrains
\emph{generation only} and does not affect the logprob-based metrics. The
continuation metric reads the model's next-token log-probabilities over
$\{\code{0},\code{1},\textsc{end}\}$ and compares them to the stored \Fmac{} reference
by Jensen--Shannon. The distributional induction metric teacher-force-scores
the \Fmac{} reference's candidate outputs by sequence log-probability (via
completions-echo where available, or via bit-by-bit chat
log-probabilities) to form a model distribution over candidates, again
scored by Jensen--Shannon. An endpoint probe reports which log-probability
and grammar features a server supports before a run. A prompt-ablation
protocol re-runs a fixed model subset under systematic variations of the
prompt template (instruction wording, example formatting, query framing) to
bound the score's sensitivity to presentation choices; ablation runs are
reported with the panel artefacts.

\subsection{Model panel and controlled comparisons}
Because \Fmac{}-ICL scores a model against a fixed, exact \Fmac{} reference, every
member of an evaluation panel is measured against the same target, and the
frozen \Fmac{} references, prompt format, decoding settings and task sample are held
constant so that only the model varies. We therefore choose the panel not to
maximise leaderboard rank but to isolate specific factors, organised as the
four controlled comparisons below. Whether \Fmac{}-ICL tracks general capability or measures a distinct axis
could be settled by correlating these rows against published scores on
general (e.g.\ MMLU-Pro, GPQA) and reasoning-oriented (e.g.\ AIME,
ARC-AGI) benchmarks; that comparison is not attempted here and is left to
future work. All models are served through the
OpenAI-compatible harness; locally served models are run as quantised weights
on commodity dual-GPU hardware, with the quantisation level recorded against
every result. The panel reported here comprises \textbf{105} runs: \textbf{30}
open models through an NVIDIA NIM endpoint, a nine-model dated OpenAI
ladder (GPT-4o May 2024 through GPT-5.6), Gemini and gemma-4 rows through
Vertex and the Gemini developer API, two Grok rows through the xAI API, three
Claude rows through the Anthropic Messages API (Fable~5, Opus~4.8, and the
Fable-with-Opus-fallback stack), and \textbf{56}
local
\code{llama.cpp} GGUF builds, covering the families GPT, Gemini, Grok, Claude, gpt-oss, phi-4, Qwen3.5,
Qwen3.6, Qwen3-30B, Ministral-3, gemma-4, Nemotron-3, DeepSeek-V4, Mistral,
Llama-3, MiniMax and a tail of small entries (GLM, Step $\times2$, gemma-2, Solar,
Sarvam, Laguna, Inkling, Dracarys, Qwen3-Next, Mixtral), and instantiating
all four controlled comparisons below.
The complete run-level registry (every model, serving route,
precision, usable readouts, billed cost, reasoning-effort setting, and
audit notes) is provided in the Supplementary Information
(\S\ref{app:registry}). The $17$ metered frontier rows (four
providers, including the nine-model generational ladder and the frontier
reasoning systems) were billed at \$$76.70$ in total (per-row costs
in the registry), so the benchmark is reproducible on a modest budget;
the NVIDIA NIM rows were served under a developer account without
per-row metering, and the local panel costs only electricity.

\paragraph{API serving routes.}
Proprietary rows are measured over the exact route that exposes the
required readout, established by per-model probes (July 2026; scripts in
the repository): OpenAI GPT-4-era models through Chat Completions with
\code{logprobs}; GPT-5.1 and later through the Responses API with
reasoning effort \code{none}; the original GPT-5 (accuracy only) through
the Responses API at effort \code{minimal}; Gemini 2.5 through the native
Vertex \code{generateContent} API with \code{response\_logprobs} (the sole
Google route that returns distributions); Gemini 3.x (accuracy only)
through the Gemini developer API or Vertex; Grok (accuracy only) through
the direct xAI API (chat route for the non-reasoning 4.20 row; the
Responses route at low reasoning effort for the 4.5 row; both routes
silently ignore log-probability requests); Claude (accuracy only) through the Anthropic Messages
API. The \code{temperature} parameter is rejected on current Claude
models (``deprecated for this model'', HTTP 400), thinking is bounded with
\code{effort: low} on Fable~5, and the fallback row uses the server-side
refusal-fallback beta with Opus~4.8 as the substitute. Every row records
its route, and per-call disk caches make each run resumable without
re-billing.

\paragraph{Machine dependence, and what invariance does not buy here.}
Algorithmic probability is invariant across universal reference machines
only up to an additive constant \cite{li2019kolmogorov}: for machines $U,V$
there is $c_{UV}$ with $|K_U(s)-K_V(s)|\le c_{UV}$ for all $s$. Which
machine to choose is a substantive and much-discussed problem
\cite{sterkenburg2018}, and the freedom is not innocuous: for any
computable predictor there is a universal prior making it optimal
\cite{leike2015bad}. That guarantee is asymptotic in
description length, and it is worth being explicit that our budget is far
below the regime where it bites. At $L_{\max}=13$ a program costs
$\lceil 14\log_2 6\rceil = 37$ bits, so every program description in this
paper costs at most $37$ bits ($38$ on \sFmac{}, which prepends a polarity
bit; the Levin-weighted families add an Elias-$\delta$ halt-time code, at
most $17$ further bits at the $1024$-step budget). Those totals are the
same order as a plausible translation constant between \Fmac{} and another
machine. The invariance
theorem therefore provides no protection at this scale: the object we
compute is a specific bounded mixture over a specific finite machine, not
an approximation to any machine-independent quantity, and the same is true
of every bounded-enumeration method in this literature. Two consequences
are worth stating plainly. First, absolute \Fmac{}-ICL-A values are
meaningful only relative to the conventions of
Table~\ref{tab:conventions}. Second, the empirical claims that should be
trusted furthest are the ones that do not depend on the target's identity
at all: the non-monotone solved-set instability, the polarity and
complement-equivariance readouts, the post-training ordering, and the
serving results of \S\ref{sec:frontier}.

One piece of external evidence bounds how idiosyncratic the machine is. The
Coding Theorem Method (CTM) estimates algorithmic probability from output
\emph{frequency} over an exhaustively enumerated space of $(n,2)$ Turing
machines \cite{delahaye2012,solertoscano2014}: a different machine family,
a different enumeration, and a frequency-based rather than a length-based
estimator. Over all $4{,}095$ complement classes of binary strings up to
length $12$, the certified description lengths of \sFmac{} agree with
published CTM values at Pearson $0.91$--$0.92$ (Spearman $0.84$--$0.86$),
against $0.79$--$0.81$ (Spearman $0.65$--$0.71$) for the unsymmetrised
machine \cite{cdm2026}, where that comparison is derived in full; it is
posted as a preprint alongside this paper, so the agreement can be checked
independently of anything reported here. Two things follow. The complexity \emph{ordering}
our target is built on is not an artefact of one instruction set, since two
independently constructed machine families rank short strings alike; and the
polarity symmetrisation is externally validated rather than merely
convenient, because it closes most of the gap to the other family.

What that agreement does \emph{not} do is rescore this panel. CTM is a
blank-tape construction and no published conditional CTM table exists, so
there is no CTM-derived posterior $P(s\mid x,\text{examples})$ against
which to score models.

Read together with the budget control of \S\ref{sec:audit}, however, the
two results bracket the objection from both sides. The CTM comparison varies
the \emph{machine} and holds the object fixed: two independently
constructed families, one length-based and one frequency-based, rank short
strings alike. The budget control varies the \emph{object} and holds the
machine fixed: re-serving the conditional posteriors from $L\le6$ through
the complete $L\le13$ store, with a change of weighting convention on top,
moves the per-run ordering by a rank correlation of $0.975$ and leaves the
headline count at $77$ of $80$. So the complexity ordering our target is
built on is robust across machine families, and the conditional posterior
built on that ordering is robust to how the mixture is formed. What neither
result covers, and what we therefore do not claim, is the conjunction: a
conditional posterior served from a \emph{different} family. That is the
natural next test, and it is an enumeration project rather than a rescore: the cached model distributions would carry over unchanged, but the
target would have to be built from scratch.

A further caution for that test: an instruction-symbol \emph{permutation}
is not a change of machine at all under length-only weighting. If $\pi$
relabels the instruction alphabet, then $F_\pi(p)=F(\pi(p))$ and
$p\mapsto\pi(p)$ is a length-preserving bijection of the valid-program set
whenever $\pi$ respects the bracket structure, so the induced output
measure is \emph{identical}: $m_{F_\pi}=m_{F}$. Relabelling tests, and
reflections implemented as a swap of the two move instructions, are
therefore vacuous by construction. A meaningful variation has to change the
input/output convention, the instruction set, or the family.

Relatedly, the enumerated universe is Turing-complete only as an
idealisation: with a $128$-cell tape and a $1024$-step budget the objects
enumerated are a family of finite automata, and universality is a property
of the unbounded machine that motivates the instruction set, not of the
instrument.

\paragraph{Where the missing mass goes.}
Solomonoff's mixture is a \emph{semi}measure because programs may fail to
halt; a bounded enumeration adds further deficits, and all of them are
absorbed into the normaliser $Z$ unless stated. For the frozen datasets
the sources are as follows. Non-halting within the step budget is the only
one present in the unbounded idealisation and the largest here, carrying
$6.2\%$ of the valid-program prior mass over $L\le13$ and rising
monotonically with length from $0\%$ at $L=1$ to $24\%$ at $L=13$. Then
programs beyond $L_{\max}$ (which carry $1.4\%$ of that mass), and the
rounding in the length code $\ell(L)=\lceil (L{+}1)\log_2 6\rceil$, which
also makes the per-length mass non-monotone: the $L{=}14$ shell is
heavier than the $L{=}13$ shell, so the budget cuts immediately before a
heavier one. An exactly normalised code $6^{-(L+1)}$ removes the
non-monotonicity and is preferable in future builds.

The two scored families handle the deficit differently, which we state
because it affects what each measures. Continuation retains the
non-halting mass by mapping it onto the terminator, so its target is a
distribution over $\{\code0,\code1,\textsc{end}\}$ with no deficit.
Induction strips the non-halting outcome and the storage-tail bucket from
the candidate set and renormalises, so its target is conditioned on the
program halting within budget. Induction scores are therefore
halting-conditioned and support-truncated (top-$24$ storage plus a
residual bucket), and the ``zero posterior mass'' errors of
\S\ref{sec:eval-panel} are zero under that truncated support rather than
under the unbounded mixture.

\paragraph{Statistical reporting.}
Hypothesis tests are two-sided unless explicitly labelled directional; the
late-versus-early continuation sign test is one-sided for the stated
improvement direction and is accompanied by a checkpoint-cluster interval.
Local decoding is deterministic
(temperature $0$) so local runs are single-seed; API rows are only as
reproducible as their serving stack (re-running Claude Fable~5 on
byte-identical prompts reproduces $86\%$ of answers,
\S\ref{sec:frontier}), and run-to-run variation is instead
characterised directly through the quantisation replicas and the paired
serving comparison. Analyses whose runs share weights (quantisation
ladders) are clustered by base model before inference: the polarity test
is a $t$-test over the $46$ base-model means, not over the $81$ runs
(the open-panel subset reported alongside uses its own $37$ means).
Confidence intervals are $95\%$ nonparametric bootstraps at the dependence
unit stated for each analysis. The operational continuation threshold bands
cross checkpoint-family and complement-twin clusters and are simultaneous over
the complete displayed threshold grid. Given the panel's exploratory breadth
we designate two findings as confirmatory: the
sub-reference headline ($78$ of $81$ runs and $45$ of $46$ base models; sign
test $p<10^{-9}$) and the
post-training fidelity decline (replicated in two independent families,
with drops an order of magnitude above the bootstrap CI); all other
contrasts, including the polarity-twin bias, are reported descriptively
with exact $n$ and unadjusted $p$-values and should be read as such.

\subsubsection{(a) Same model, varying quantisation}
For a fixed checkpoint we sweep the quantisation level (e.g.\ \textsc{bf16},
\textsc{8-}, \textsc{4-}, \textsc{3-} and \textsc{2-bit}) and measure how each
\Fmac{}-ICL facet moves. This serves two ends. Scientifically, because the
continuation and induction-JS metrics are computed from token
log-probabilities (Eqs.~\eqref{eq:cont},\,\eqref{eq:js}), they are sensitive
to the logit perturbation quantisation introduces; we expect posterior
fidelity (JS to the \Fmac{} reference) to degrade \emph{before} single-answer
accuracy does, a dissociation the facets can resolve; the panel outcome is
reported in \S\ref{sec:eval-panel}. Practically, the sweep
identifies the smallest precision at which a model's score is statistically
indistinguishable from its full-precision score, fixing the most economical
serving setting for the rest of the panel and bounding the bias that low-bit
quantisation injects into the headline number.

\subsubsection{(b) Same family, increasing parameters}
Within one model family we hold the training recipe fixed and increase the
parameter count (a small dense member, a mid-size dense member, and the
largest member that fits, including mixture-of-experts (MoE) variants for
which the total and active parameter counts differ). This isolates the effect
of scale on algorithmic in-context reasoning: whether the per-family mean JS
to the \Fmac{} reference (Eq.~\eqref{eq:famjs}) falls monotonically, whether the Bayes
sample complexity (Eq.~\eqref{eq:bsc}) the model can absorb rises, and whether
sample-complexity efficiency (Eq.~\eqref{eq:sceff}) approaches one. For MoE
members we report both total and active parameters, since the two scale the
relevant computation differently.

\subsubsection{(c) Same model: base, instruction- and reasoning-tuned}
For a family that releases a base (pre-trained-only) checkpoint alongside
instruction-tuned and explicit-reasoning (``thinking'') variants, we compare
the three at matched scale. The base model is a pure next-token predictor and
gives the cleanest test of whether a model's \emph{raw} continuation prior is
aligned with the algorithmic prior, a comparison the log-probability metrics
support directly, even though a base model may not follow the few-shot answer
format (the grammar-constrained answer call handles formatting; the JS metrics
do not depend on it). The reasoning variant tests the opposite hypothesis:
explicit deliberation may raise induction accuracy and lower sample complexity
while \emph{worsening} token-level posterior fidelity, if it sharpens the model's
distributions past the \Fmac{} reference's own uncertainty. The harness exposes the
reasoning mode as a toggle, so the three are evaluated through an otherwise
identical pipeline.

\subsubsection{(d) Different families}
Holding scale roughly constant, we compare models from distinct developers and
pre-training pipelines (for example Qwen, Gemma, GPT-OSS, Nemotron, GLM,
DeepSeek, Mistral and Llama). Because the \Fmac{} reference is recipe-independent, a
difference in the \Fmac{}-ICL score across families at matched size reflects a
difference in what pre-training and alignment instil about algorithmic
structure, not a difference in the yardstick. This is the comparison behind the family
effect of \S\ref{sec:modelspace}. We do not report it here as a
separate matched-size ladder: the family effect it targets is reported
instead in \S\ref{sec:modelspace}, where family explains distributional
position ($F=3.93$, $p=0.0002$). A matched-size cross-family ladder
remains a natural extension.

\emph{Controls.} Across all four comparisons the frozen \Fmac{} references, the curated
task set, the prompt templates, the decoding configuration
(grammar-constrained for accuracy, log-probabilities for the divergence
metrics) and the random seeds are fixed; only the factor under study varies.
Confidence intervals are obtained by resampling tasks (decoding is
deterministic throughout), so a reported gap between two panel members can be
read as significant or not.

\subsection{Reproducibility and the precomputed dataset}
The exact \Fmac{} references are computed once and frozen into a single self-contained
JSON dataset (curated induction tasks with full per-shot posteriors, and
continuation tasks with per-position \Fmac{}-reference next-token distributions), so
model evaluation requires neither the large enumeration dumps nor the joint
computation at run time. All randomness (task sampling, input orderings,
candidate sets) is seeded. The released dataset is the one every panel row is scored on. An
\emph{extended} build is also released, used for the validation figures
and available for future work: it holds
$480$ induction tasks ($240$ curated, each with its polarity twin; $8$ shots;
hidden programs of length $L=5$--$9$, difficulty-balanced into equal
Bayes-sample-complexity quotas) and $300$
continuation tasks ($150$ length-$\times$-complexity-balanced pairs, each
with its twin; up to nine revealed positions, one per output bit plus the
terminator), built with seed $0$; its
induction \Fmac{} references mix over the complete $L\le13$ budget
(Table~\ref{tab:conventions}). The model panel of \S\ref{sec:eval-panel} was
evaluated on the \emph{pipeline-scale build}, which shares the
continuation family byte-for-byte (so every cached continuation call transfers)
and serves a smaller induction family of $120$ tasks ($60$ curated $+$ twins)
from the exact small-program joint (hidden programs $L\le6$, mixture over
$L\le6$); \S\ref{sec:eval-panel} verifies exactly that the two instruments
test models equivalently on these tasks (\Fmac{} reference divergence $0.028$ bits,
run-level rank correlation $0.975$, headline unchanged). The
validation figures use $300$ tasks (Fig.~\ref{fig:foils}) and $500$ tasks
per family (Fig.~\ref{fig:families}), seed $0$; their generating script is
released with the code. Local software: Python 3.12.12, \code{pycdm} 0.1.0,
NumPy 2.4.6, SciPy 1.17.1, Matplotlib 3.10.9, scikit-learn 1.9.0 and
h5py 3.16.0; API clients \code{openai} 2.41.0 and \code{anthropic}
0.120.0. The enumeration and the conditional dumps were produced with
Taichi 1.7.4 on the CUDA backend, on a single NVIDIA GeForce RTX 4090
Laptop GPU ($16$\,GiB) with driver 561.09; local model serving used
\code{llama.cpp} on commodity dual-GPU hardware. Every stochastic step
in the build (task sampling, input orderings, candidate-set selection)
uses seed $0$, and each panel row is a single run: local decoding is
deterministic at temperature $0$, while API rows are only as reproducible
as their serving stack, which we measure rather than assume
(\S\ref{sec:frontier}).

\normalsize

\section{Data availability}
\small
The frozen \Fmac{}-ICL benchmark dataset and the precomputed reference-%
predictor results that support the findings of this study are archived at
\texttt{10.5281/zenodo.0000000} (final DOI to be disclosed) and are also
released with the code (below). The large intermediate enumeration
artefacts can be regenerated from the pipeline.
\normalsize

\section{Code availability}
\small
The \Fmac{}-ICL benchmark, the reference machine and predictors, the GPU
enumeration pipeline, the certification tools and the evaluation harness are
released under the MIT licence at the release tagged for this manuscript;
the repository URL is to be disclosed. The submission bundle also contains a
self-contained \code{codes/} snapshot with frozen panel results, reconstruction
caches, a SHA-256 manifest, and the prior-free audit and plotting commands.
The documented reconstruction and plotting commands regenerate
Figs.~\ref{fig:pf0-all}--\ref{fig:pf2-truth},
Figs.~\ref{fig:pf6-operational}--\ref{fig:pf9-extrapolation}, and
Fig.~\ref{fig:pf4-revision}, together with their machine-readable cell- and
run-level tables, clustered tests, provenance fields, exclusions and scope
report.
\normalsize

\section{Acknowledgements}
\small The authors gratefully acknowledge the support of Oxford Immune
Algorithmics and of King's College London.\normalsize

\section{Author contributions}
\small \textbf{Luan Ozelim:} Conceptualization, Methodology, Software,
Validation, Investigation, Data curation, Formal analysis, Visualization,
Writing -- original draft, Writing -- review \& editing.
\textbf{Hector Zenil:} Conceptualization, Supervision, Funding
acquisition.\normalsize

\section{Competing interests}
\small The authors declare no competing interests.\normalsize

\section{Use of AI tools disclosure statement}

Generative AI tools were used to assist with language editing, LaTeX formatting, checking derivations, proposing objections to arguments, and organising the material. All mathematical arguments, proofs, results, and interpretations originated and were verified by the authors, who assumes full responsibility for the manuscript.

\clearpage
\appendix
\setcounter{equation}{0}
\renewcommand{\theequation}{\Alph{section}.\arabic{equation}}
\small
\part*{Supplementary Information}
\noindent\emph{Supplementary sections \ref{app:trace}--\ref{app:index}
are worked numerical examples.}
\medskip
\section{The symmetrised machine: legitimacy, invariance, tightening}
\label{app:sfproofs}

Write $F$ for the reference machine, $p$ for one of its programs, $\ell(p)$
for its code length, and $\bar{\;\cdot\;}$ for bitwise complement. Define
$\sFmac{}$ as in the main text: read one polarity bit $\beta$, run $F$, and
complement the output iff $\beta=1$. Its programs are the strings $\beta p$,
so $\ell(\beta p)=\ell(p)+1$ and the program set is $F$'s doubled. Three
properties hold exactly rather than asymptotically.

\paragraph{(i) Legitimacy.} $\sFmac{}$ is an ordinary prefix machine and its
prior is a bounded sub-probability measure with the same total mass as $F$'s:
\begin{equation}
\sum_{\beta\in\{0,1\}}\sum_{p} 2^{-\ell(p)-1}
 \;=\; 2\cdot\tfrac12\sum_{p} 2^{-\ell(p)}
 \;=\; \sum_{p} 2^{-\ell(p)} .
\end{equation}
The polarity bit is paid for out of the doubled population, so nothing is
created or destroyed. Consequently every quantity we report on $\sFmac{}$ is
an estimator \emph{computed on a machine}, not a post-hoc average of two
runs of $F$, which matters because a post-hoc average is not the
algorithmic probability of anything.

\paragraph{(ii) Exact invariance.} For any input $x$ and output $s$,
grouping the $\sFmac{}$ programs by their polarity bit gives
\begin{equation}
\sP(s\mid x) \;=\; \tfrac12\bigl[P(s\mid x) + P(\bar s\mid \bar x)\bigr],
\label{eq:sfid-proof}
\end{equation}
since $\beta p$ maps $x\mapsto s$ exactly when $p$ maps $x\mapsto s$
($\beta=0$) or $p$ maps $\bar x\mapsto\bar s$ ($\beta=1$). Substituting
$(\bar s,\bar x)$ for $(s,x)$ in \eqref{eq:sfid-proof} exchanges the two terms and
leaves the sum unchanged, so
\begin{equation}
\sP(s\mid x) \;=\; \sP(\bar s\mid \bar x) \quad\text{identically.}
\end{equation}
The identity is a symmetry of the machine, not a property of any dataset,
and it holds cell by cell rather than on average. Reversal invariance is
inherited from $F$ unchanged, since the polarity bit acts only on output
polarity. The first-position law follows: summing \eqref{eq:sfid-proof} over the
outputs beginning with $\code1$ and over their complements beginning with
$\code0$ gives equal totals, so $P(\code0\text{ first})=\tfrac12$ exactly.

\paragraph{(iii) Tightening.} $\sFmac{}$ admits every description $F$ admits,
via $\beta=0$, \emph{and} the complement route, via $\beta=1$. A one-sided
upper bound on description length can therefore only improve:
\begin{equation}
\hat K_{\sFmac{}}(s\mid x) \;=\;
\min\bigl(\hat K(s\mid x),\, \hat K(\bar s\mid \bar x)\bigr) + 1
\;\le\; \hat K(s\mid x) + 1 ,
\end{equation}
with equality exactly when the complement is not the cheaper face. The $+1$
is the polarity bit, the price of recording which of the two runs
produced $s$, and paying it is what makes the estimator a description
length on a machine rather than an unnormalised pooled count. In the
companion enumeration the complement route supplies the shorter program for
a substantial minority of outputs, so the mean effect of symmetrisation is a
\emph{reduction} in certified description length despite the extra bit.

\input{./tables/prior_free_appendix}
\section{Panel registry}\label{app:registry}
\input{./tables/panel_table}

\section{The prompts}\label{app:prompts}

Every score in this paper is read from a model's response to one of two
prompts, held fixed across all $105$ rows up to the per-row serving wrapper
recorded in the registry. They are reproduced here
verbatim; the ablation of \S\ref{sec:audit} varies exactly these strings.

\paragraph{Induction.} The examples are rendered one per line as
\code{input -> output}, the query line is left open, and the instruction
asks for a bare bit string. Answers are read either as free text or under a
grammar constraint (Methods).

\begin{quote}\small\ttfamily\raggedright
Each line maps an input bit string to a fixed hidden program's output bit
string. Give ONLY the output for the last line: a bit string like 011,
with no words, no code, no explanation. If unsure, still answer with your
best bit string.\\[4pt]
0010 -> 01\\
010 -> 01\\
0000 ->
\end{quote}

\noindent The empty input is written \code{()} so that a zero-length
string is visible rather than blank. At $n=0$ examples the block between
the instruction and the query line is empty, and the model is asked for the
output of an unseen program on one input, which is what makes the
zero-shot column a pure prior read-out.

\paragraph{Continuation.} The next-token family reveals the output one
character at a time and reads the served distribution over
$\{\code0,\code1,\textsc{end}\}$ from the log-probabilities of a
single-token reply. The terminator is the full stop.

\begin{quote}\small\ttfamily\raggedright
A fixed program over bit strings was run on the input below and produced an
output, shown so far. Reply with EXACTLY ONE character: the next output bit
('0' or '1'), or '.' if the output has ended.\\[4pt]
input: 01101111\\
output so far: 1000\\
next character:
\end{quote}

\noindent \code{output so far} is written \code{(empty)} at the first
revealed position. The three classes are recovered from the returned
top-$k$ tokens by exact single-character match, with the
first-character folding described in the Limitations as the robustness variant.

\paragraph{Serving wrappers.} Chat-served models receive the prompt as a
single user turn. Models whose default template emits a reasoning block
before the answer, and base checkpoints with no chat template at all, are
served through \code{/v1/completions} with a fixed wrapper that prefills an
empty, already-closed reasoning block so the first generated token is the
answer. The wrappers are recorded per row in the registry; as an example,
the Qwen family uses
\begin{quote}\small\ttfamily\raggedright
<|im\_start|>user\\
\{prompt\}<|im\_end|>\\
<|im\_start|>assistant\\
<think></think>
\end{quote}
\noindent with the reasoning block opened and closed immediately so the
next generated token is the answer, and the Mistral family the analogous
\code{[INST]\{prompt\}[/INST][THINK][/THINK]}.
Equivalent wrappers exist for the gpt-oss, MiniMax, DeepSeek and GLM
families; all are in the released harness. Substituting the raw wrapper for
a model's chat template is one of the ablation arms, and is the arm that
moves the score most (\S\ref{sec:audit}).

\noindent\emph{The worked examples below use the raw machine \Fmac{} except
where \sFmac{} is itself the subject, and,
where a weighting is needed, the length-only prior: the minimal setting
that keeps every number legible and hand-checkable. The frozen benchmark
serves \sFmac{} \Fmac{} references with time-penalised weights
(Table~\ref{tab:conventions}); the constructions are identical, with the
polarity mixture of Eq.~\eqref{eq:sfid} layered on top.}

\section{Worked example: the reference machine}\label{app:trace}
Program $p=\code{>}\,\code{+}\,\code{<}$ (base-5 index $\mathrm{rule}=35$,
length $L=3$) on input $x=\code{10}$. The head starts at the tape centre $c$,
with $x$ written rightward ($\text{cell}_c{=}\code1$, $\text{cell}_{c+1}{=}\code0$).
Execution halts when the program counter passes the last instruction:

\begin{center}
\begin{tabular}{@{}clccc@{}}
\toprule
step & instruction & head & pc & $(\text{cell}_c,\text{cell}_{c+1})$\\
\midrule
0 (start) & --             & $c$   & 0 & $(\code1,\code0)$\\
1 & \code{>} (right)        & $c{+}1$ & 1 & $(\code1,\code0)$\\
2 & \code{+} (flip)         & $c{+}1$ & 2 & $(\code1,\code1)$\\
3 & \code{<} (left)         & $c$   & 3 $\Rightarrow$ halt & $(\code1,\code1)$\\
\bottomrule
\end{tabular}
\end{center}
The head visited cells $\{c,c{+}1\}$, so the output is the head-excursion
window $\code{11}$, with halt time $t_p=3$ and output length $2$. Mapped over
the input set, this program has behaviour $(\code{01},\code{01},\code{11})$ and
contributes one of the behaviour entries counted in
Appendix~\ref{app:table}.

\section{Worked example: encoding length and prior weight}\label{app:plen}
Equation~\eqref{eq:plen} with $\log_2 6=2.584963\ldots$ gives, and
Eq.~\eqref{eq:prior} the per-program weight $w=2^{-\ell(L)}$; the last column
is the total prior mass of all programs of that length:

\begin{center}
\begin{tabular}{@{}rrrrr@{}}
\toprule
$L$ & $\ell(L)$ (bits) & $2^{-\ell(L)}$ & $5^{L}$ & $5^{L}\,2^{-\ell(L)}$\\
\midrule
1 & 6  & $1.56\times10^{-2}$ & 5      & 0.0781\\
2 & 8  & $3.91\times10^{-3}$ & 25     & 0.0977\\
3 & 11 & $4.88\times10^{-4}$ & 125    & 0.0610\\
4 & 13 & $1.22\times10^{-4}$ & 625    & 0.0763\\
5 & 16 & $1.53\times10^{-5}$ & 3{,}125 & 0.0477\\
6 & 19 & $1.91\times10^{-6}$ & 15{,}625 & 0.0298\\
\bottomrule
\end{tabular}
\end{center}
A single program at $L{=}1$ carries $2^{5}=32\times$ the weight of one at
$L{=}3$; the prior concentrates on short programs, which is why summing
$L=1..L_{\mathrm{eval}}$ is exact to far beyond float precision once any short
consistent program exists.

\section{Worked example: Elias-$\delta$ time penalty}\label{app:elias}
Equation~\eqref{eq:elias} for representative halt times, with the resulting
Levin time weight $2^{-\delta(t)}$:

\begin{center}
\begin{tabular}{@{}rrrrr@{}}
\toprule
$t$ & $b(t)$ & $\gamma(t)$ & $\delta(t)$ & $2^{-\delta(t)}$\\
\midrule
1   & 1  & 1  & 1  & $5.00\times10^{-1}$\\
2   & 2  & 3  & 4  & $6.25\times10^{-2}$\\
4   & 3  & 5  & 5  & $3.13\times10^{-2}$\\
8   & 4  & 7  & 8  & $3.91\times10^{-3}$\\
16  & 5  & 9  & 9  & $1.95\times10^{-3}$\\
1024& 11 & 21 & 17 & $7.63\times10^{-6}$\\
\bottomrule
\end{tabular}
\end{center}
A length-3 witness ($\ell=11$) that halts at $t=4$ ($\delta=5$) therefore has
certified description length $\hat K=\ell+\delta=11+5=16$ bits
(Eq.~\eqref{eq:kcost}) and Levin weight
$2^{-(\ell+\delta)}=2^{-16}\approx1.53\times10^{-5}$ (Eq.~\eqref{eq:masst}).

\section{Worked example: the behaviour table (sufficient statistic)}\label{app:table}
Take $M=1$, $L_{\max}=3$, so the input set is
$(\varepsilon,\code0,\code1)$ and $\sum_{L=1}^{3}5^{L}=155$ programs are
enumerated. They realise $29$ distinct behaviours with total mass
$Z_{\mathrm{tot}}=0.236816$, the total enumerated prior mass including the
non-halting outcome (distinct from the halting-only normaliser $Z(x)$ of
Eq.~\eqref{eq:condprob}). Writing each behaviour as
$b=(b(\varepsilon),b(\code0),b(\code1))$ and using Eqs.~\eqref{eq:table},\,
\eqref{eq:prior}:

\begin{center}
\begin{tabular}{@{}lrr@{}}
\toprule
behaviour $b$ & mass $\sum_{p\vdash b} w(p)$ & \#programs\\
\midrule
$(\bot,\bot,\bot)$ (never halts)        & $1.333\times10^{-1}$ & 106\\
$(\code{00},\code{00},\code{01})$        & $2.148\times10^{-2}$ & 6\\
$(\code1,\code1,\code0)$                  & $1.611\times10^{-2}$ & 2\\
\bottomrule
\end{tabular}
\end{center}
The map behaviour~$\mapsto$~mass is everything the \Fmac{} reference consumes;
the program count column is what the GPU cross-check (gate~1) reproduces
byte-for-byte.

\section{Worked example: the exact induction posterior}\label{app:posterior}
Hidden behaviour $b^\star=(\code{00},\code{00},\code{01})$ from
Appendix~\ref{app:table}; the query is the output on the empty input
(truth $=\code{00}$), and examples are revealed in the order
$(\code0\!\to\!\code{00})$, then $(\code1\!\to\!\code{01})$. The posterior
predictive (Eq.~\eqref{eq:posterior}) and the marginal foil's regret
(Eq.~\eqref{eq:foils}) are:

\begin{center}
\begin{tabular}{@{}clll@{}}
\toprule
$n$ & examples & \Fmac{} $P(\cdot\mid x_q,\mathcal{E}_n)$ (top) & ref.\ LL / marg.\ regret\\
\midrule
0 & --                        & $\bot{:}0.565,\ \code{00}{:}0.186,\ \code1{:}0.070,\dots$ & $2.43$ / $1.66$ bits\\
1 & $\code0\!\to\!\code{00}$  & $\code{00}{:}1.000$                                       & $0.00$ / $0.00$ bits\\
2 & $+\,\code1\!\to\!\code{01}$ & $\code{00}{:}1.000$                                     & $0.00$ / $1.00$ bits\\
\bottomrule
\end{tabular}
\end{center}
One example pins the program (the posterior collapses to $\code{00}$), so this
task has Bayes sample complexity $\mathrm{BSC}=1$ (Eq.~\eqref{eq:bsc}). The
pattern-completion foil, shown two \emph{different} outputs by $n{=}2$, hedges
between them and pays $\approx 1$ bit of regret while the \Fmac{} reference pays none.

\section{Worked example: continuation next-token}\label{app:cont}
This example enumerates programs of length $L\le5$ under raw \Fmac{} with
length-only weights, so every value below is hand-checkable; the full
benchmark uses the Levin-weighted \sFmac{} mixture over $L\le13$. For input
$x=\code{10}$ the conditional marginal $P(\cdot\mid x)$ is tabulated in
Appendix~\ref{app:sf}. Applying Eq.~\eqref{eq:cont} as the prefix $r$ grows:

\begin{center}
\begin{tabular}{@{}lccc@{}}
\toprule
prefix $r$ & $P^F(\code0,\code1,\textsc{end}\mid x,r)$ & entropy (bits)\\
\midrule
$\varepsilon$ & $(0.576,\ 0.424,\ 0.000)$ & 0.983\\
$\code0$       & $(0.329,\ 0.412,\ 0.259)$ & 1.559\\
$\code{01}$    & $(0.081,\ 0.046,\ 0.873)$ & 0.669\\
\bottomrule
\end{tabular}
\end{center}
The distribution sharpens toward \textsc{end} once the prefix completes a
high-mass output. The input-agnostic pooled foil is close here
($\mathrm{JS}(P^F\,\|\,\text{pooled})=0.0058$ bits at $r=\varepsilon$),
illustrating why continuation is a sensitive fidelity probe yet only weakly
separates these foils (cf.\ Fig.~\ref{fig:families}a).

\section{Worked example: the symmetrised machine \sFmac{}}\label{app:sf}
Continuing from Appendix~\ref{app:cont} ($L\le5$, length-only weights,
valid programs): the \sFmac{} conditional for $x=\code{10}$ mixes the raw
conditional at $x=\code{10}$ with the complemented conditional at the
complemented input $\bar x=\code{01}$ (Eq.~\eqref{eq:sfid}):

\begin{center}
\begin{tabular}{@{}lccc@{}}
\toprule
$s$ & $P(s\mid\code{10})$ & $P(\bar s\mid\code{01})$ &
$\sP(s\mid\code{10})=\tfrac12[\,\cdot+\cdot\,]$\\
\midrule
\code{01} & 0.207 & 0.050 & 0.128\\
\code{10} & 0.207 & 0.234 & 0.221\\
\code{0}  & 0.149 & 0.137 & 0.143\\
\code{1}  & 0.036 & 0.094 & 0.065\\
\code{00} & 0.105 & 0.053 & 0.079\\
\code{11} & 0.089 & 0.239 & 0.164\\
\bottomrule
\end{tabular}
\end{center}
Both columns are normalised distributions, so the mixture is one as well.
Computing the full table both ways confirms the exact invariance:
$\max_s\big|\sP(s\mid\code{10})-\sP(\bar s\mid\code{01})\big|=0$ to machine
precision: the complement twin's \Fmac{} reference \emph{is} the flipped original.
The first-position next-token distribution
$(q_{\code0},q_{\code1},q_{\textsc{end}})$ makes the correction visible:
raw \Fmac{} gives the $(0.576,\,0.424,\,0.000)$ of
Appendix~\ref{app:cont} on $x=\code{10}$ (the tape's zero preference), whereas \sFmac{} gives $(0.420,\,0.580,\,0.000)$ on
$x=\code{10}$ and exactly $(0.580,\,0.420,\,0.000)$ on $x=\code{01}$. The
residual asymmetry now tracks the \emph{input} (short programs tend to echo
its leading bit), not the tape: averaged over any complement-closed input
set the first-bit law is $\tfrac12$ exactly.

\section{Worked example: abductive posterior}\label{app:abd}
Candidates $\mathcal{C}=\{\code0,\code1,\code{10},\code{11}\}$, hidden true
input $\code{10}$, revealed output $\code{01}$, uniform prior $\pi=1/4$.
Equation~\eqref{eq:abd} reweights each candidate by its prefix mass
$\mu(x,r)$:

\begin{center}
\begin{tabular}{@{}lcccc@{}}
\toprule
prefix $r$ & $P^F(\cdot\mid r)$ over $(\code0,\code1,\code{10},\code{11})$ & $P^F(\text{true})$ & entropy (bits)\\
\midrule
$\varepsilon$ & $(0.250,\ 0.250,\ 0.250,\ 0.250)$ & 0.250 & 2.000\\
$\code0$       & $(0.299,\ 0.233,\ 0.233,\ 0.235)$ & 0.233 & 1.991\\
$\code{01}$    & $(0.131,\ 0.269,\ 0.269,\ 0.330)$ & 0.269 & 1.932\\
\bottomrule
\end{tabular}
\end{center}
Each revealed token reweights the candidates by how much of their output mass
is consistent with the prefix; the evidence-ignoring prior-only foil stays at
$\pi$, giving $\mathrm{JS}(P^F\,\|\,\text{prior-only})=0.0181$ bits at
$r=\code{01}$. With these tiny length-only marginals the update is modest; at
$L{=}13$ the posterior concentrates sharply (Fig.~\ref{fig:families}b).

\section{Worked example: Jensen--Shannon divergence}\label{app:js}
For $p=(0.7,0.2,0.1)$ and $q=(0.4,0.4,0.2)$ over $\{\code0,\code1,\textsc{end}\}$,
Eq.~\eqref{eq:js} gives $m=\tfrac12(p+q)=(0.55,0.30,0.15)$ and
\begin{equation}
  \mathrm{KL}(p\,\|\,m)=0.0681,\quad
  \mathrm{KL}(q\,\|\,m)=0.0652,\quad
  \mathrm{JS}=\tfrac12(0.0681)+\tfrac12(0.0652)=0.0667\ \text{bits},
\end{equation}
finite and symmetric without any smoothing.

\section{Worked example: the \Fmac{}-ICL macro-average}\label{app:score}
From the frozen Qwen evaluation, the continuation family has mean divergence
$\bar D_{\mathrm{cont}}=0.3973$ (Eq.~\eqref{eq:famjs}). To keep the arithmetic
legible we take continuation alone, so the macro-average
(Eq.~\eqref{eq:ficl}) reduces to a single family:
\begin{equation}
  \mathrm{FICL}=1-\tfrac{1}{1}\,\bar D_{\mathrm{cont}}=1-0.3973=0.6027,
\end{equation}
with keystroke-reference thresholds $1-\overline{\mathrm{JS}}_{\text{hedge}}=0.8695$ and
$1-\overline{\mathrm{JS}}_{\text{committed}}=0.4792$, and median
sample-complexity efficiency $0.333$ (Eq.~\eqref{eq:sceff}). With both
families scored, $\mathrm{FICL}=\tfrac12\big[(1-\bar D_{\mathrm{ind}})+(1-\bar
D_{\mathrm{cont}})\big]$, each family weighted equally regardless of its task
count. On the anchored scale of Eq.~\eqref{eq:skill} this run falls on the
\emph{lower} branch: its mean divergence $0.3973$ exceeds the hedged keystroke
divergence $\overline{\mathrm{JS}}_{\text{hedge}}=0.1305$, so
$\mathrm{FICL\text{-}A}<\tfrac12$ and the run is scored by its position between
the worst-case predictor and the keystroke reference rather than between the
reference and Bayes. This is the headline configuration of
\S\ref{sec:eval-panel}.

\section{Worked example: the dumped program joint}\label{app:index}
The dump records, per length $L$ and input $x$, an inverted index
$\text{clean output}\mapsto\text{sorted rule list}$. For $L=3$, $x=\code0$
(the $5^3=125$ length-3 programs) the index begins:

\begin{center}
\begin{tabular}{@{}lrl@{}}
\toprule
output $s$ & \#rules & first rule identities\\
\midrule
\code{00}  & 10 & $5,12,19,26,37,44,60,61,\dots$\\
\code{000} & 4  & $1,6,25,30$\\
\code0      & 3  & $79,84,89$\\
\code{01}  & 3  & $7,35,51$\\
\code{10}  & 3  & $11,27,55$\\
\bottomrule
\end{tabular}
\end{center}
This input has $35$ halting programs over $11$ distinct outputs (the CSR
arrays \code{rules\_ptr}/\code{rules} pack the lists; \code{bit\_length}/
\code{out\_word\_ptr}/\code{out\_words} pack the outputs). Serving the
hidden program of Appendix~\ref{app:posterior}
($b^\star=(\code{00},\code{00},\code{01})$, query on the empty input, examples
$\code0\!\to\!\code{00}$ then $\code1\!\to\!\code{01}$) by Eq.~\eqref{eq:serve}
(intersecting the rule lists across examples, re-running the survivors on
the query, and weighting lengths $L=1,2,3$ by $2^{-\ell(L)}$) reproduces the
behaviour-table joint \emph{exactly}:

\begin{center}
\begin{tabular}{@{}clc@{}}
\toprule
$n$ & served $P(\cdot\mid x_q,\mathcal{E}_n)$ (top) & $\max\big|\text{served}-\text{joint}\big|$\\
\midrule
0 & $\bot{:}0.565,\ \code{00}{:}0.186,\ \code1{:}0.070,\ \code{000}{:}0.041,\dots$ & $0$\\
1 & $\code{00}{:}1.000$ & $0$\\
2 & $\code{00}{:}1.000$ & $0$\\
\bottomrule
\end{tabular}
\end{center}
The $0$-shot row is served by the bounded universe scan (no positive example
to intersect); the $1$- and $2$-shot rows by list intersection plus
seed-and-rerun. Agreement is bit-exact ($0$ to machine precision), which is
the property that lets the dumped joint serve the induction \Fmac{} references at the
complete $L\le13$ budget, where the small-program joint is intractable.

\normalsize
\end{document}

%% file: tables/prior_free_appendix.tex
\section{The reference-machine objection: finite-cover bounds and prior-robust diagnostics}
\label{app:priorfree}

A natural objection to scoring against the \Fmac{} reference is that this
posterior is Bayes-optimal only under the declared \sFmac{} prior. A language
model need not hold that prior, and the standard prompt does not disclose it.
Under another universal reference machine,
description lengths change by an additive invariance constant
\cite{li2019kolmogorov}; on finite problems that constant can be of the
same order as the descriptions being compared. The question is therefore
not whether every alternative prior numerically agrees with
$P^{F}$ at the present budget, but which qualitative constraints follow for
the narrower class of alternative predictors that uniformly finitely cover
$P^{F}$. Arbitrary priors can exclude the \Fmac{} process and need satisfy no
such bound.

Two distinctions are essential. First, a finite experiment cannot exclude
an arbitrarily large finite invariance constant: on any fixed finite
horizon, every full-support predictor has some finite multiplicative
covering constant. What can be tested without choosing that constant is
whether the \emph{minimum constant required by the observations remains
bounded as evidence grows}. Second, the continuation and induction
families have different information structures. Continuation is a genuine
prequential reveal: the symbol scored at one step is revealed before the
next. The induction family instead re-probes a held-out query after adding
new demonstrations; its previously scored query answer is not itself
revealed. The pathwise and cumulative-loss theorems below therefore apply
directly to continuation, and to any prequential version of induction, but
not to the sequence of repeated held-out induction probes.

\paragraph{Setting and finite coverage.}
Fix all exogenous side information $x$ for an episode: the input, prompt
format, and, where appropriate, the schedule of queries or reveals. Let
$P^{F}_{x}$ be the exact reference measure on a sequentially revealed
string $\sigma_{1:T}=\sigma_1\cdots\sigma_T$ over a finite alphabet
$\mathcal A$. Let $Q_x$ be a second fixed predictor. We call the serving
rule \emph{coherent} when its normalized conditionals are functions of the
available history and therefore induce a joint measure by the chain
product. Hidden provider state or routing must either be held fixed or
included in $x$ for this interpretation.

We say that $Q$ \emph{$C$-covers} $P^{F}$ when one constant $C<\infty$
satisfies
\begin{equation}
  Q_x(\sigma_{1:T})
  \;\ge\;
  2^{-C}P^{F}_x(\sigma_{1:T})
  \qquad
  \text{for every }x,T,\sigma_{1:T}.
  \label{eq:cover}
\end{equation}
The quantifier over all horizons is important: on a single frozen finite
benchmark, positivity alone would make some finite $C$ exist. Equation
\eqref{eq:cover} is the uniform statement corresponding to a fixed prior
penalty rather than a horizon-dependent one.

Any Bayesian mixture over measures that contains $P^{F}$ as an
expert with weight $w>0$ satisfies Eq.~\eqref{eq:cover} with
$C=-\log_2 w$. Standard universal semimeasures obey the analogous
dominance property for computable environments; because the scored objects
here are normalized predictive distributions, we state the argument for
measures and need only the explicit inequality in Eq.~\eqref{eq:cover}.
Because the released enumerator is computable, changing the reference
machine may change the associated dominance constant, but not its
dependence on the amount of subsequently revealed data. We use coverage
only as a necessary consequence of the specific objection that a model is
implementing a fixed alternative universal mixture that does not exclude
the computable reference process a priori; it is not a definition of every
conceivable form of rational inference.

\paragraph{Bound 1: pathwise regret, with no sampling assumption.}
For every sequentially revealed string, curated or not, Eq.~\eqref{eq:cover}
implies
\begin{equation}
  R_T(Q)
  \;:=\;
  \sum_{t=1}^{T}
  \log_2
  \frac{P^{F}_t(\sigma_t\mid\sigma_{<t},x)}
       {Q_t(\sigma_t\mid\sigma_{<t},x)}
  \;=\;
  \log_2\frac{P^{F}_x(\sigma_{1:T})}
                 {Q_x(\sigma_{1:T})}
  \;\le\;C.
  \label{eq:pathregret}
\end{equation}
Equivalently, the cumulative log-loss of $Q$ can exceed that of the
reference by at most $C$ bits. No expectation is taken, so curation cannot
invalidate the statement.

This gives a directly observable lower bound on every admissible covering
constant:
\begin{equation}
  C^{\rm path}_{\rm req}(T)
  \;:=\;
  \max_{1\le s\le T}[R_s(Q)]_+.
  \label{eq:creqpath}
\end{equation}
Every fixed finite cover requires
$\sup_T C^{\rm path}_{\rm req}(T)\le C$. Thus
$C^{\rm path}_{\rm req}(T)$ itself is a prior-free diagnostic: it can be
computed without specifying the rival prior or $C$. The theorem predicts
boundedness, not a particular numerical ceiling. A sustained increase of
the required bits as the horizon is extended is therefore evidence against
a fixed-cover explanation; a finite observed horizon supplies a lower
bound on $C$, not a proof that no larger finite $C$ exists.

\paragraph{Bound 2: a finite expected KL budget.}
When the reveal stream is distributed according to $P^{F}$, the chain
rule for relative entropy and Eq.~\eqref{eq:cover} give the standard
universal-prediction bound \cite{solomonoff1978,hutter2005uai}
\begin{equation}
  \sum_{t=1}^{T}
  \mathbb E_{P^{F}}
  \mathrm{KL}\!\left(P^{F}_t\,\middle\|\,Q_t\right)
  \;=\;
  \mathrm{KL}\!\left(P^{F}_{1:T}\,\middle\|\,Q_{1:T}\right)
  \;\le\;C,
  \label{eq:klsum}
\end{equation}
where KL is measured in bits.

To transfer this statement to the reported Jensen--Shannon geometry we can
avoid the square-root loss introduced by Pinsker's inequality.

\emph{Lemma (one-sided KL domination of standard JS).}
For probability distributions $p,q$ on a common alphabet, with the
standard Jensen--Shannon convention
\[
  \mathrm{JS}(p\|q)
  =\tfrac12\mathrm{KL}(p\|m)+\tfrac12\mathrm{KL}(q\|m),
  \qquad m=\tfrac12(p+q),
\]
and the same logarithm base in both divergences,
\begin{equation}
  \mathrm{JS}(p\|q)
  \;\le\;
  \kappa_{\rm JS}\,\mathrm{KL}(p\|q),
  \qquad
  \kappa_{\rm JS}:=\frac{\ln 2}{2}.
  \label{eq:jskl}
\end{equation}
The constant is sharp. One proof is by functional domination of
$f$-divergences \cite{sasonverdu2016}: with natural logarithms, use the
canonical generators
\[
 f_{\rm KL}(u)=u\ln u-u+1,
 \qquad
 f_{\rm JS}(u)=\tfrac12\!\left[
 u\ln\frac{2u}{1+u}+\ln\frac{2}{1+u}\right],
\]
for which
$\sup_{u\ge0} f_{\rm JS}(u)/f_{\rm KL}(u)=\ln2/2$; the ratio is unchanged
by a common change of logarithm base.

Combining Eqs.~\eqref{eq:klsum} and \eqref{eq:jskl}, define
\begin{equation}
  j_t
  :=
  \mathbb E_{P^{F}}
  \mathrm{JS}\!\left(P^{F}_t\,\middle\|\,Q_t\right),
  \qquad
  J_T:=\sum_{t=1}^{T}j_t.
\end{equation}
Then
\begin{equation}
  J_T\le \kappa_{\rm JS}C
  \qquad\text{for all }T,
  \label{eq:jsbudget}
\end{equation}
and hence
\begin{equation}
  \sum_{t=1}^{\infty}j_t<\infty,
  \qquad
  j_t\to0.
  \label{eq:jssummable}
\end{equation}
Thus a finite-cover predictor has a finite lifetime budget of expected JS
disagreement with the reference along a $P^{F}$-distributed reveal
stream.

Writing the Ces\`aro mean as
\begin{equation}
  \bar J_T:=\frac{J_T}{T},
\end{equation}
Eq.~\eqref{eq:jssummable} implies the stronger asymptotic form
\begin{equation}
  \bar J_T
  =\frac{J_\infty}{T}+o(T^{-1}),
  \qquad
  J_\infty:=\sum_{t=1}^{\infty}j_t<\infty.
  \label{eq:cesarorate}
\end{equation}
The $1/T$ term is therefore not merely a loose envelope for the Ces\`aro
mean: it is the leading form generated by a finite cumulative JS budget
(unless $J_\infty=0$, in which case the predictors agree identically in
this metric).

Coverage also implies $P^{F}\ll Q$ on the corresponding infinite
reveal process; the Blackwell--Dubins merging theorem therefore gives the
same qualitative conclusion without reference to the numerical value of
$C$ \cite{blackwell1962}.

\paragraph{A $C$-free Richardson-type tail diagnostic.}
Equation~\eqref{eq:cesarorate} suggests cancelling the unknown
$J_\infty/T$ transient rather than estimating its amplitude. For any
$T_2>T_1$, define
\begin{equation}
  D^{\rm tail}_{\rm JS}(T_1,T_2)
  :=
  \frac{T_2\bar J_{T_2}-T_1\bar J_{T_1}}
       {T_2-T_1}
  =
  \frac{J_{T_2}-J_{T_1}}{T_2-T_1}.
  \label{eq:jstail}
\end{equation}
This is not only analogous to Richardson cancellation: it is exactly the
mean expected JS contributed by the newly added evidence block. Every
finite-cover predictor satisfies
\begin{equation}
  D^{\rm tail}_{\rm JS}(T_1,T_2)\to0
  \label{eq:jstailzero}
\end{equation}
as the block moves into the tail. If the observed mean divergence instead
approaches a positive floor $\bar J_T\to\delta>0$, then for any fixed scale
ratio $T_2/T_1>1$,
\begin{equation}
  D^{\rm tail}_{\rm JS}(T_1,T_2)\to\delta.
\end{equation}
For doubling, Eq.~\eqref{eq:jstail} reduces to
\begin{equation}
  D^{\rm tail}_{\rm JS}(T,2T)=2\bar J_{2T}-\bar J_T.
  \label{eq:jsrichardson}
\end{equation}
The reference value zero contains no $C$ and no rival-prior parameter.

A useful dimensionless companion is
\begin{equation}
  I^{\rm tail}_{\rm JS}(T)
  :=
  \frac{J_{2T}-J_T}{J_T}
  =
  \frac{2\bar J_{2T}-\bar J_T}{\bar J_T},
  \label{eq:jsindex}
\end{equation}
when $J_T>0$. A finite nonzero disagreement budget gives
$I^{\rm tail}_{\rm JS}(T)\to0$, whereas a genuine constant divergence
floor gives $I^{\rm tail}_{\rm JS}(T)\to1$. This ratio is a diagnostic of
the plateau alternative, not a sufficient test for coverage: an
unbounded but very slowly growing cumulative budget can also have a
vanishing dyadic ratio. The theorem-level requirement is boundedness of
$J_T$ itself.

For completeness, Eq.~\eqref{eq:jsbudget} also yields the finite-horizon
lower bound
\begin{equation}
  C^{\rm JS}_{\rm req}(T)
  :=
  \frac{J_T}{\kappa_{\rm JS}}
  =
  \frac{2T\bar J_T}{\ln2}
  \;\le\; C.
  \label{eq:creqjs}
\end{equation}
Unlike Eqs.~\eqref{eq:jstail}--\eqref{eq:jsindex}, this statistic is in
bits and explicitly estimates the minimum constant required by the
observed expected divergence.

\paragraph{Bound 3: reference-free truth budget for a deterministic reveal.}
Continuation admits an even stronger statement that does not mention
$P^{F}$ at all. Let $Q$ be any Bayesian mixture over deterministic
hypotheses, and suppose the true deterministic hypothesis $h^{\dagger}$
has prior weight $w_{\dagger}>0$. Along the realised reveal sequence,
\begin{equation}
  Q(\sigma_{1:T}\mid x)
  \ge w_{\dagger},
\end{equation}
so the model's cumulative truth log-loss obeys
\begin{equation}
  L_T^{\rm truth}
  :=
  \sum_{t=1}^{T}-\log_2 Q_t(\sigma_t\mid\sigma_{<t},x)
  =-\log_2Q(\sigma_{1:T}\mid x)
  \le -\log_2w_{\dagger}.
  \label{eq:truthbudget}
\end{equation}
Thus $L_T^{\rm truth}$ must remain bounded under \emph{every} fixed prior
that assigns the truth positive mass. This is a necessary condition for
Bayesian induction in the realisable, noise-free sequential setting, and its
loss function is reference-free: $P^{F}$ appears nowhere in
Eq.~\eqref{eq:truthbudget}, which scores only the predictor's own probability
on the symbol that actually arrived.

One distinction should be drawn sharply here, because the two halves of
``reference-free'' do not travel together. The \emph{condition} is free of the
reference machine; the \emph{evidence} is not. The strings scored are outputs
of enumerated \sFmac{} programs, selected by a curation defined on that same
enumeration, so the diagnostic constrains a rival prior on this particular
family of computable stories rather than on computable stories in general. A
different reference machine would supply different stories, and nothing here
excludes a predictor that spends less of its budget on those. What the bound
removes is the need for the rival to share our prior, our machine, or our
posterior in order to be measured; what it does not remove is the origin of
the data.

Two immediate consequences are
\begin{equation}
  \sum_{t=1}^{\infty}
  \bigl[1-Q_t(\sigma_t\mid\sigma_{<t},x)\bigr]
  <\infty,
  \label{eq:truthprob}
\end{equation}
and a finite number of MAP mistakes. Indeed,
$1-q\le -\ln q=(\ln2)(-\log_2q)$, and every wrong MAP prediction has
$Q_t(\sigma_t\mid\cdot)\le1/2$ and therefore costs at least one bit.
Hence
\begin{equation}
  N_{\rm MAP}(T)
  \le L_T^{\rm truth}
  \le -\log_2w_{\dagger}.
  \label{eq:mistakebudget}
\end{equation}
The quantities
\begin{equation}
  L_T^{\rm truth},
  \qquad
  \frac{L_{T_2}^{\rm truth}-L_{T_1}^{\rm truth}}{T_2-T_1},
  \qquad
  \frac{N_{\rm MAP}(T)}{T}
  \label{eq:truthdiagnostics}
\end{equation}
are therefore prior-free diagnostics in the same sense as
Eq.~\eqref{eq:creqpath}: no alternative prior or $w_{\dagger}$ is needed
to compute them. The theorem predicts bounded cumulative truth loss and a
vanishing tail loss and mistake density.

\paragraph{A pathwise JS corollary requiring no $P^F$ sampling.}
The deterministic-truth bound also gives a second route to JS merging that
is distinct from the $P^F$-expectation argument in
Eqs.~\eqref{eq:jsbudget}--\eqref{eq:jsindex}. Let $P$ and $Q$ be any two
Bayesian mixtures over deterministic hypotheses, and suppose both assign
positive mass $w_P,w_Q>0$ to the realised computable truth
$\sigma_{1:\infty}$. Writing $\delta_{\sigma_t}$ for the point mass on the
realised symbol and measuring JS in bits \cite{sasonverdu2016},
\begin{align}
 \mathrm{JS}(P_t\|Q_t)
 &\le \operatorname{TV}(P_t,Q_t) \notag\\
 &\le \operatorname{TV}(P_t,\delta_{\sigma_t})
       +\operatorname{TV}(Q_t,\delta_{\sigma_t}) \notag\\
 &= [1-P_t(\sigma_t)]+[1-Q_t(\sigma_t)].
 \label{eq:pathjsstep}
\end{align}
Using $1-u\le-\ln u$ and the two truth-loss bounds therefore gives
\begin{equation}
 \sum_{t=1}^{T}\mathrm{JS}(P_t\|Q_t)
 \le
 (\ln2)\left[-\log_2 w_P-\log_2 w_Q\right]
 \qquad\text{for every }T.
 \label{eq:pathjsbudget}
\end{equation}
Consequently, on this individual reveal path, the new-block JS contribution
vanishes and its dyadic ratio tends to zero whenever the accumulated
disagreement is nonzero:
\begin{equation}
 D^{\rm tail,path}_{\rm JS}(T,2T)\to0,
 \qquad
 I^{\rm tail,path}_{\rm JS}(T)\to0.
 \label{eq:pathjslimit}
\end{equation}
This result neither draws the story from $P^F$ nor averages histories under
$P^F$. It applies in particular when $P=P^F$ contains the enumerated program
that produces the benchmark story and $Q$ is another universal Bayesian
mixture, since such a mixture assigns positive mass to every computable
deterministic truth. It also extends to any fixed finite average of such
paths, although the bound may differ by story. Thus the observed dyadic JS
index has a pathwise prior-robust interpretation in addition to its curated
descriptive interpretation. Equation~\eqref{eq:historychain} remains the
appropriate theorem for a population expectation over random $P^F$
histories; the two claims should not be conflated.

The positive-mass condition is essential and does not supply a uniform rate.
Across the unrestricted class $w_Q$ may be arbitrarily small, making the
right-hand side of Eq.~\eqref{eq:pathjsbudget} arbitrarily large. Indeed, any
finite sequence of full-support predictive conditionals can be embedded in a
proper Bayesian construction that reproduces those conditionals through the
observed horizon, retains a positive but arbitrarily small component on the
true continuation, and converges only later. No finite-horizon hypothesis
test can therefore reject the union of \emph{all} proper priors assigning
positive truth mass. A statistical rejection requires either a prescribed
upper bound on the prior penalty, a prescribed convergence tolerance at the
observed horizon, or a restricted prior family.

\paragraph{Approximate tracking rather than exact Bayesianity.}
If a served model $M$ is claimed only to approximate a covering inductor
$Q$, the metric property of $\sqrt{\mathrm{JS}}$ \cite{endres2003} gives
a sharper transfer inequality than a direct squaring bound. Over any
common averaging law, Minkowski's inequality yields
\begin{equation}
  \sqrt{\overline{\mathrm{JS}}(P^{F},M)}
  \le
  \sqrt{\overline{\mathrm{JS}}(P^{F},Q)}
  +
  \sqrt{\overline{\mathrm{JS}}(Q,M)}.
  \label{eq:transfer}
\end{equation}
On a $P^{F}$-distributed sequential reveal,
Eq.~\eqref{eq:jsbudget} can be substituted into the first term. The exact
alternative-prior hypothesis is the special case $M=Q$; without an
independent bound on the approximation term, an unrestricted claim that
$M$ merely ``tracks'' some rational predictor is not falsifiable.

\paragraph{Why the repeated held-out induction probes are different.}
The preceding cumulative budgets require a genuine prequential sequence:
the outcome scored at one step must be part of the evidence available at
later steps. That condition holds for continuation. It does \emph{not}
hold for the current induction protocol, where the model repeatedly
predicts the same held-out query after new demonstration pairs are added,
while its previous query target is never revealed.

In particular, a Bayesian held-out answer need not be monotone even
in a realisable, noiseless deterministic hypothesis class. Consider three
hypotheses with prior weights
\[
  \pi(h^{\dagger})=0.20,\qquad
  \pi(h_1)=0.50,\qquad
  \pi(h_2)=0.30,
\]
where $h^{\dagger}$ is the truth. On the held-out query, let
$h^{\dagger}$ and $h_1$ predict $A$ and $h_2$ predict $B$. Before seeing
evidence, $A$ is the correct MAP answer with mass $0.70$. Suppose the
next demonstration eliminates $h_1$ but not $h_2$. The posterior then
assigns normalized masses $0.40$ to the true $A$ hypothesis and $0.60$ to
the surviving $B$ hypothesis, so the Bayes MAP becomes wrong. A later
demonstration that eliminates $h_2$ restores the correct answer. Thus an
exact Bayesian learner can move
\[
  \text{correct}\;\longrightarrow\;\text{wrong}\;\longrightarrow\;\text{correct}
\]
without noise and without any decrease in posterior mass on the truth.
Posterior odds among surviving hypotheses remain fixed; what changes is
which competitors survive.

Consequently, the number of solved$\to$unsolved transitions in the
existing held-out induction curves is an empirical instability statistic,
not by itself a prior-free impossibility theorem. Once every surviving
hypothesis in a \emph{specified} hypothesis class agrees on the query, the
answer is frozen under every prior supported on that class; but changing
the reference machine can enlarge the hypothesis class, so that
finite-support-collapse statement is not reference-machine independent.

\paragraph{A prequential induction audit restores the prior-free theorem.}
The induction tasks already contain an ordered sequence of demonstrations
$(x_1,o_1),\ldots,(x_N,o_N)$. A theorem-grounded companion audit can score
them prequentially: at step $t$, show only
$(x_1,o_1),\ldots,(x_{t-1},o_{t-1})$, ask for the distribution of the
output at $x_t$, score the realised $o_t$, and then reveal $(x_t,o_t)$
before step $t+1$. The exact \Fmac{} posterior for this query is available
from the same program-set intersection backend.

This produces a genuine sequential reveal and therefore inherits
Eqs.~\eqref{eq:pathregret}--\eqref{eq:truthdiagnostics}. In particular,
for every alternative finite cover the required pathwise bits
$C^{\rm path}_{\rm req}(T)$ must remain bounded, while for every Bayesian
mixture that gives the true deterministic program positive weight the
cumulative truth log-loss $L_T^{\rm truth}$ must remain bounded. These two
statistics provide the reference-machine robustness that repeated
held-out-query forgetting cannot.

\paragraph{Finite-horizon impossibility and the correct interpretation of
``prior-free''.}
No statistic computed on a fixed finite, full-support panel can logically
exclude \emph{every} finite $C$. If the tested domain contains only
finitely many strings and $Q(\sigma)>0$ wherever
$P^{F}(\sigma)>0$, then
\begin{equation}
  C_{\max}
  =
  \max_{\sigma\ \text{in the tested domain}}
  \log_2\frac{P^{F}(\sigma)}{Q(\sigma)}
  <\infty.
  \label{eq:finiteimpossibility}
\end{equation}
An exact support violation, $Q(\sigma)=0<P^{F}(\sigma)$, immediately
rules out finite coverage, but ordinary softmax predictors generally have
full support before API truncation.

The role of the diagnostics above is therefore precise. They require no
chosen rival prior and no guessed invariance constant. At finite horizon
they report the minimum bits already required; over extensible horizons a
fixed alternative prior predicts bounded cumulative budgets and vanishing
tail contributions. A persistent positive divergence or log-loss floor
forces the required budget to grow without bound and is asymptotically
incompatible with every finite constant. This is the sense in which the
test is prior-free.

\paragraph{What the present panel licenses.}
The strongest theorem-grounded statements come from the sequentially
revealed continuation family. Equation~\eqref{eq:pathregret} applies to
its realised paths even under the length--complexity curation, so
$C^{\rm path}_{\rm req}(T)$ can be computed on the frozen panel without
assuming that the sampled outputs are typical under $P^{F}$.
Equation~\eqref{eq:truthbudget} gives a second, reference-free audit on the
same paths, provided the complete served probability of every realised
symbol is retained.

The empirical results and figures are reported in the main Results section
(\S\ref{sec:sequential-results}); here we retain the proof, estimand and
sampling conditions needed to interpret them.

The frozen caches support two distinct estimands. The all-panel estimand is the
\emph{operational three-class predictor} $\widetilde Q$ used by the benchmark.
At every history it folds recorded top-$k$ token mass onto
$\{0,1,\textsc{end}\}$, adds $10^{-9}$ to each class, and normalises. Once this
transformation is declared as the predictor rather than an estimator of a
provider's native distribution, it is a proper conditional rule, is positive
on every scored class, and induces a coherent joint measure by the chain
product. Epsilon smoothing therefore does not invalidate
Eqs.~\eqref{eq:pathregret} and \eqref{eq:truthbudget} for $\widetilde Q$; it
changes the estimand and can mechanically limit any loss attached to a missing
class. We make neither the transformation nor the limitation implicit.

The all-panel reconstruction recovers 23,998 of 24,000
configuration--trajectory pairs (155,984 realised-symbol cells) from all 80
usable configurations. Only 991 cells ($0.635\%$) are at the $10^{-9}$ floor
and 3,313 ($2.12\%$) are at or below $10^{-6}$; 68 configurations have no
floor event and 74 have at most two. The reported sensitivity removes a whole
trajectory if any realised symbol touches the floor. It retains 23,069
endpoints, so it tests whether a small number of clipped events drives the
all-panel relation while preserving the sequential unit.

The second estimand is available for six frontier runs whose endpoint caches
retain folded masses of $0$, $1$, and END \emph{before} smoothing. We define an
explicit \emph{top-$k$-projected scoring rule} by normalising those three
observed masses, without clipping or epsilon replacement. This is also a
proper conditional distribution and coherent point estimand for
Eqs.~\eqref{eq:pathregret} and \eqref{eq:truthbudget}; it is not asserted to
equal the provider's latent full-vocabulary distribution. If the realised
class has zero recorded mass at any position, the complete trajectory is
censored from this point diagnostic. The retained counts are 298, 299, 222,
49, 75, and 126 of 300 trajectories for Gemini~2.5~Pro and GPT-5.1 through
GPT-5.6-sol, respectively.

The reconstruction also preserves the distinction between projection and
native distribution. Allocating all unreported mass either toward or away
from the realised class gives rigorous native-probability and cumulative-loss
intervals in the released tables. The largest unreported mass is below 0.020
in every retained cache (below 0.0027 for Gemini~2.5~Pro and 0.00014 for
GPT-5.1), but the paper figures report only the exactly identified projected
estimand. A clean complete-support rerun manifest is retained for estimating
the native serving rule directly.

Figures~\ref{fig:pf6-operational} and \ref{fig:pf7-smoothing} report the
all-panel operational estimand and its floor-free sensitivity;
Figs.~\ref{fig:pf1-coverage} and \ref{fig:pf2-truth} report the six-cache
pre-smoothing projection. Because continuation ends at different positions,
all cumulative quantities are stopped after END. This avoids conditioning
later positions on the subset of longer strings. The exact finite-panel
fraction above any fixed ceiling requires no $p$-value: a realised path with
$C^{\rm path}_{\rm req}>C_0$ is a direct counterexample to that particular
ceiling on the tested domain. To avoid selecting $C_0$ after inspecting the
data, we report the complete threshold-survival curve. Simultaneous 95\%
bands over its displayed grid come from crossed resampling of 45 checkpoint
families (quantisations kept together) and 150 complement-twin task clusters.
They measure stability to panel composition, not uncertainty under
$P^F$ sampling. The separate late-versus-early direction test uses the
checkpoint family as its unit: 38 of 45 cluster means improve, giving the
one-sided exact sign-test value $p=1.56\times10^{-6}$ and a cluster-bootstrap
95\% interval $[-0.087,-0.043]$ bits for the mean change. This sign test is
conditional on exchangeable direction signs across the 45 checkpoint families;
it does not convert the deliberately assembled model panel into a random sample
of all LLMs. None of these finite curves or tests rejects every possible larger
finite $C$.

By contrast, the expected JS budget of
Eq.~\eqref{eq:jsbudget} refers to a $P^{F}$-distributed reveal stream;
the curated continuation sample deliberately over-represents atypical
length and complexity strata, so its raw mean JS should not be inserted
into Eq.~\eqref{eq:creqjs} as though it were that expectation. A dedicated
prior-sampled audit, or an exact importance reweighting whose inclusion
probabilities are known, is required for the quantitative JS-budget test.
This restriction is specific to the expectation in PF-3. It has no bearing on
the casewise identity in Eq.~\eqref{eq:pathregret}: averaging casewise PF-1
values under the curated law may be descriptive, but each individual
inequality remains exact regardless of how that case entered the panel.

\paragraph{Why the expectation law cannot be replaced by a size weight.}
The $P^F$ expectation in Eq.~\eqref{eq:jsbudget} is forced by the chain rule, not chosen for
convenience. Writing $H_t=\sigma_{<t}$,
\begin{equation}
 \mathrm{KL}(P^F_{1:T}\|Q_{1:T})
 =
 \sum_{t=1}^{T}
 \mathbb E_{H_t\sim P^F}
 \mathrm{KL}\!\left(P^F_t(\cdot\mid H_t)\|Q_t(\cdot\mid H_t)\right).
 \label{eq:historychain}
\end{equation}
Coverage bounds the joint KL on the left; averaging the conditional terms
under another history law does not equal that quantity. In particular,
deliberately over-sampling histories that are rare under $P^F$ can leave a
large curated mean even when their $P^F$-weighted contribution is summable.

If histories were sampled from a known law $q_{{\rm eval},t}$ with adequate
support, the relevant importance identity would instead be
\begin{equation}
 j_t
 =
 \mathbb E_{(x,h)\sim q_{{\rm eval},t}}
 \left[
 \frac{\mu(x)P^F(h\mid x)}{q_{{\rm eval},t}(x,h)}
 \mathrm{JS}\!\left(P^F_t(\cdot\mid h,x)\|Q_t(\cdot\mid h,x)\right)
 \right],
 \label{eq:importancehistory}
\end{equation}
where $\mu$ is the chosen target law over exogenous inputs. The numerator is
a \emph{prefix mass},
$P^F(h\mid x)=\sum_{p:p(x)\succeq h}w(p)$, aggregating every compatible
program. It is neither $2^{-|h|}$ nor the weight of one generating program.

The benchmark selection denominator is also not a length code. Construction
assigns equal quotas to final output lengths 3--8 and to three
conditional-complexity terciles, draws proportionally to $P^F(s\mid x)$
within each stratum without replacement, rejects duplicate and mirrored
pairs, and then inserts the complement twin. Multiplying by $P^F(s\mid x)$
again would therefore approximately square the within-stratum mass rather
than undo selection. Exact Horvitz--Thompson correction would require the
joint prefix inclusion probabilities generated by the sequential
without-replacement scheme. More importantly, the released design assigns
zero inclusion probability to positive-$P^F$ histories outside the selected
length support and contains no model call on their unobserved branches. No
finite weight can reconstruct those missing terms. A reconstructed
within-support weighting remains a useful curation sensitivity, but it is not
the expectation in Eq.~\eqref{eq:historychain}.

\paragraph{The longitudinal finite-scale tests.}
For both operational JS and realised-symbol truth loss, the released analysis
forms stopped dyadic blocks at $T=1,2,4$: an episode contributes zero after
END. Quantisations are averaged within 45 checkpoint families and complement
twins within 150 base-task clusters. Point estimates weight these two units
equally; simultaneous 95\% bands use a crossed checkpoint-by-twin bootstrap
and a max-deviation critical value over all three displayed scales. At $T=4$,
checkpoint-family ratios are also compared with the theoretically determined
midpoint $1/2$ between the finite-budget limit zero and positive-floor limit
one using an exact one-sided sign test. This is a finite-scale directional
test, not a test that assumes the ratio must already have reached its limit.

The JS tail contribution remains $0.271$ bits per step and its aggregate index
$0.666$; 37 of 45 checkpoint-family indices lie closer to one
($p=7.69\times10^{-6}$). At the same time, 44 of 45 families have a smaller
$T=4$ than $T=2$ contribution ($p=1.31\times10^{-12}$), so the data support
improvement but not entry into a near-zero tail regime. The reference-free
truth tail remains $2.315$ bits per step with aggregate ratio $0.752$; 39 of
45 checkpoint-family ratios lie closer to one ($p=2.71\times10^{-7}$).
Whole-trajectory floor removal leaves both conclusions unchanged.

No draw from $P^F$ is needed merely to \emph{define} these dyadic statistics.
Given any observed nonnegative sequence $z_1,z_2,\ldots$, its new-block mean
and ratio can be computed as
\[
  \frac{1}{T}\sum_{t=T+1}^{2T}z_t,
  \qquad
  \frac{\sum_{t=T+1}^{2T}z_t}{\sum_{t=1}^{T}z_t}.
\]
The sampling law enters when those descriptive quantities are connected to a
theorem. For the JS panels, the empirical position mean is
$\widetilde j_t=\mathbb E_{q_{{\rm eval},t}}
\mathrm{JS}(P^F_t\|Q_t)$, not the
$j_t=\mathbb E_{P^F}\mathrm{JS}(P^F_t\|Q_t)$ in
Eq.~\eqref{eq:jsbudget}. Consequently, the curated dyadic JS result is a
finite-scale statement about the observed panel; without $P^F$ sampling or
valid importance weights it is not a test of the $P^F$-expected asymptotic
limit in Eq.~\eqref{eq:jstailzero}. The exact sign tests likewise quantify
consistency across the 45 checkpoint families under the curated law, not tail
probability under $P^F$.

This restriction on estimating the $P^F$ expectation does not remove the
separate pathwise interpretation proved in
Eqs.~\eqref{eq:pathjsstep}--\eqref{eq:pathjslimit}. Because each benchmark
story is the deterministic output of an enumerated program, two Bayesian
mixtures that both assign its truth positive mass must have a bounded
pathwise JS sum and hence dyadic index tending to zero. The truth-loss panels
have the still more direct status of using no \Fmac{} posterior inside the
loss. Both diagnostics therefore test a necessary finite-scale signature of
an alternative universal prior on the observed computable stories without
requiring $P^F$ history sampling. Even so, finite observations cannot reject
every such mixture: the bounds involving $w_Q$ can exceed the budget consumed
by position eight, and all released episodes terminate by position nine.
Together the diagnostics show that the discrepancy is not merely exact
numerical mismatch with the chosen \Fmac{} prior at the observed scale; they
do not prove reference-machine invariance or exclude arbitrarily delayed
universal merging.

\paragraph{Finite-horizon hypotheses supported by the frozen panel.}
The strongest nonparametric hypothesis test available here is scale-local.
For a prespecified practical tolerance $\varepsilon$, define
\[
 H_{0,\varepsilon}^{\rm JS}:
 D^{\rm tail,path}_{\rm JS}(4,8)\le\varepsilon,
 \qquad
 H_{1,\varepsilon}^{\rm JS}:
 D^{\rm tail,path}_{\rm JS}(4,8)>\varepsilon,
\]
where the population is the crossed checkpoint-family/twin evaluation law.
Inverting the simultaneous 95\% interval $[0.208,0.333]$ rejects every
$H_{0,\varepsilon}^{\rm JS}$ with $\varepsilon<0.208$ bits per step while
controlling multiplicity over the displayed scales. The analogous truth-loss
interval $[1.527,3.104]$ rejects every tolerance below $1.527$ bits per step.
Reporting the complete confidence intervals avoids choosing a favourable
$\varepsilon$ after inspection. Independently, the exact sign test of
$H_0:\Pr(I^{\rm tail}(4)>1/2)\le1/2$ rejects for JS
($37/45$, $p=7.69\times10^{-6}$) and truth loss
($39/45$, $p=2.71\times10^{-7}$). These tests support the statement that the
panel has not entered a practically near-zero tail regime and that its
current geometry is closer to a persistent increment than to a spent finite
budget. Their confidence refers to composition of the curated panel. They do
not test the unrestricted ``some proper prior'' null, which is not finitely
falsifiable without a bound on its prior penalty or convergence rate.

The late-slope equivalence and positive-intercept fits are deliberately
secondary. The former uses alive-conditioned position means and displays the
TOST result for every margin rather than choosing one after inspection. The
latter fits $\bar J_T=\delta+a/T$ only on $T=4,\ldots,8$; its positive
intercepts are a parametric extrapolation sensitivity because
Eq.~\eqref{eq:cesarorate} permits an
arbitrary finite transient and does not assert that this form is exact at the
observed positions. Finally, every released continuation ends by position
nine. With END made absorbing, all later disagreement and loss increments are
zero by construction. The frozen experiment can therefore identify the
finite-scale signature but cannot observe a genuinely positive
infinite-horizon asymptote.

For induction, the existing held-out divergence curves remain valid
measurements of fidelity to the designated \Fmac{} posterior, and the
observed solved-set instability remains a useful behavioural description.
They should not, however, be presented as distribution-free proofs that no
Bayesian prior can generate the observed revisions. The proposed
prequential induction audit supplies the missing sequential test while
leaving the current posterior-fidelity measurement intact.

Figure~\ref{fig:pf4-revision} in the main Results section is the descriptive
control that the frozen data do license. The exact reference posterior itself moves
correct$\to$wrong 46 times and wrong$\to$correct 406 times across 480
tasks; all reference tasks finish correct, and all 46 adverse revisions
begin at the first or second example. Thus revision per se is not a
Bayesian impossibility. The model panel has substantially more and later
churn, which remains a behavioural finding rather than a telescoping-loss
theorem.

The companion reconstruction and plotting commands now produce the all-panel
operational PF-1/PF-2 curves, their floor-free sensitivity, the six-run
pre-smoothing PF-1/PF-2 validation, the stopped dyadic and extrapolation
sensitivities, and the PF-4 induction control, all with explicit provenance and
scope fields. The machine-readable scope report still
marks PF-3 unsupported by the curated sampling law and PF-5 pending exact
prequential targets and model calls. These omissions are different. PF-3 is a
population expectation and is not needed for the casewise result. PF-5 needs a
different query at each step: the current cache repeatedly predicts the fixed
held-out induction query. Auditing 3,422 distinct prequential prompt keys finds
only two matches in representative complete caches, confirming that the
missing probabilities cannot be recovered by rearranging existing responses.

The print-only reference remains informative for a different reason: it
locates the panel relative to a concrete loop-free algorithmic mixture.
Its bracketing of the panel is structural evidence about the kind of
measure the models resemble, not by itself a proof that every finite
translation constant is impossible.

\paragraph{Scope.}
The argument now separates four claims that should not be conflated.
(i) Uniform finite coverage gives a pathwise constant-regret bound on any
genuine reveal stream. (ii) Under $P^{F}$ sampling, the expected KL
and JS disagreements have finite cumulative budgets, yielding the
$C$-free tail diagnostic of Eq.~\eqref{eq:jstail}. (iii) In a realisable
deterministic reveal, any Bayesian prior with positive truth weight has a
finite cumulative truth-loss and mistake budget, in a loss that does not
mention the reference machine, though the stories on which it is measured are
still generated by one. (iv) Repeated held-out induction queries are not
reveals, so neither cumulative-loss telescoping nor monotone correctness
holds for them without additional assumptions.

Within that scope the invariance-constant objection becomes testable
without choosing the constant: a fixed alternative prior can move the
finite ceiling, but it cannot turn a bounded sequential disagreement
budget into one that continues to accumulate with evidence. The finite
panel can establish lower bounds and scaling signatures; an extensible
prequential audit can test the stronger asymptotic claim directly.

%% file: tables/panel_table.tex
\begin{center}\footnotesize\setlength{\tabcolsep}{3.5pt}
\begin{longtable}{@{}p{3.3cm}llcrl p{2.5cm}@{}}
\caption{Panel registry: every run, its serving route and precision, the families with usable readouts (\code{/lp} = log-probabilities exposed), the billed API cost in USD and the reasoning-effort setting used (local rows carry no metered cost), and audit notes.}\\
\toprule
Model & Precision & Route & Readouts & USD & Effort & Notes\\
\midrule\endfirsthead
\toprule Model & Precision & Route & Readouts & USD & Effort & Notes\\\midrule\endhead
\bottomrule\endfoot
\multicolumn{7}{@{}l}{\emph{Local panel}}\\
MiniMax-M2.7 & Q8\_0 & local & ind+cont/lp & -- & -- & --\\
Ministral-3-14B-Base-2512 & Q8\_0 & local & ind+cont/lp & -- & -- & --\\
Ministral-3-14B-Instruct-2512 & Q8\_0 & local & ind+cont/lp & -- & -- & --\\
Ministral-3-14B-Reasoning-2512 & Q8\_0 & local & ind+cont/lp & -- & -- & --\\
Qwen3-30B-A3B-Base & Q8\_0 & local & ind+cont/lp & -- & -- & --\\
Qwen3-30B-A3B-Instruct-2507 & Q8\_0 & local & ind+cont/lp & -- & -- & --\\
Qwen3-30B-A3B-Thinking-2507 & Q8\_0 & local & ind+cont/lp & -- & -- & --\\
Qwen3.5-0.8B & Q8\_0 & local & ind+cont/lp & -- & -- & --\\
Qwen3.5-27B & Q8\_0 & local & ind+cont/lp & -- & -- & --\\
Qwen3.5-2B & Q8\_0 & local & ind+cont/lp & -- & -- & --\\
Qwen3.5-4B & Q8\_0 & local & ind+cont/lp & -- & -- & --\\
Qwen3.5-9B & Q8\_0 & local & ind+cont/lp & -- & -- & --\\
Qwen3.6-27B & Q3\_K\_M & local & ind+cont/lp & -- & -- & --\\
Qwen3.6-27B & Q4\_K\_M & local & ind+cont/lp & -- & -- & --\\
Qwen3.6-27B & Q5\_K\_M & local & ind+cont/lp & -- & -- & --\\
Qwen3.6-27B & Q6\_K & local & ind+cont/lp & -- & -- & --\\
Qwen3.6-27B & Q8\_0 & local & ind+cont/lp & -- & -- & --\\
Qwen3.6-27B & UD-IQ2\_M & local & ind+cont/lp & -- & -- & --\\
Qwen3.6-35B-A3B & BF16 & local & ind+cont/lp & -- & -- & --\\
Qwen3.6-35B-A3B & Q8\_0 & local & ind+cont/lp & -- & -- & --\\
Qwen3.6-35B-A3B & UD-IQ1\_M & local & ind+cont/lp & -- & -- & --\\
Qwen3.6-35B-A3B & UD-IQ2\_M & local & ind+cont/lp & -- & -- & --\\
Qwen3.6-35B-A3B & UD-Q3\_K\_M & local & ind+cont/lp & -- & -- & --\\
Qwen3.6-35B-A3B & UD-Q4\_K\_M & local & ind+cont/lp & -- & -- & --\\
Qwen3.6-35B-A3B & UD-Q5\_K\_M & local & ind+cont/lp & -- & -- & --\\
Qwen3.6-35B-A3B & UD-Q6\_K & local & ind+cont/lp & -- & -- & --\\
gemma-4-12b-it & Q8\_0 & local & ind+cont/lp & -- & -- & --\\
gemma-4-26B-A4B-it & BF16 & local & ind+cont/lp & -- & -- & --\\
gemma-4-26B-A4B-it & Q8\_0 & local & ind+cont/lp & -- & -- & --\\
gemma-4-26B-A4B-it & UD-IQ2\_M & local & ind+cont/lp & -- & -- & --\\
gemma-4-26B-A4B-it & UD-Q3\_K\_M & local & ind+cont/lp & -- & -- & --\\
gemma-4-26B-A4B-it & UD-Q4\_K\_M & local & ind+cont/lp & -- & -- & --\\
gemma-4-26B-A4B-it & UD-Q5\_K\_M & local & ind+cont/lp & -- & -- & --\\
gemma-4-26B-A4B-it & UD-Q6\_K & local & ind+cont/lp & -- & -- & --\\
gemma-4-31B-it & Q3\_K\_M & local & ind+cont/lp & -- & -- & --\\
gemma-4-31B-it & Q4\_K\_M & local & ind+cont/lp & -- & -- & --\\
gemma-4-31B-it & Q5\_K\_M & local & ind+cont/lp & -- & -- & --\\
gemma-4-31B-it & Q6\_K & local & ind+cont/lp & -- & -- & --\\
gemma-4-31B-it & Q8\_0 & local & ind+cont/lp & -- & -- & --\\
gemma-4-31B-it & UD-IQ2\_M & local & ind+cont/lp & -- & -- & --\\
gemma-4-E2B-it & Q8\_0 & local & ind+cont/lp & -- & -- & --\\
gemma-4-E4B-it & Q8\_0 & local & ind+cont/lp & -- & -- & --\\
gpt-oss-20b & F16 & local & ind+cont/lp & -- & -- & --\\
gpt-oss-20b & Q2\_K & local & ind+cont/lp & -- & -- & --\\
gpt-oss-20b & Q3\_K\_M & local & ind+cont/lp & -- & -- & --\\
gpt-oss-20b & Q4\_K\_M & local & ind+cont/lp & -- & -- & --\\
gpt-oss-20b & Q5\_K\_M & local & ind+cont/lp & -- & -- & --\\
gpt-oss-20b & Q6\_K & local & ind+cont/lp & -- & -- & --\\
gpt-oss-20b & Q8\_0 & local & ind+cont/lp & -- & -- & --\\
phi-4 & F16 & local & ind+cont/lp & -- & -- & --\\
phi-4 & Q2\_K & local & ind+cont/lp & -- & -- & --\\
phi-4 & Q3\_K\_M & local & ind+cont/lp & -- & -- & --\\
phi-4 & Q4\_K\_M & local & ind+cont/lp & -- & -- & --\\
phi-4 & Q5\_K\_M & local & ind+cont/lp & -- & -- & --\\
phi-4 & Q6\_K & local & ind+cont/lp & -- & -- & --\\
phi-4 & Q8\_0 & local & ind+cont/lp & -- & -- & --\\
\multicolumn{7}{@{}l}{\emph{API panel}}\\
claude-fable-5 & API & Anthropic & ind & 8.10 & low & accuracy only\\
claude-fable-5+opus-4.8-fb & API & Anthropic & ind & 8.23 & low & accuracy only\\
claude-opus-4-8 & API & Anthropic & ind & 4.90 & off & accuracy only\\
deepseek-v4-flash & API & NIM & ind+cont/lp & -- & -- & --\\
deepseek-v4-pro & API & NIM & ind & -- & -- & accuracy only\\
dracarys-llama-3.1-70b-instruct & API & NIM & ind+cont/lp & -- & -- & --\\
gemini-2.5-pro & API & Vertex & ind+cont/lp & 8.80 & 128 tok & --\\
gemini-3.1-pro-preview & API & Vertex & ind & 4.30 & 128 tok & accuracy only\\
gemini-3.6-flash & API & Gemini API & ind & 4.37 & managed & accuracy only\\
gemma-2-2b-it & API & NIM & ind & -- & -- & accuracy only\\
gemma-4-26b-a4b-it & API & Gemini API & ind & -- & -- & accuracy only\\
gemma-4-31b-it & API & Gemini API & ind & -- & -- & accuracy only\\
glm-5.2 & API & NIM & ind & -- & -- & accuracy only\\
gpt-4.1-2025-04-14 & API & OpenAI & ind+cont/lp & 1.64 & -- & --\\
gpt-4o-2024-05-13 & API & OpenAI & ind+cont/lp & 4.54 & -- & --\\
gpt-4o-2024-11-20 & API & OpenAI & ind+cont/lp & 2.39 & -- & --\\
gpt-5-2025-08-07 & API & OpenAI & ind & 0.30 & minimal & accuracy only\\
gpt-5.1-2025-11-13 & API & OpenAI & ind+cont/lp & 1.96 & none & --\\
gpt-5.2-2025-12-11 & API & OpenAI & ind+cont/lp & 1.93 & none & --\\
gpt-5.4-2026-03-05 & API & OpenAI & ind+cont/lp & 2.91 & none & --\\
gpt-5.5-2026-04-23 & API & OpenAI & ind+cont/lp & 5.96 & none & --\\
gpt-5.6-sol & API & OpenAI & ind+cont/lp & 4.88 & none & --\\
gpt-oss-120b & API & NIM & ind & -- & -- & accuracy only\\
gpt-oss-20b & API & NIM & ind & -- & -- & accuracy only\\
grok-4.20-0309-non-reasoning & API & xAI & ind & 0.15 & n/a & accuracy only\\
grok-4.5 & API & xAI & ind & 11.34 & low & accuracy only\\
inkling & API & NIM & ind & -- & -- & answers degenerate (excl.); accuracy only\\
laguna-xs-2.1 & API & NIM & ind+cont/lp & -- & -- & --\\
llama-3.1-70b-instruct & API & NIM & ind+cont/lp & -- & -- & --\\
llama-3.1-8b-instruct & API & NIM & ind+cont/lp & -- & -- & --\\
llama-3.2-3b-instruct & API & NIM & ind+cont/lp & -- & -- & --\\
llama-3.3-70b-instruct & API & NIM & ind & -- & -- & accuracy only\\
llama-3.3-nemotron-super-49b-v1 & API & NIM & ind+cont/lp & -- & -- & --\\
minimax-m3 & API & NIM & ind & -- & -- & answers degenerate (excl.); accuracy only\\
mistral-large-3-675b-instruct-2512 & API & NIM & ind+cont/lp & -- & -- & cont.\ unservable (retired)\\
mistral-nemotron & API & NIM & ind+cont/lp & -- & -- & --\\
mistral-small-4-119b-2603 & API & NIM & ind+cont/lp & -- & -- & --\\
mixtral-8x7b-instruct-v0.1 & API & NIM & ind & -- & -- & accuracy only\\
nemotron-3-nano-30b-a3b & API & NIM & ind+cont/lp & -- & -- & --\\
nemotron-3-super-120b-a12b & API & NIM & ind+cont/lp & -- & -- & --\\
nemotron-3-ultra-550b-a55b & API & NIM & ind+cont/lp & -- & -- & --\\
nemotron-mini-4b-instruct & API & NIM & ind & -- & -- & accuracy only\\
nvidia-nemotron-nano-9b-v2 & API & NIM & ind & -- & -- & answers degenerate (excl.); accuracy only\\
qwen3-next-80b-a3b-instruct & API & NIM & ind+cont/lp & -- & -- & --\\
qwen3.5-397b-a17b & API & NIM & ind+cont/lp & -- & -- & --\\
sarvam-m & API & NIM & ind+cont/lp & -- & -- & --\\
solar-10.7b-instruct & API & NIM & ind & -- & -- & answers degenerate (excl.); accuracy only\\
step-3.5-flash & API & NIM & ind & -- & -- & answers degenerate (excl.); accuracy only\\
step-3.7-flash & API & NIM & ind & -- & -- & answers degenerate (excl.); accuracy only\\
\end{longtable}
\end{center}